%% file: Estimator.tex
\documentclass[lettersize,journal]{IEEEtran}
\usepackage{amsmath,amsfonts}
\newtheorem{definition}{Definition}
\usepackage[ruled,vlined]{algorithm2e}
\usepackage{algpseudocode}
\usepackage{balance}
\usepackage{array}
\usepackage{multirow}
\usepackage[caption=false]{subfig}
\usepackage{textcomp}
\usepackage{stfloats}
\usepackage{url}
\usepackage{verbatim}
\usepackage{graphicx}
\usepackage{cite}
\usepackage{color}
\begin{document}

\title{Estimator: An Effective and Scalable Framework for Transportation Mode Classification over Trajectories}

\author{Danlei Hu, Ziquan Fang, Hanxi Fang, Tianyi Li,~\IEEEmembership{Member,~IEEE},\\ Chunhui Shen, Lu Chen, Yunjun Gao,~\IEEEmembership{Member,~IEEE}
\thanks{Manuscript received July 29, 2022. \textit{(Corresponding Author: Lu Chen)}}
\thanks{Danlei Hu, Ziquan Fang, Lu Chen and Yunjun Gao are with the College of Computer Science, Zhejiang University, Hangzhou 310007, China (e-mail: dlhu@zju.edu.cn; zqfang@zju.edu.cn; luchen@zju.edu.cn; gaoyj@zju.edu.cn).}
\thanks{Hanxi Fang is with the School of Earth Science, Zhejiang University, Hangzhou 310007, China (e-mail: hanxif@zju.edu.cn).}
\thanks{Tianyi Li is with the Department of Computer Science, Aalborg University, Denmark (e-mail: tianyi@cs.aau.dk).}
\thanks{Chunhui Shen is with the Alibaba Group, Hangzhou 310007, China (e-mail: zjusch2@163.com).}}

\markboth{IEEE TRANSACTIONS ON INTELLIGENT TRANSPORTATION SYSTEMS, VOL. 23, NO. 1, July 2022}%
{Shell \MakeLowercase{\textit{et al.}}: Estimator: An Effective and Scalable Framework for Transportation Mode Classification over Trajectories}


\maketitle
\begin{abstract}
Transportation mode classification, the process of predicting the class labels of moving objects' transportation modes, has been widely applied to a variety of real-world applications, such as traffic management, urban computing, and behavior study. 
However, existing studies of transportation mode classification typically extract the explicit features of trajectory data but fail to capture the implicit features that affect the classification performance. In addition, most of the existing studies also prefer to apply RNN-based models to embed trajectories, which is only suitable for classifying small-scale data. To tackle the above challenges, we propose an \underline{e}ffective and \underline{s}calable framework for transporta\underline{ti}on \underline{m}ode classific\underline{a}tion over GPS trajec\underline{tor}ies, abbreviated Estimator. Estimator is established on a developed CNN-TCN architecture, which is capable of leveraging the spatial and temporal hidden features of trajectories to achieve high effectiveness and efficiency. Estimator partitions the entire traffic space into disjointed spatial regions according to traffic conditions, which enhances the scalability significantly and thus enables parallel transportation classification. Extensive experiments using eight public real-life datasets offer evidence that Estimator i)  achieves superior model effectiveness (i.e., 99\% \textit{Accuracy} and 0.98 \textit{F1-score}), which outperforms state-of-the-arts substantially; ii) exhibits prominent model efficiency, and obtains 7--40x speedups up over state-of-the-arts learning-based methods; and iii) shows high model scalability and robustness that enables large-scale classification analytics.
\end{abstract}


\input{Introduction}

\input{Definitions}

\input{Method}

\input{Experiments}

\input{RelatedWork}

\input{Conclusion}

\balance
\bibliographystyle{abbrv}
\bibliography{ref}


\end{document}

%% file: Introduction.tex
\section{Introduction}
\label{sec:intro}

With the ubiquitous uses of GPS-equipped devices and mobile computing services (e.g., Twitter, Weibo, and other location-based apps), massive GPS trajectories are collected. The collected data enables to describe the mobility characteristic of moving objects such as bikes, buses, taxis, and pedestrian~\cite{15overview}. Moreover, it has motivated various trajectory-based pattern mining tasks that provide location-based services such as traffic management~\cite{13trafficplanning, 21Survey}, urban computing~\cite{16reviewapplication}, and behavior study~\cite{20scalable}. Being a typical and fundamental pattern mining task, transportation mode classification aims to classify trajectories from different moving objects according to their travel modes, e.g.,  taxi-mode and walking-mode. Note that, a moving object could change its transportation modes during traveling. For example, a city commuter may first take a bus, then take the subway, and finally walk, when commuting from home to his/her working place. Given a trajectory dataset generated by different moving objects without label information (for the sake of trajectory privacy preserving~\cite{18WWWcriminal}), enabling to detect the transportation mode of any trajectory can benefit a variety of services, such as traffic planning~\cite{13trafficplanning}, tourism demand analysis~\cite{19review}, and mobility study~\cite{14mininglocation}. Two motivating examples illustrate this.


\begin{figure}[tb]
 \centering
 \hspace{-0.25cm}
 \subfloat[Traffic Planning]{
  \includegraphics[width=4cm]{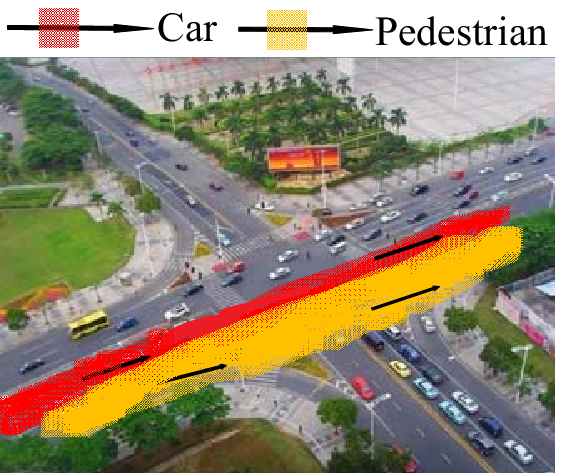}}
 \hspace{1.4mm}
 \subfloat[Travel Demand Analysis]{
  \includegraphics[width=4cm]{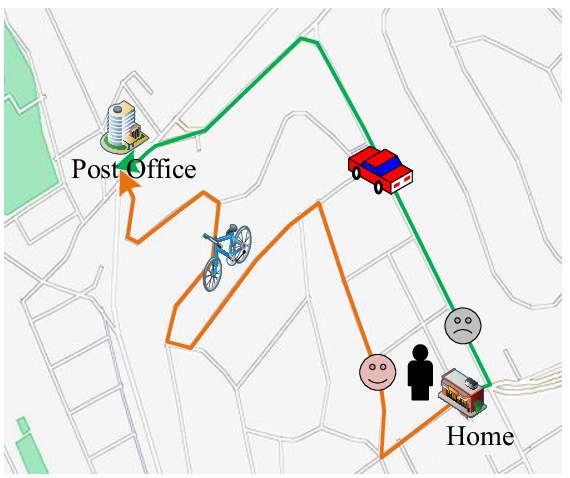}}
 \caption{Transportation Mode Classification}
 \label{fig:example}
 \vspace{-6mm}
\end{figure}

As shown in Fig.~\ref{fig:example}(a), transportation mode classification can find there are more pedestrian trajectories than vehicle trajectories crossing this crossroad in a certain time period. Hence, a flyover or underpass may be constructed to ease traffic congestion. 

Fig.~\ref{fig:example}(b) shows two paths marked by green and orange colors, respectively, between home and post office. With transportation classification over historical trajectories between two locations, we identify that the green path is generally traveled by cars, while the orange path is generally traveled by bikes, and thus, we can recommend the desirable mode for the user according to his/her habits.

Existing studies of transportation mode classification primarily focus on  \textbf{trajectory classification}, which leverages the spatio-temporal features of moving object trajectories to facilitate classification~\cite{18classificationreview, 16reviewapplication, 19review}. Specifically, existing transportation mode classification studies can be generally divided into two categories: traditional machine learning based methods and deep-learning based methods. Methods in the former category manually extract the explicit features (e.g., speed, acceleration, and stay time) of trajectories, feed these features into classifiers (e.g., $k$-nearest neighbour ($k$NN) model~\cite{08zhengyuMobility}, hidden markov (HMM) model~\cite{14HMM}, and support vector machine (SVM) model~\cite{12transportationSVM}). However, they cannot capture the complex and hidden features of trajectories due to their simple and manual-dependent operations~\cite{16Endo, 14mininglocation}. Thus, the effectiveness of the classification is limited, especially when the moving objects share very similar mobility characteristics, such as taxi and private car trajectories.


Deep learning has been successfully applied to facilitate tons of services, such as text classification~\cite{20Text}, image classification~\cite{20image}, and trajectory representation learning~\cite{19representation, 19representation2,21ICCV}, via capturing the non-linear and hidden features. This motivates researchers to study deep-learning based transportation mode classification analyses, which can be mainly categorized into CNN-based methods~\cite{16Endo, 19TKDE}, RNN-based methods~\cite{17ISKE, 19LSTM, 19temporalgraphs}, and CNN-RNN based methods~\cite{19GRU,20CNNwithRNN}.
Convolution neural network (CNN) based methods typically transform a GPS trajectory into a gray image via feature extraction (i.e., pixel calculation), where the pixel values in the image represent the average speed, the time duration, and other features of the trajectory. Then, existing image classification methods are applied to trajectory classification. However, CNN-based methods focus on the spatial dependencies but ignore the temporal correlations among sampling points in a trajectory. To capture it, recurrent neural network (RNN) based models are developed, which is able to learn time-series sequences. As trajectories can be formulated as a type of time-series sequences, RNN based models can feed trajectories into RNNs to extract their temporal features for transportation classification. More recently, a new branch of studies combines CNN and RNN, i.e., CNN-RNN based methods~\cite{16Endo, 19TKDE}, which takes advantages of CNNs and RNNs to consider both spatial and temporal information of trajectories to improve the performance of classification. However,  ST-GRU~\cite{19GRU}, the state-of-the-art method, suffers from poor performance of RNNs when conducting parallel processing, which degrades its model efficiency and scalability.

Classification effectiveness, training efficiency, and model scalability are primary aspects to measure the performance of  learning-based transportation classification~\cite{19TKDE}.
With these in mind, we revisit the problem of transportation mode classification, and propose an \underline{e}ffective and \underline{s}calable framework for transporta\underline{ti}on \underline{m}ode classific\underline{a}tion over GPS trajec\underline{tor}ies, termed \textbf{Estimator}, while address challenges below. 

\textit{Challenge I: How to capture the implicit spatio-temporal characteristics of  trajectories to improve the effectiveness of classification analysis?} In addition to extracting the explicit features (e.g., travel speed and stay time of moving objects) that are already achieved by existing approaches, we aim to extract hidden mobility features to improve classification.
For instance, although taxis and private cars share similar moving features as they both follow vehicle mode, they have different hidden mobility patterns. 
Specifically, private cars typically travel around several fixed locations (e.g., home and working place), while taxis do not. Capturing such hidden features like fixed locations in certain time periods  helps to distinguish the trajectories of taxis and private cars more effectively. However, there are mainly two challenges for capturing the implicit spatio-temporal characteristics of trajectory data. First, the mobility features generally vary a lot across different traffic regions~\cite{21Compression}, which are difficult to be captured by existing deep-learning based studies~\cite{16Endo, 19GRU}.
For example, even for the same taxi, its traveling speed is faster in the suburbs while is slower in the city center. 
However, in this case, existing deep-learning based studies~\cite{16Endo, 19GRU} that treat the entire city-wide training trajectories as a whole cannot identify the varied transportation mode of the taxi's trajectory. Second, the features of moving trajectories are unknown and dynamic during the moving process (e.g., whether it has periodic features and how the features evolve), which makes the feature embedding challenging. 

\textit{Challenge II: How to overcome the limited parallel processing of RNNs to improve the efficiency and scalability of trajectory classification?} Although the state-of-the-art CNN-RNN based method~\cite{19GRU} outperforms both CNN-based and RNN-based methods, it still exhibits limited capacity of parallel processing  due to the inherent computation mechanism of RNNs. Specifically, RNNs scan the sampling points of a trajectory one by one, meaning that RNNs cannot process the next point until the current point has been processed.
Consequently, the RNNs-depended methods fail to deal with large-scale classification. Furthermore, although CNNs are able to process sampling points of a trajectory in parallel, the size of convolution kernel limits the ability of capturing spatio-temporal correlations due to the different lengths of trajectories. Put differently, existing CNN based methods cannot efficiently process sequences with varying lengths, degrading the scalability of classification.


To address the first challenge, we extract periodic features of taxis and private cars to distinguish their mobility features. In addition, we partition the entire traffic space into disjointed regions to handle varying traffic conditions. During this process, different features in partitions with different traffic environment are extracted, and thus the effectiveness of classification can be enhanced (cf. ablation study in Section~\ref{sec:exe}). To tackle the second challenge, we develop Estimator, a unified CNN-TCN architecture for effective and efficient transportation mode classification analyses. We extract spatial features by CNN model, and propose to employ temporal convolutional network (TCN)~\cite{18TCN} to capture the temporal informatoin of trajectories efficiently.
Further, based on the partitioned traffic space, we extend CNN-TCN architecture for parallel transportation mode classification to improve model scalability. Overall, this paper makes the following contributions: 

\begin{itemize}\setlength{\itemsep}{-\itemsep}
    \item \textit{CNN-TCN Architecture.} We construct a CNN-TCN architecture to capture the spatio-temporal mobility characteristics of moving objects for transportation mode classification. To the best of our knowledge, this is the first proposal to apply TCN to trajectory data.

    \item \textit{Hidden Mobility Feature Extraction.} Estimator considers both explicit and implicit mobility features. In terms of explicit features, we capture speed, azimuth, stay time, etc. In terms of implicit features, we capture periodic patterns of taxis and private cars to distinguish different mobility characteristics. Based on this, Estimator enables fine-grained feature embedding to improve effectiveness.

    \item \textit{Partition and Parallel Training.} We partition the whole city into disjointed areas, i.e., urban center, urban area, and suburb. Moreover, we extend CNN-TCN architecture for parallel transportation mode classification to further improve the model scalability.

    \item \textit{Extensive Experiments.} We conduct extensive experiments on eight real-world trajectory datasets that offer insight into Estimator and demonstrate that it is able to outperform the state-of-the-art competitors in terms of machine learning based and deep learning based methods.

\end{itemize}

\vspace{-1mm}
The rest of the paper is organized as follows. 
Section~\ref{sec:problem} presents the problem statement. Section~\ref{sec:method} details our proposed framework Estimator. Section~\ref{sec:exe} reports the experimental results, and the related work is reviewed in Section~\ref{sec:related}. Finally, Section~\ref{sec:conclusion} concludes the paper.

%% file: Definitions.tex
\section{Preliminaries and Problem Definition}
\label{sec:problem}

We proceed to introduce preliminary definitions, based on which, we define the problem of trajectory mode classification.

Trajectories can be collected by smartphone sensors~\cite{20CNNwithRNN} or GPS-based equipment~\cite{19TKDE}. Different from GPS-based trajectories which is mainly consisted of the locations and timestamps of sampling points, the sensor-based trajectories contain not only timestamps but also the information of sensor gravity, gyroscope, ambient light and so on. However, the complete records are often not available. In this case, it is necessary to discover transportation mode classification only using locations and timestamps. Thus, this paper focuses on GPS trajectories. A raw trajectory is defined as follows:


\begin{definition}\label{defn:raw trajectory}
    {\bf (Raw Trajectory)} \textit{A raw trajectory $T$ is a time-ordered sequence that consists of GPS points  $p_i (1 \leq i \leq N)$, i.e., $T=\{p_1, p_2, ..., p_N\}$, where $N$ denotes the length of a trajectory. Each GPS point $p_i$ is  in the form of $\left \langle id, l_o, l_a, t_s \right \rangle$, where $id$ is the identifier,  $l_o$ is the longitude, $l_a$ is the latitude, and $t_s$ is the time when $p$ occurs.}
\end{definition}

A trajectory may contain thousands of GPS points (i.e., $N$ is extremely large), and thus, its transportation mode may change during the traveling of the moving object. Following existing methods, trajectory segmentation can be used to detect the evolution of transportation mode. Specifically, we use the stay point detection method~\cite{08staypoint} to split a raw trajectory into several successive segments to ensure each trajectory segment only features one mode. Unless stated otherwise, we assume that trajectories are segmented in the rest of paper.  

\begin{definition}\label{defn:mapped trajectory}
    {\bf (Mapped Trajectory)} \textit{A mapped trajectory is a trajectory image, which consists of $w\times h$ uniform grid cells. The pixels of each grid cell are represented by an RGB tuple of the mean azimuth, average speed, and stay time of all trajectory points located in this grid cell.}
\end{definition}

\begin{figure}[tb]
	\centering
	\vspace{-3mm}
	\includegraphics[width=0.37\textwidth]{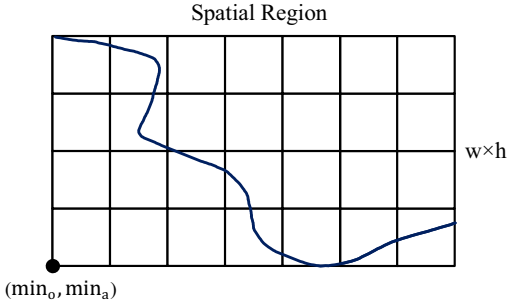}
	\vspace{-3mm}
	\caption{Trajectory Spatial Region}
	\label{fig:image}
    \vspace{-5mm}
\end{figure}

\begin{algorithm}[t]
	\caption{Mapped Trajectory Generation}\label{alg:map algorithm}
    \small
    \LinesNumbered

	\KwIn{a trajectory $T=\{p_1, \dots, p_N\}$, the number of grid cells $w \times h$, left-bottom corner $(\textit{min}_o, \textit{min}_a)$}
    \KwOut{the mapped trajectory ${I_s}$}
    {Initialize} ${I_s\in \mathbf{R}^{w\times h\times 3}\leftarrow 0}$, ${I_M\in \mathbf{R}^{w\times h\times 4}\leftarrow 0}$\\

    \For{each $p_i$ in $T$}
	{
       $ x_i = \lfloor\frac{p_i.l_o- \textit{min}_o }{h}\rfloor$, $ y_i = \lfloor\frac{p_i.l_a- \textit{min}_a }{w}\rfloor$ \\
            update $[n, p_s, p_e, d]$ of $I_M(x_i,y_i)$ // the number of points, start point, end point and covered distance in grid cell $(x_i, y_i)$\\
	}
   
   \For{each grid cell $(x, y)$}
   {
    ${p_s=I_M(x,y,1)}$\\
    ${p_e=I_M(i,j,2)}$\\
    $I_s(i,j,0)\leftarrow$ Azimuth angle between ${p_{s}}$ and ${p_{e}}$ \\
                ${ST=p_{e}.t_s-p_{s}.t_s}$ // stay time\\
                ${I_s(i,j,1)\leftarrow \frac{I_M(i,j,3)}{ST}}$  // speed\\
                ${I_s(i,j,2)\leftarrow ST}$											\\
         } 
    \Return{${I_s}$}
\end{algorithm}

The spatial region traversed by a mapped trajectory is a minimum bounding rectangle to include all of its GPS points, as shown in Fig.~\ref{fig:image}. However, images must have same number of pixels that are directly fed to the training model. Thus, we fix the number of grid cells to $w \times h$ in the mapped image~\cite{16Endo}. 
For each sample point $p_i = \left \langle l_o, l_a, t_s \right \rangle$ in a trajectory, its mapped grid cell $(x_i, y_i)$ can be calculated as:

\begin{equation}\small
      x_i = \lfloor\frac{p_i.l_o- \textit{min}_o}{h}\rfloor,  y_i = \lfloor\frac{p_i.l_a- \textit{min}_a}{w}  \rfloor.
\end{equation}
where ($\textit{min}_a$, $\textit{min}_o$) is the left-bottom corner of the spatial region of this trajectory. For each grid cell $(x, y)$, we collect the number of points $n$ in $(x, y)$, the start point $p_s$ (i.e., the earliest point occurred in $(x, y)$), the end point $p_e$ (i.e., the latest point occurred in $(x, y)$), and the covered distance $d$ (i.e., the trajectory length in $(x, y)$).  

\begin{figure}[tb]
	\centering
	\vspace{-3mm}
	\hspace{-0.25cm}
	\includegraphics[width=0.4\textwidth]{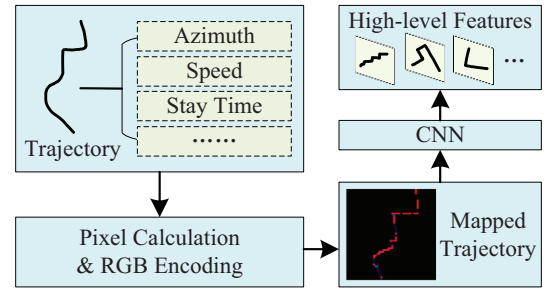}
	\vspace{-2mm}
	\caption{Trajectory Mapping}
	\vspace{-4mm}
	\label{fig:map}
\end{figure}

As shown in Fig.~\ref{fig:map}, we first map an original trajectory into a mapped image according to Definition~\ref{defn:mapped trajectory}, and then calculate its moving features (i.e., pixels) in each grid cell of the mapped image. The different colors in image grid cells represent different mobility characteristics of the moving object. We detail how to transform an original trajectory into a trajectory image in Algorithm~\ref{alg:map algorithm}. Finally, the mapped image is fed to CNN in order to obtain high level features.

\begin{definition}
\label{defn:TMC}
    {\bf (Transportation Mode Classification)} \textit{Given a set of raw trajectories generated from  moving objects, i.e., $\mathcal{T} = \{T_1, T_2, ..., T_M\}$, we aim to train a classifier to classify them into different groups based on their transportation manners below:}
\end{definition}
\vspace{-6mm}

\begin{align}\small
f(\theta): \mathcal{T} \rightarrow {\mathcal{T}_1, \mathcal{T}_2, ..., \mathcal{T}_K}.
\end{align}
where $\mathcal{T}_j$ $(1 \leq j \leq K)$ denotes a specific transportation mode such as bike, taxi, private-car, bus, or subway, $K$ denotes the total number of transportation manners, and $\theta$ is the target parameter set of Estimator (to be detailed in Section~\ref{sec:method}).

%% file: Method.tex
\section{The Proposed Method}
\label{sec:method}

We first present the CNN-TCN architecture, and then introduce the optimization techniques including hidden mobility extraction and data partition, which are the basis of Estimator.

\begin{figure*}[tb]
	\centering
	\vspace{-2mm}
	\includegraphics[width=0.85\textwidth]{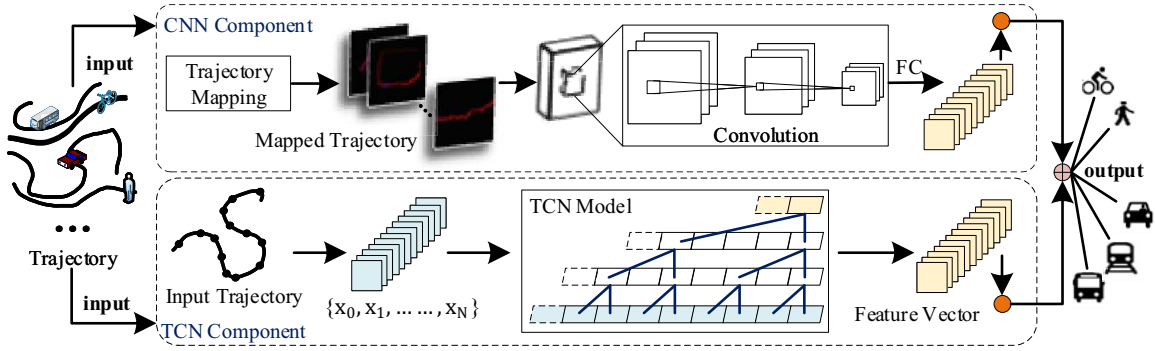} 
	\vspace{-2mm}
	\caption{The CNN-TCN Architecture for Estimator}
    \vspace{-2mm}
	\label{fig:overview}
	\vspace{-3mm}
\end{figure*}

\subsection{CNN-TCN Model}
We combine CNN and TCN to obtain a CNN-TCN model, as shown in Fig.~\ref{fig:overview}. Specifically, we feed the given trajectories into CNNs and TCNs simultaneously to capture the spatial and temporal dependencies of trajectory data for transportation mode classification. We detail the two components below.

\textbf{CNN Component.} As depicted in the upper part of Fig.~\ref{fig:overview}, we transform raw trajectories into mapped trajectories (trajectory images) through trajectory mapping, i.e., $T \rightarrow MT$ (cf. Definition~\ref{defn:mapped trajectory}). During the mapping, the spatial features of trajectories can be obtained and represented as pixels of images. Next, we feed the mapped trajectories into CNNs for representation learning, i.e., $MT \rightarrow CT$. Specifically, we utilize ResNet50~\cite{Resnet50} for trajectory embedding while capturing the mobility features for transportation mode classification. The training process is detailed as follows.   

\begin{align}\small
{MT}_{l+1}&=h({MT}_{l})+\mathcal{F_C}\left({MT}_{l}, W_{l}\right) \label{eq:19},\\
CT=&{MT}_{0}+\sum_{i=0}^{L-1} \mathcal{F_C}\left({MT}_{i}, W_{i}\right) \label{eq:20},
\end{align}
where the mapped trajectory ${MT}$ is the input of CNN model, ${CT}$ is the output of CNN model, $h(·)$ and ${\mathcal{F_C(·)}}$ denote identity mapping and residual in ResNet, respectively, ${W_l}$ denotes the convolution operation in the $l$-th layer, and $L$ denotes the total number of layers. Based on Eqs.~\ref{eq:19}-\ref{eq:20}, we provide the CNN-based loss function below. 

\begin{equation}\small
L_C=-\frac{1}{n}[CT\ln \hat{CT}+(1-CT) \ln (1-\hat{CT})],
\end{equation}
where CT and $\hat{CT}$ represent the labeled transportation mode and predicted transportation mode, respectively.

\textbf{TCN Component.} Generally, each token in TCNs only contains a timestamp. However, we feed a large number of trajectories into TCNs, and thus, massive timestamps are included in each token. Different from general time series processing, we use multi-channel to record trajectory sequence. Specifically, when feeding trajectories into CNNs (in the upper part) to capture their spatial dependencies, we also feed raw trajectories into TCNs (in the lower part) to capture their temporal correlations at the same time, i.e., $T \rightarrow TT$, as shown in  the lower part of Fig.~\ref{fig:overview}. As observed, the TCN model enables reading/embedding an input trajectory sequence in a parallel fashion. The detailed process is as follows.  
\vspace{-2mm}

\begin{align}\small
D(p)=\left({T} *_{d} f\right)(p)&=\sum_{i=0}^{k_s-1} f(i) \cdot {T}_{p.id-d \cdot i} \label{eq:22},\\
TT= Activation&(T^{'} + \mathcal{F}_T(T^{'})),
\end{align}
where $D(\cdot)$ denotes the dilated convolution operation (i.e., $T \rightarrow T^{'}$) on the point ${p}$ of the trajectory sequence ${T}$, ${d}$ is the dilated factor, ${k_s}$ is the size of filter ${f}$ (i.e., kernel size), \textit{Activation} denotes the active function (i.e., \textit{ReLU}), and $TT$ is the output of TCN model.  
The dilated convolution operation facilitates to capture all the time span of ${T}$. In Eq.~\ref{eq:22}, ${T_{p.id -d\cdot i}}$ denotes the history token of ${T}$, in order to capture the information in the past time. ${\mathcal{F}_T(·)}$ denotes the residual in TCN. In experiments, we set ${k_s=3}$ and ${d=[1,2,4,8]}$ 
following previous work~\cite{18TCN}. The TCN-based loss function  ${L_T}$ is computed as follows:
\begin{equation}\small
    L_T =-\frac{1}{n} [TT\ln \hat{TT}+(1-TT) \ln (1-\hat{TT})],
\end{equation}
where ${TT}$ and ${\hat{TT}}$ represent the labeled transportation mode and predicted transportation mode, respectively.

\textbf{CNN-TCN Combination.}
We propose the overall training loss function of CNN-TCN model by summing up the CNN-based Loss and TCN-based Loss.
\begin{align}\small
\mathcal{L}=&\alpha L_C+\beta L_T,\\
\alpha=\frac{e^{{r}_{1}}}{e^{{r}_{1}}+e^{{r}_{2}}}&, \beta=\frac{e^{{r}_{2}}}{e^{{r}_{1}}+e^{{r}_{2}}}.
\end{align}where $\alpha$ and $\beta$ are coefficients, and ${r_1}$ and ${r_2}$ denote the accuracy of CNN-based  model and TCN-based model. We calculate the weight coefficients by softmax function.

\subsection{Hidden Mobility Extraction}
As discussed in Section~\ref{sec:intro}, moving objects such as taxis and private cars share similar mobility characteristics (e.g., moving speed) since they both belong to vehicle transportation, which makes it challenging for existing methods and our CNN-TCN architecture to distinguish them apart. Fortunately, we observe that the private cars show period moving features (i.e., hidden temporal features) in geographic space.
Motivated by this, given a private car's trajectory $T = \{p_1, p_2, ..., p_N\}$, we first extract its time sequence based on the temporal dimension, $T^{(t_s)} = \{p_1.t_s, p_2.t_s, ..., p_N.t_s\}$. Next, we employ the seasonal-trend decomposition procedure (STL)~\cite{1990STL} to capture the period features hidden in $T^{(t_s)}$.
\par STL methods decompose the time series sequence to get trend data, seasonal data and residual data, which is denoted by ${TR_v}$, ${S_v}$, ${R_v}$, respectively. Based on these, the sequence data ${Y_v}$ is formulated as: $Y_v=TR_v+S_v+R_v$.

Note that periodic data and seasonal data are quite similar, except that the time frequency of seasonal data fluctuations is fixed, while that of periodic data is not. 
Considering the time correlation among different trajectories of one moving object, we aim to extract the seasonal features of taxis and private cars $S_v$. The first step is detrending, which subtracts the trend component ${TR_v}^{(k)}$ in the previous iteration (i.e. $k$-th iteration) from the sequence data $Y_v$:


\begin{equation}\small
    C_v^{k+1} = {Y_v-{TR_v}^{(k)}},
\end{equation}

Then, cycle-subseries smoothing is used to get the smoothing ${{C_v}^{(k+1)}}$  (i.e., temporary seasonal series) after extending the cycle-subseries forward and backward for one cycle respectively, while the cycle-subseries low-pass filtering is used to obtain ${{L_v}^{(k+1)}}$. Here, cycle-subseries is a sequence composed of points in the same position in each cycle. Based on the above illustration, the seasonal data ${{S_v}^{(k+1)}}$ is obtained:
 \begin{equation}\small
 {S_v}^{(k+1)}={C_v}^{(k+1)}-{L_v}^{(k+1)} \label{Sv},
 \vspace{-2mm}
 \end{equation}
In each iteration of inner loop, the seasonal data updates after a series of detrending and smoothing steps as below: 
 \begin{equation}\small
      S_v^{(k+1)}=S_v^{(k)}\label{Sv+1},
      \vspace{-2mm}
 \end{equation}
\par Based on the above processing, we capture the period features (i.e. seasonal data in the trajectory) hidden in $T^{(t_s)}$, i.e., $T^{(t_s)} \rightarrow H$. Next, we add it to the raw trajectory $T$ to get $\hat{T}$, which denotes the trajectory with additional period features $H$. For each point $p_i$ in $T$, $\hat{p_i}.t_s$ in $\hat{T}$ is calculated as:
\begin{align}\small
\begin{split}
\hat{p_i}.t_s&=p_i.t_s + w\times H_i \\
(1\leq i\leq &N, p_i\in T, H_i\in S_v).
\end{split}
\end{align}
\par Finally, we replace the raw trajectory $T$ with $\hat{T}$ when feeding trajectory to TCNs. This way, Estimator enables to capture the implicit mobility features between taxis and private cars, which significantly improves the classification performance.

\begin{figure}[tb]
	\centering
	\vspace{-1mm}
	\subfloat[Serial Computing]{
		\includegraphics[width=0.235\textwidth]{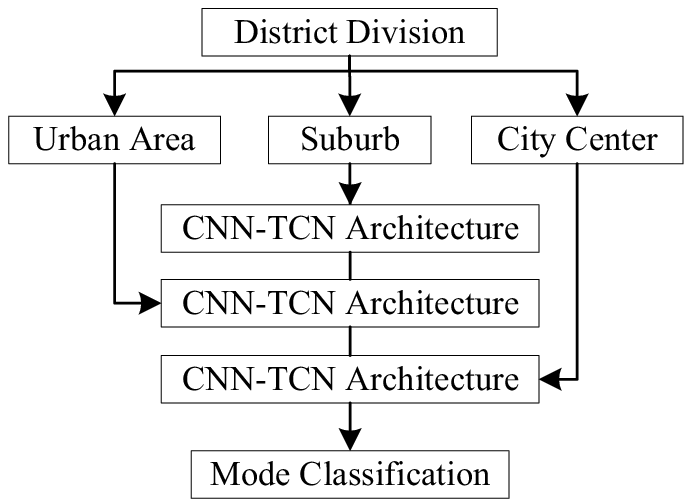}}
	\hspace{-2mm}
	\subfloat[Parallel Computing]{
		\includegraphics[width=0.235\textwidth]{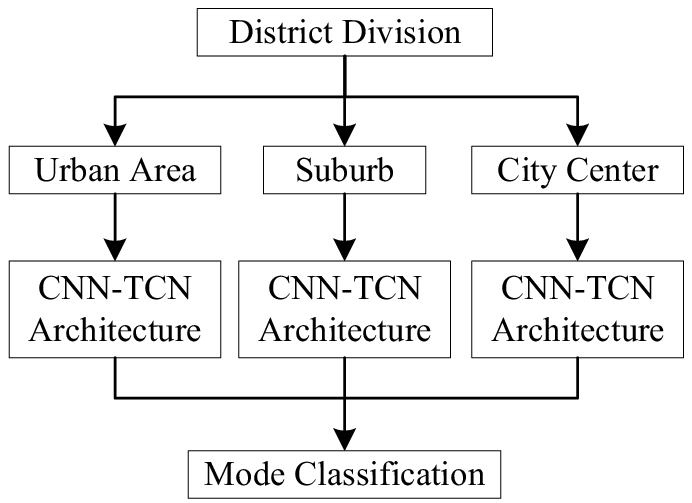}}
	\caption{Model Training Architecture}
	\vspace{-5mm}
	\label{fig:centraAnddistri}
\end{figure}

\subsection{Data Partition}
We detail another optimization for CNN-TCN architecture, i.e., data partition and parallel computing.

Considering the varying traffic conditions in a city, (e.g., a car may drive fast in suburbs while drive slow in city blocks), we split the entire city region into the urban area, suburb, and urban center by administrative divisions and the traffic road layout. Based on this, we train  CNN-TCN architecture in the three partitions to capture mobility features of different traffic conditions, as shown in Fig.~\ref{fig:centraAnddistri}(a). However, such serial model training in three partitions is inefficient. Therefore, we design a parallel training mechanism, as depicted in Fig.\ref{fig:centraAnddistri}(b). Specifically, we divide the training data into three sub-datasets based on their spatial locations. Note that, a trajectory may include thousands of GPS points and may cross more than one partitions. In this case, we assign this trajectory to the partition containing maximum number of GPS points. After the parallel model training, we collect and fuse the results (i.e., compute the union of the labeled or predicted transportation modes in each partition) from each partition, as we target the city-level transportation mode classification.

%% file: Experiments.tex
\section{Experiments}
\label{sec:exe}
We report on extensive experiments aimed at evaluating the performance of Estimator. We first present the experimental settings. Then, we compare the classification performance of Estimator with the state-of-the-arts. Next, we study the model efficiency and scalability. Finally, we report the effects of hidden mobility and district partition on classification performance. All implementation codes have been released online$\footnote{\footnotesize https://github.com/ZJU-DAILY/Estimator}$. 

\begin{table}[t]
\caption{Statics of Evaluated Datasets}
\vspace{-2mm}
\begin{tabular}{|p{1.6cm}<{\centering}|p{1.5cm}<{\centering}|p{1.2cm}<{\centering}|p{1.2cm}<{\centering}|p{1.2cm}<{\centering}|}
\hline
\textbf{Datasets}                    & \textbf{Number of Trajectories} & \textbf{Max. Length} & \textbf{Min. Length} & \textbf{Ave. Length} \\ \hline
\textbf{Walk}        & 6,600                   & 19837                    & 20                    & 250                    \\ \hline
\textbf{Bike}        & 3,787                   & 22080                   & 22                    & 500                   \\ \hline
\textbf{Bus}         & 6,002                   & 19605                    & 24                    & 450                    \\ \hline
\textbf{Subway}      & 1,017                   & 21995                    & 21                    & 450                    \\ \hline
\textbf{Private car} & 2,662                   & 8395                    & 21                   & 400                    \\ \hline
\textbf{Taxi}        & 1,661                   & 10841                    & 19                    & 450                    \\ \hline
\textbf{Train}       & 3,251                   & 38294                    & 21                    & 800                    \\ \hline
\textbf{Geolife}       & 24,980                   & 38294                    & 19                    & 450                    \\ \hline
\end{tabular}
\label{tab:dataset}
\vspace{-6mm}
\end{table}

\subsection{Experimental Settings}

\textbf{Datasets.} We verify Estimator using Geolife dataset~\cite{Geolife}$\footnote{\footnotesize https://www.microsoft.com/en-us/download/details.aspx?id=52367}$, which contains 17621 GPS points collected from 182 users. To the best of our knowledge, this is the only public GPS trajectory dataset with transportation mode labels~\cite{19TKDE}, which has been widely used by existing studies of transportation mode classification of GPS trajectories~\cite{19GRU,17ISKE,19TKDE}. Geolife dataset contains eleven labeled kinds of transportation modes. Nevertheless, due to the small percentage of the boat, run, airplane, and motorcycle, we mainly detect the remaining seven modes, i.e., walk, bike, bus, subway, private car, taxi, and train. As shown in Table~\ref{tab:dataset}, we generate seven additional datasets from GeoLife (each dataset contains one transportation mode). Based on this, we train and test Estimator on eight datasets (i.e., Walk, Bike, Bus, Subway, Private Car, Taxi, Train and Geolife) respectively, in order to better explore the performance of Estimator for each transportation mode and prove it is optimal for 
discovering both single transportation mode and multiple transportation modes.

\textbf{Baselines.} We compare Estimator with representative traditional machine-learning based methods (i.e., ${k}$NN, RF, SVM, and DT) and the state-of-the-art deep-learning based methods (i.e., ST-GRU and SECA). Specifically, the baselines are: 

\begin{table*}[t]
\vspace{-2mm}
\caption{Classification Performance (i.e., \textit{ACC} and \textit{F1}) of all the Methods}
\vspace{-2mm}
\begin{tabular}{|p{1.8cm}<{\centering}|p{1.4cm}<{\centering}|p{0.6cm}<{\centering}|p{1.2cm}<{\centering}|p{1.2cm}<{\centering}|p{1.2cm}<{\centering}|p{1.2cm}<{\centering}|p{1.2cm}<{\centering}|p{1.2cm}<{\centering}|p{1.2cm}<{\centering}|p{1.2cm}<{\centering}|}
\hline
\multicolumn{2}{|c|}{\makebox[0.05\textwidth][c]{Methods}}                                        &     & \textbf{Geolife}                     &\textbf{ Walk}  &\textbf{ Bike}  & \textbf{Bus}   & \textbf{Subway} &\textbf{ Pri.Car} & \textbf{Taxi}  & \textbf{Train} \\ \hline
                                         &                           & ACC & 0.532                        & 0.613     & 0.468     & 0.464     & 0.572      & 0.488       & 0.561     & 0.472     \\ \cline{3-11} 
{Machine}
                                    & {SVM}     & F1  & 0.48                           & 0.55     & 0.51     & 0.49     & 0.54      & 0.47       & 0.51     & 0.43     \\ \cline{2-11} 
                                         &                           & ACC & 0.579                       & 0.802  & 0.445  & 0.978  & 0.934   & 0.524    & 0.703  & 0.533  \\ \cline{3-11} 
{Learning}								 		& {$k$NN}     & F1  & 0.53						& 0.78  & 0.78  & 0.13  & 0.84   & 0.41    & 0.52  & 0.42  \\ \cline{2-11} 
                                         &                           & ACC & 0.694                        & 0.732  & 0.556  & 0.433  & 0.721   & 0.345    & 0.597  & 0.556  \\ \cline{3-11} 
{Based Methods}                                         & {DT}      & F1  & 0.58                        & 0.76  & 0.77  & 0.46  & 0.68   & 0.50    & 0.54  & 0.57  \\ \cline{2-11} 
                                         &                           & ACC & 0.783                        & 0.842  & 0.396  & 0.982  & 0.911   & 0.845    & 0.833  & 0.645  \\ \cline{3-11} 
										&{RF}      & F1  & 0.62                        & 0.81  & 0.78  & 0.44  & 0.76   & 0.59    & 0.61  & 0.58  \\ \hline
                                         &                           & ACC & 0.768                        & 0.746  & 0.824  & 0.756  & 0.732   & 0.818    & 0.822  & 0.856  \\ \cline{3-11} 
{Deep Learning} & {SECA}    & F1  & 0.76                        & 0.81  & 0.84  & 0.75  & 0.70   & 0.78    & 0.77  & 0.80  \\ \cline{2-11} 
                                         &                           & ACC & 0.912                        & 0.882  & 0.843  & 0.924  & 0.931   & 0.897    & 0.789  & 0.842  \\ \cline{3-11} 
{Based Methods}                                         & {ST-GRU}  & F1  & 0.88                        & 0.88  & 0.92  & 0.90  & 0.91   & 0.84    & 0.82  & 0.86  \\ \cline{2-11} 
                                         &                           & ACC & \textbf{0.992}                       & \textbf{0.993} &\textbf{0.995} & \textbf{0.997} & \textbf{0.982}  & \textbf{1.000}   & \textbf{0.979} & \textbf{0.950} \\ \cline{3-11} 
										&{Estimator} & F1  & \textbf{0.98}                        & \textbf{0.97}  & \textbf{0.98}  & \textbf{0.98}  & \textbf{0.97}   & \textbf{0.97}    & \textbf{0.98}  &\textbf{0.92}  \\ \hline
\end{tabular}
\label{tab:performance}
\vspace{-6mm}
\end{table*}

\begin{table}[]
\caption{Model Efficiency Evaluation}
\vspace{-2mm}
\begin{tabular}{|p{1.0cm}<{\centering}|p{0.8cm}<{\centering}|p{0.8cm}<{\centering}|p{0.8cm}<{\centering}|p{0.8cm}<{\centering}|p{0.8cm}<{\centering}|p{0.8cm}<{\centering}|p{0.8cm}<{\centering}|p{0.8cm}<{\centering}|}
\hline
\textbf{Dataset}  & \multicolumn{2}{c|}{\textbf{SECA}} & \multicolumn{2}{c|}{\textbf{ST-GRU}} & \multicolumn{2}{c|}{\textbf{Estimator}} \\ \cline{2-7}
 	            & TTIME                & PTIME               & TTIME                    & PTIME                   & TTIME                     & PTIME                    \\ \hline
Walk            & 9.36                & 0.30               & 51.47                 & 2.34                & \textbf{0.95}                & \textbf{0.29}                 \\ \hline
Bike              & 10.73                & 0.29               & 53.41                 &2.73                & \textbf{1.03}                 & \textbf{0.28}                 \\ \hline
Bus               & 10.11                & 0.29               & 57.16                 & 2.31                & \textbf{0.60}                 & \textbf{0.28}                \\ \hline
Subway            & 5.79                & 0.29               &34.53                 &1.36                & \textbf{0.50}                 & \textbf{0.28}                 \\ \hline
Pri. Car       & 5.33                & 0.30              & 37.93       	& 1.68                & \textbf{0.60}                & \textbf{0.28}                 \\ \hline
Taxi              & 3.57                & 0.29               & 31.63               & 1.45                & \textbf{0.43}                 & \textbf{0.27}                 \\ \hline
Train             & 3.25                & 0.29               & 28.31                & 1.11                & \textbf{0.37}                 & \textbf{0.27}               \\ \hline
GeoLife         & 20.69               & 0.32              & 117.49                  & 5.89                & \textbf{3.23}                 & \textbf{0.30}                \\ \hline
\end{tabular}
\label{tab:efficiency}
\vspace{-6mm}
\end{table}

\begin{itemize}\setlength{\itemsep}{-\itemsep}
\item \textbf{$k$NN~\cite{08zhengyuMobility}}, a classic method to distinguish different transportation modes using trajectory similarity computation; 
\item \textbf{RF~\cite{10zhengyu}}, a classifier that uses multiple trees to identify different trajectories; 
\item \textbf{SVM~\cite{12transportationSVM}}, a linear classifier for classification defined according to the largest interval in the feature space; 
\item \textbf{DT~\cite{08zhengyuTransportation}}, a classifier that constructs a decision tree to discover the classification rules hidden in trajectory data; 
\item \textbf{SECA~\cite{19TKDE}}, a semi-supervised convolutional autoencoder to learn the spatio-temporal features of trajectories for transportation mode classification; 
\item \textbf{ST-GRU~\cite{19GRU}}, a GRU-based model with a 1D-CNN to capture the local correlations, which offers the state-of-the-art results for transportation mode classification.
\end{itemize}

All baselines are trained using the hyperparameters that achieves the highest performance. Here, we list the hyperparameters corresponding to each deep-learning methods.

\textbf{SECA~\cite{19TKDE}.} We set the number of GPS points to 20, the total distance of a segment to 150 meters and total duration time of the segment less than 1 minutes. In addition, the hyperparameters $\alpha$ and $\beta$ are initialize to 1 and are both varied from 1 to 0.1 during training.

\textbf{GRU~\cite{19GRU}.} We set the soft-embedding dimension to 16 and the segment length to 10. In ST-GRU model, the kernel size in all 1D-convolution operators is set to 3. The model is trained by optimising the cross-entropy loss function on the labels, where the learning rate is set to 0.01. In addition, the model is optimised using Adam optimiser for the first 30 epochs and is finely tuned using SGD optimiser for 50 epochs, where the learning rate is set to 0.0001.

\textbf{Preprocessing.} The preprocessing mainly involves trajectory mapping. Existing methods \cite{16Endo,18classificationreview} map the raw trajectories into gray images to extract features representing spatial coordinates and stay time. We enhance this strategy by mapping raw trajectories into color images and extract features representing spatial coordinates, stay time, azimuth, and average speed. The transformed color images are the input of the CNN model.

\textbf{Hyperparameters.} We set the hyperparameters based on the model performance. Specifically, we set the granularity of the grid cell as $40\times 40$. The number of hidden layers is 8 and the number of hidden units of TCN is 25. We adopt Adam optimizer~\cite{kingma2014adam} for model training. In addition, the batch size is set to 64, the epoch is set to 20, the gradient clip is set to (i.e., no clip), the dropout is set to 0.05, and the initial learning rate is set to 0.002. We use 80\% 
and 20\% of each mode dataset for training and testing
for OCT-LSTM, respectively.
We implement Estimator in Python and Pytorch. All experiments are conducted on a server with GeForce RTX 3090, 2.40GHz GPU, and 24-GB RAM.

\textbf{Evaluation metrics.} Following most of existing trajectory classification studies~\cite{19review}, we use $\textit{ACC}$ and ${F1}$ to evaluate the quality, and use $\textit{TTIME}$ and $\textit{PTIME}$ to evaluate the efficiency. $ACC=\frac{100\%}{N} \sum_{i = 1}^{N} \left (1-\left|\frac{y_{i}-\hat{y}_{i}}{y_{i}}\right|\right)$, where $y_i$ is the actual transportation mode (i.e., ground truth) of trajectory $T_i$, $\hat{y}_i$ denotes the predicted transportation mode of trajectory $T_i$, and $N$ is the number of trajectories. \textit{ACC} can also BE derived by $\frac{TP+TN}{TP+TN+FN+FT}$, where \textit{FP}, \textit{FN}, \textit{TP}, \textit{TN} denote the number of false positive samples, false negative samples, true positive samples, and true negative samples, respectively. ${F1=\frac{2*\textit{Precision}*\textit{Recall}}{\textit{Precision}+\textit{Recall}}=\left(\frac{2}{2+\frac{F P}{T P}+\frac{F N}{T P}}\right)}$, where $\textit{Precision}=\frac{TP}{TP+TN}$ and $\textit{Recall}=\frac{TP}{TP+FN}$.
$\textit{TTIME}$ and $\textit{PTIME}$ are the training and testing time, respectively, which measure the efficiency of transportation mode classification methods. Obviously, the larger \textit{ACC} and \textit{F1} are, the better classification quality is; and the smaller \textit{TTIME} and \textit{PTIME} are, the better classification efficiency is. Note that, we perform binary classification on seven datasets with single mode, while use multi-class classification on Geolife dataset.
\vspace{-5mm}
\subsection{Classification Performance Evaluation}
We compare Estimator and all baselines in terms of the classification performance (i.e., $\textit{ACC}$ and $\textit{F1}$).
Table~\ref{tab:performance} reports the results.
As observed, deep-learning based methods  perform better than machine-learning based methods. This is because, machine-learning based methods rely on manual feature extraction and cannot capture the non-linear dependencies in trajectory data. Specifically, $k$NN and DT perform worst in all of the baselines, as they simply treat trajectories as time-series sequences, ignoring the temporal correlations between trajectory points. 
On the other hand, 
Estimator achieves the best performance in all the cases. This is because, compared with other deep-learning based methods, Estimator considers hidden mobility features and varying traffic conditions that enable embedding and learning more effectively. Based on this, it is easier for Estimator to identify transportation mode via corresponding captured features of different trajectories.

\begin{figure*} [tb]
	\centering
	\vspace{-4mm}
	\includegraphics[width=0.75\textwidth]{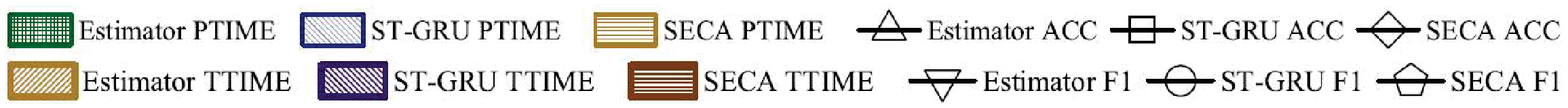}\\
    \vspace{-6mm}
    \hspace{1cm}\\
	\hspace{-0.25cm}
	\subfloat[Walk/ACC/PTIME]{
		\includegraphics[width=0.24\textwidth]{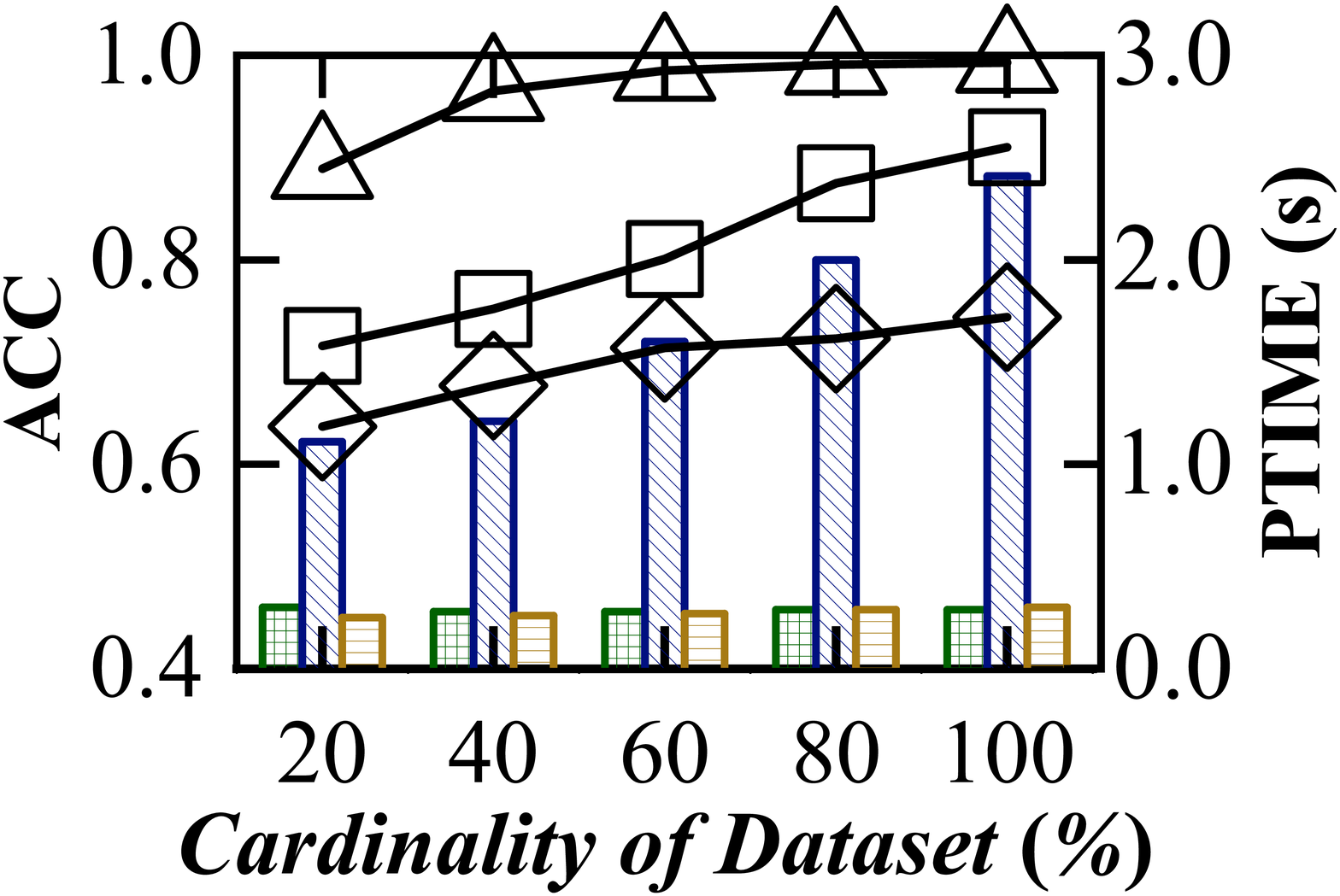}}
	\subfloat[Walk/F1/TTIME]{
		\includegraphics[width=0.24\textwidth]{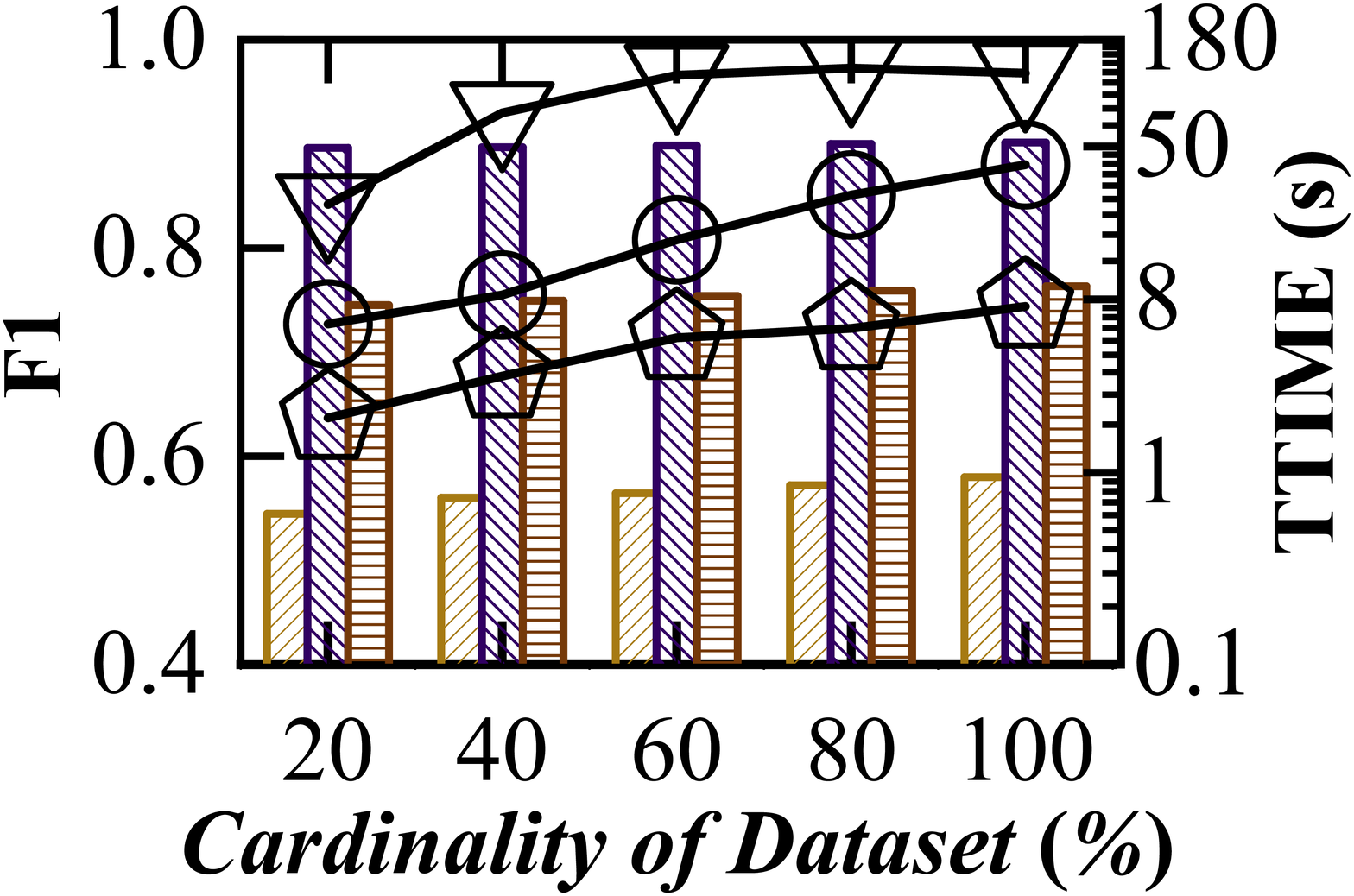}}
	\subfloat[Bike/ACC/PTIME]{
		\includegraphics[width=0.24\textwidth]{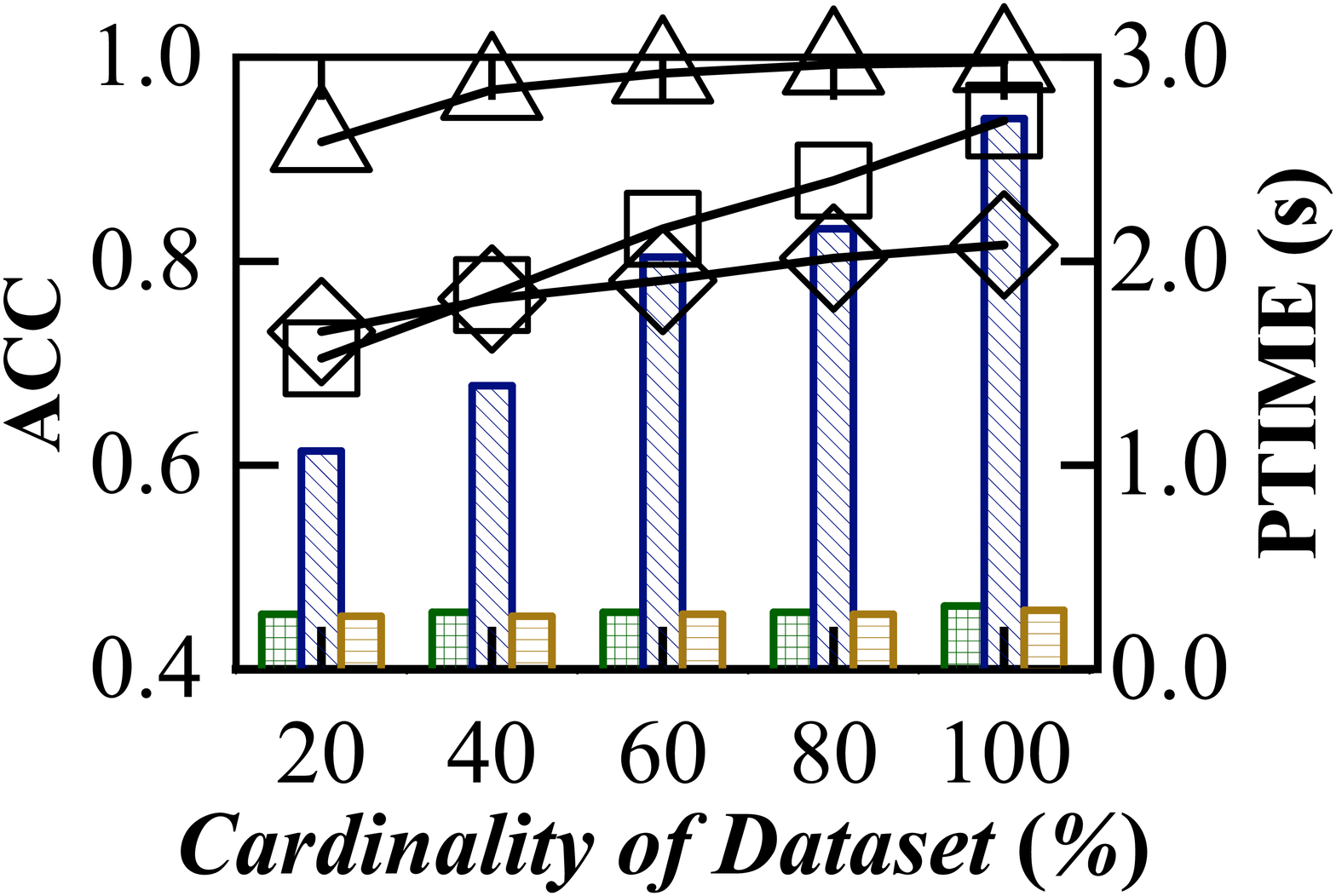}}
	\subfloat[Bike/F1/TTIME]{
		\includegraphics[width=0.24\textwidth]{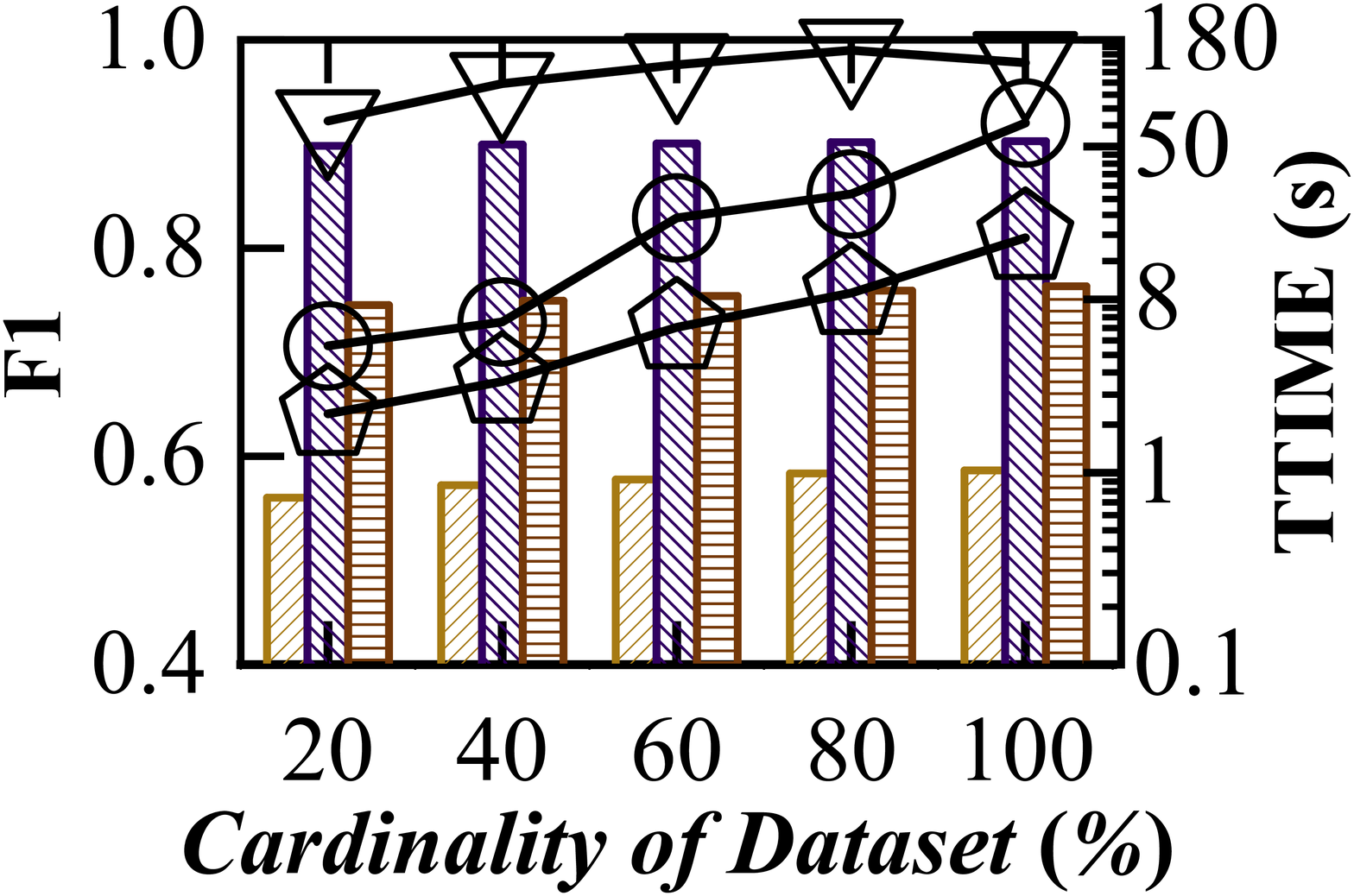}}\\
	\vspace{-2.5mm}
	\centering
	\hspace{-0.25cm}
	\subfloat[Bus/ACC/PTIME]{
		\includegraphics[width=0.24\textwidth]{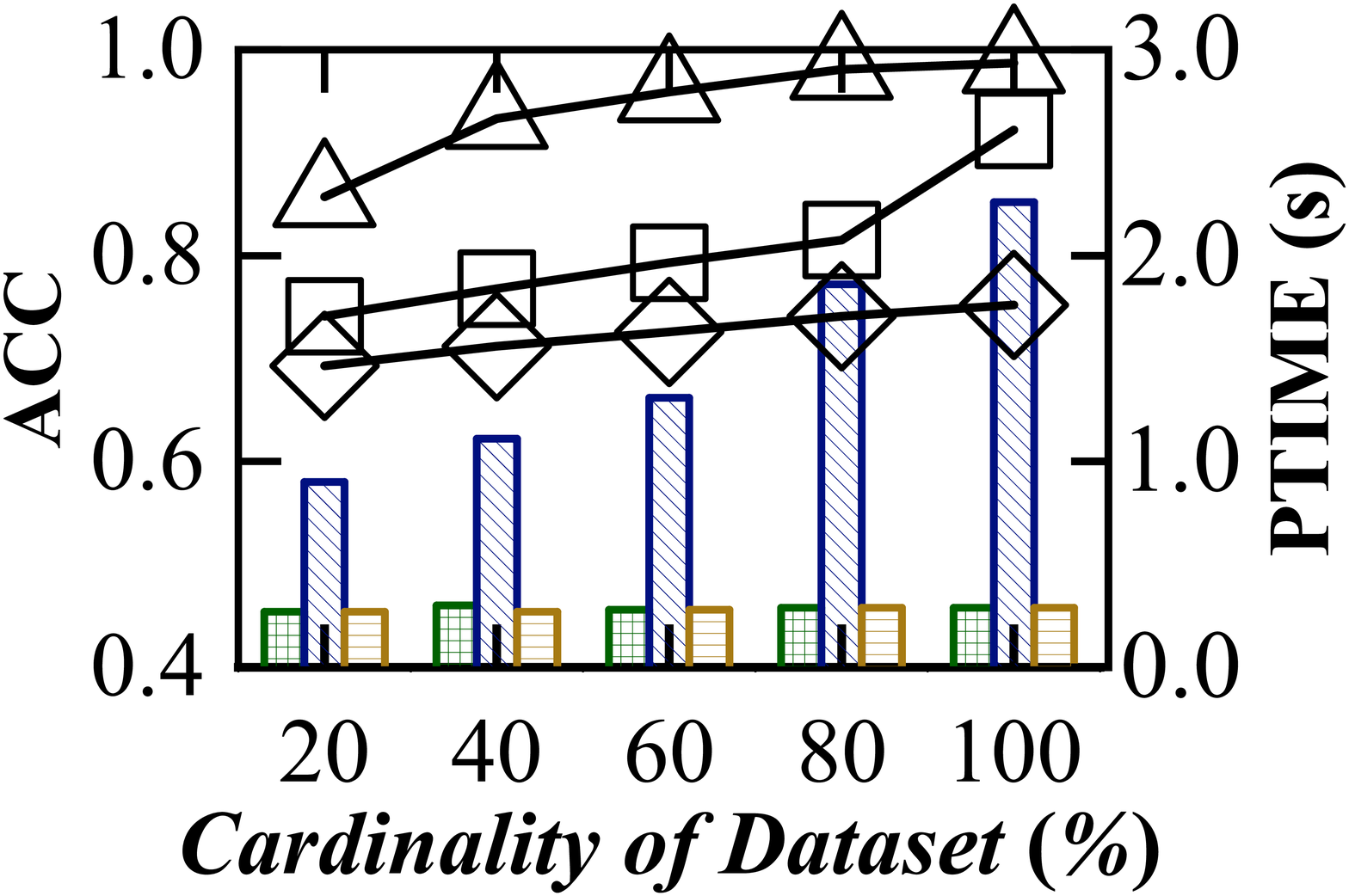}}
	\subfloat[Bus/F1/TTIME]{
		\includegraphics[width=0.24\textwidth]{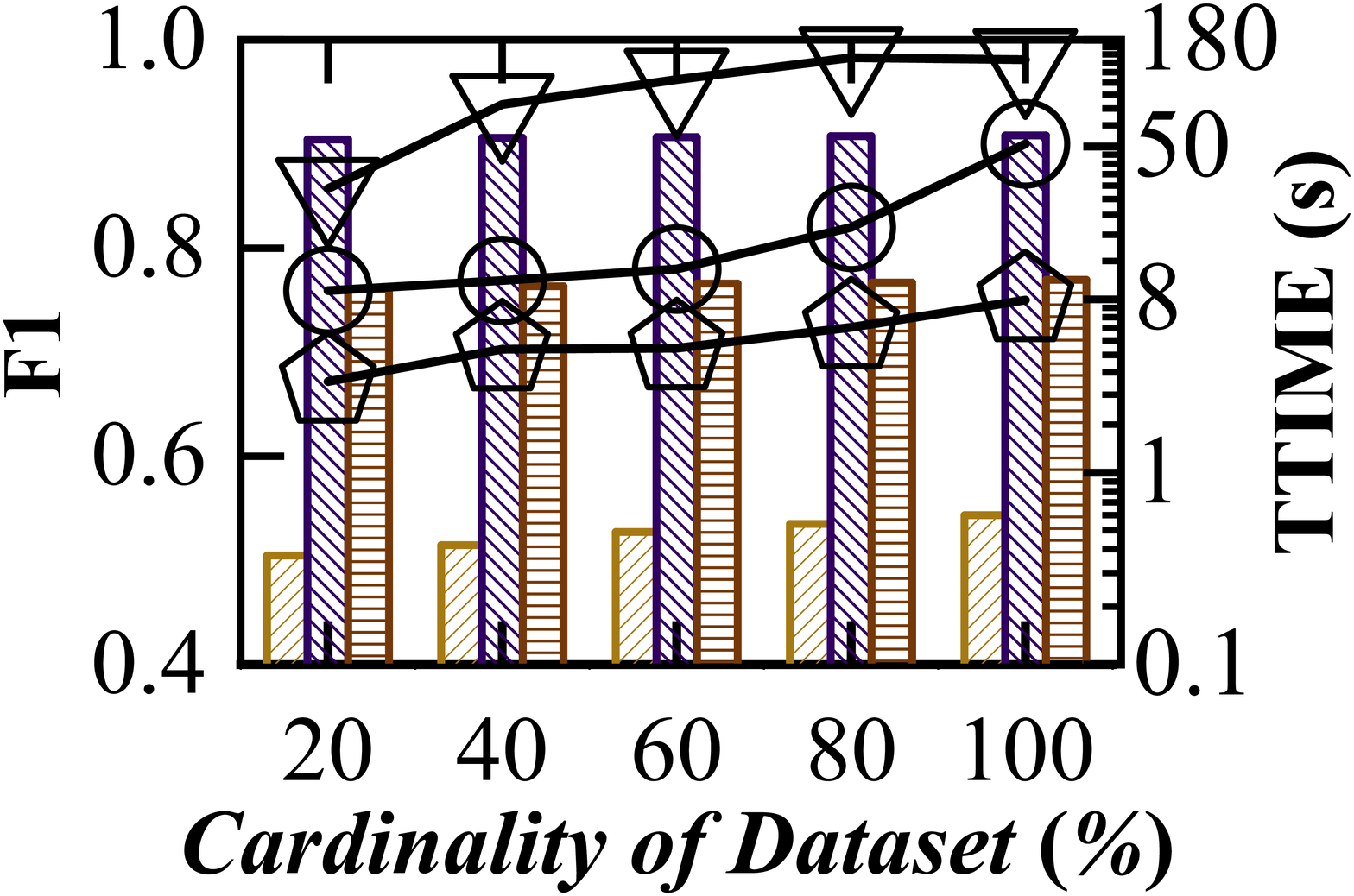}}
	\subfloat[Subway/ACC/PTIME]{
		\includegraphics[width=0.24\textwidth]{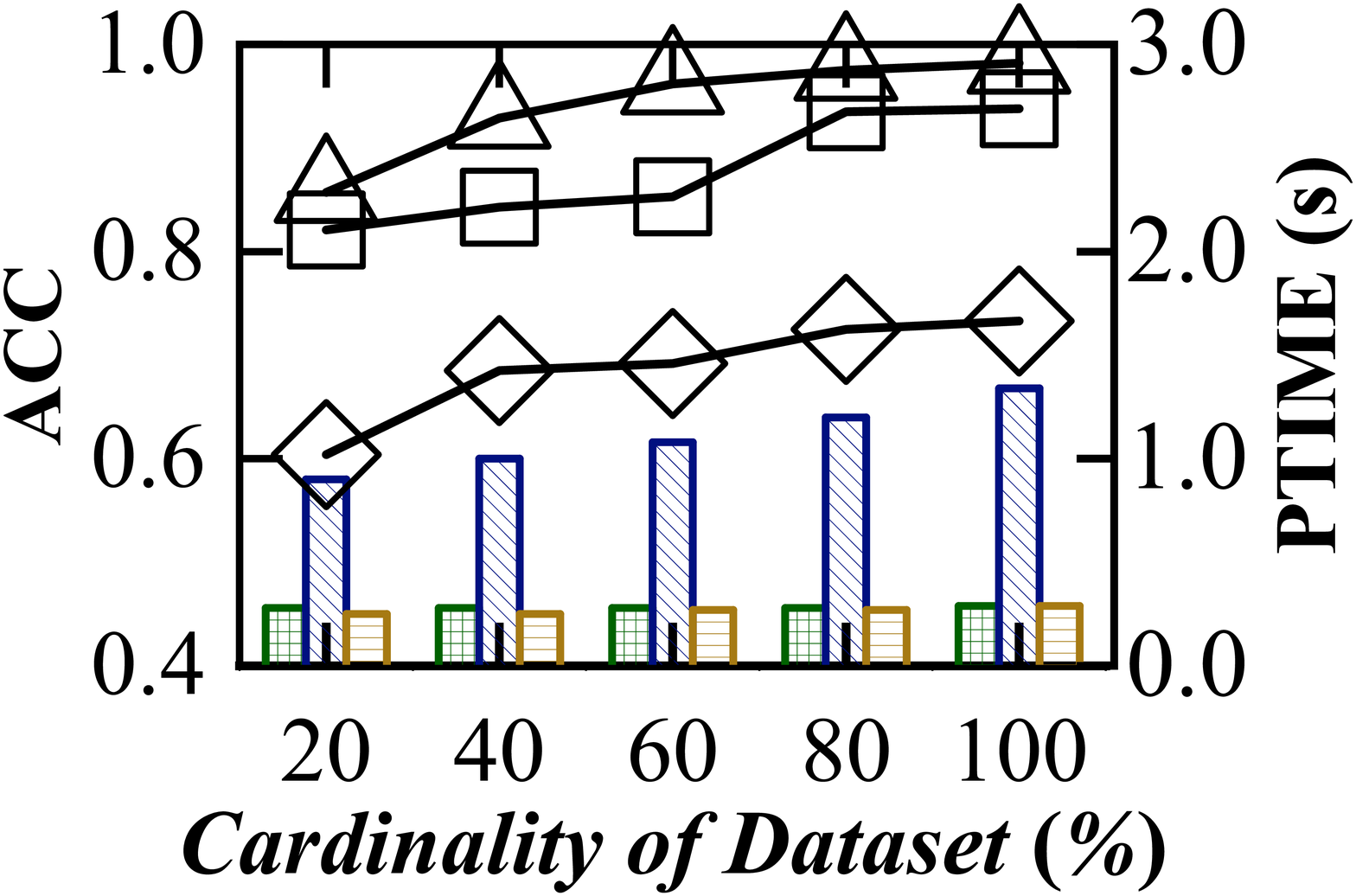}}
	\subfloat[Subway/F1/TTIME]{
		\includegraphics[width=0.24\textwidth]{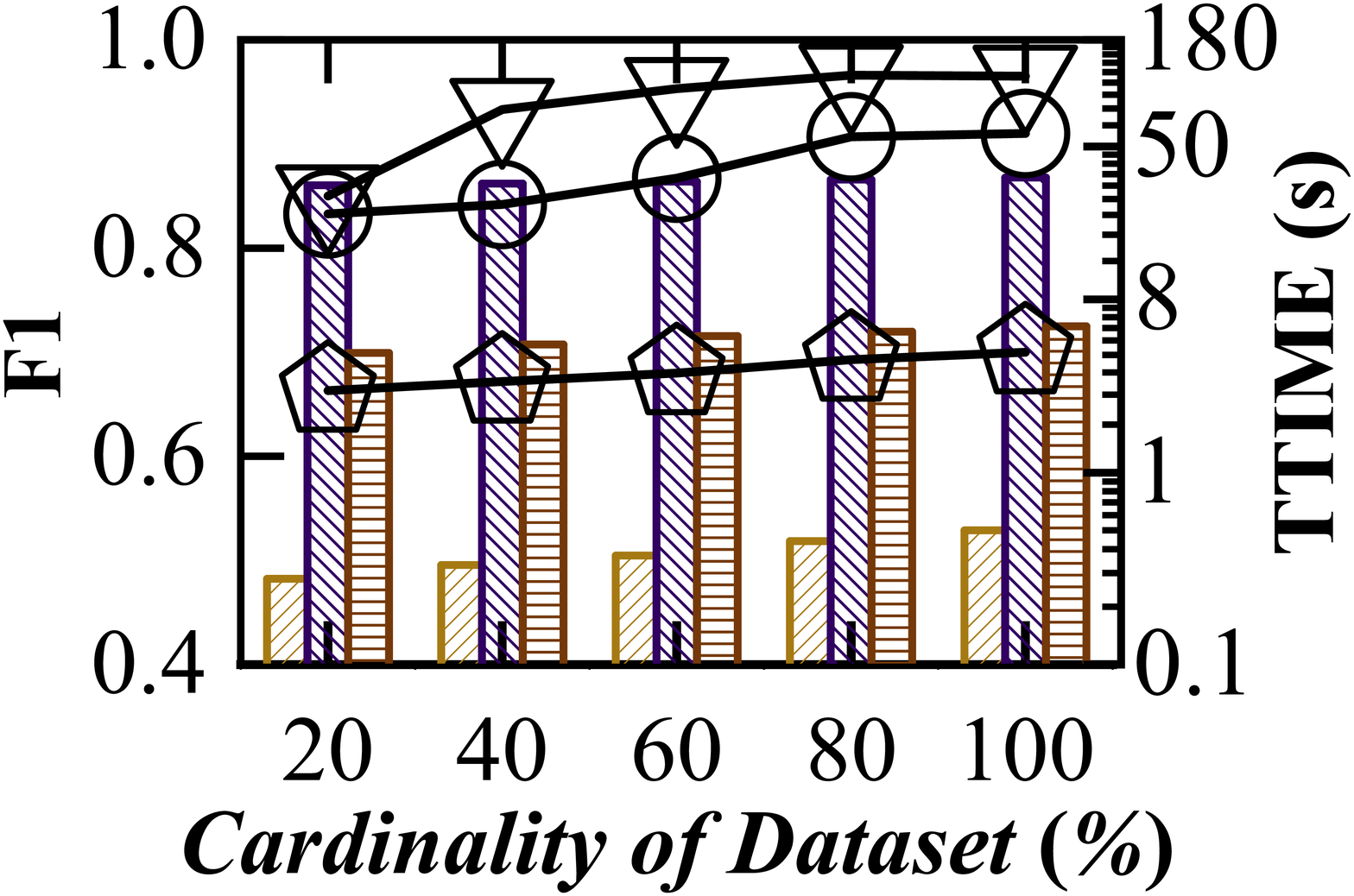}}\\
	\vspace{-2.5mm}
	\centering
	\hspace{-0.25cm}
	\subfloat[Private Car/ACC/PTIME]{
		\includegraphics[width=0.24\textwidth]{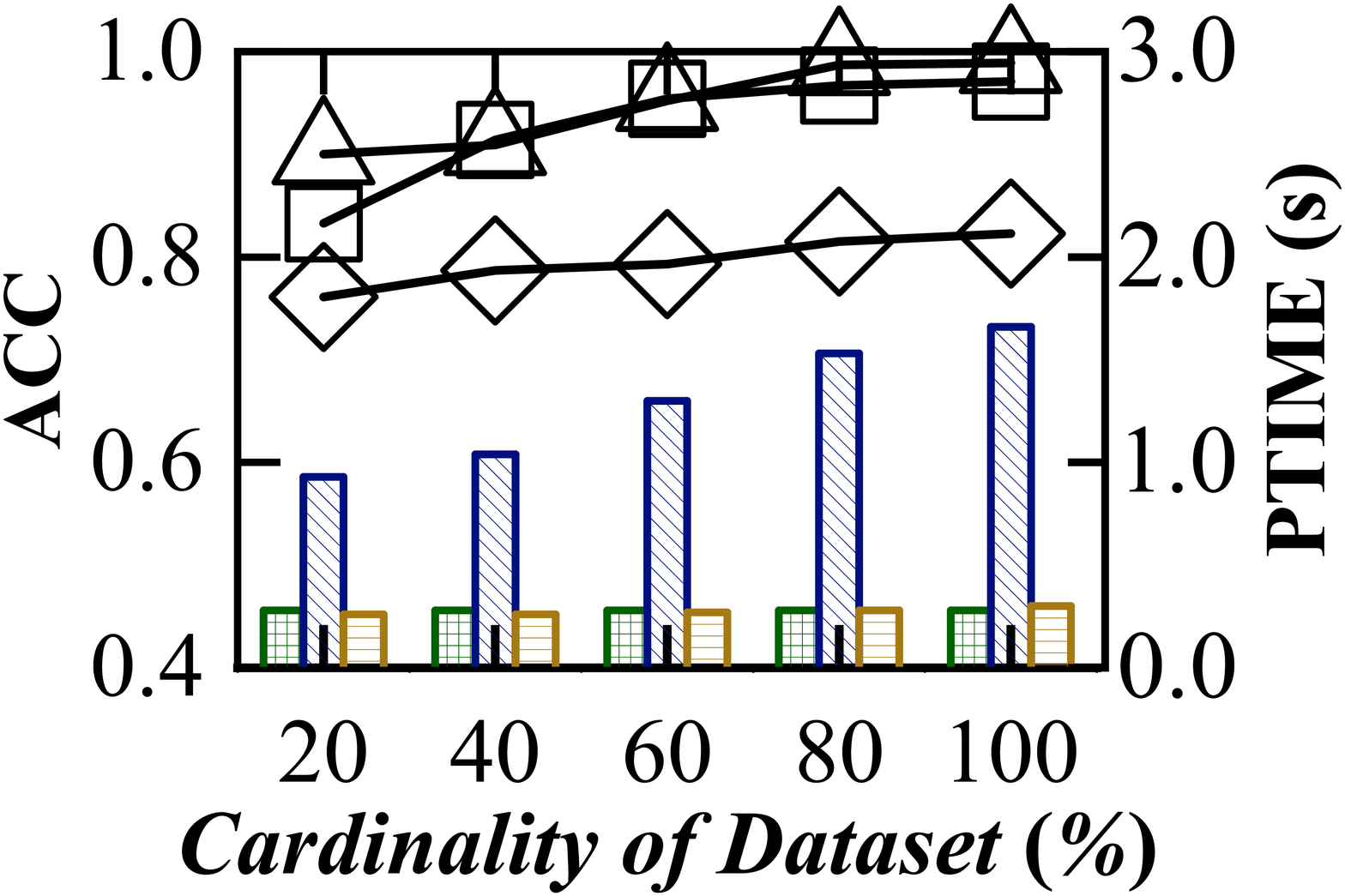}}
	\subfloat[Private Car/F1/TTIME]{
		\includegraphics[width=0.24\textwidth]{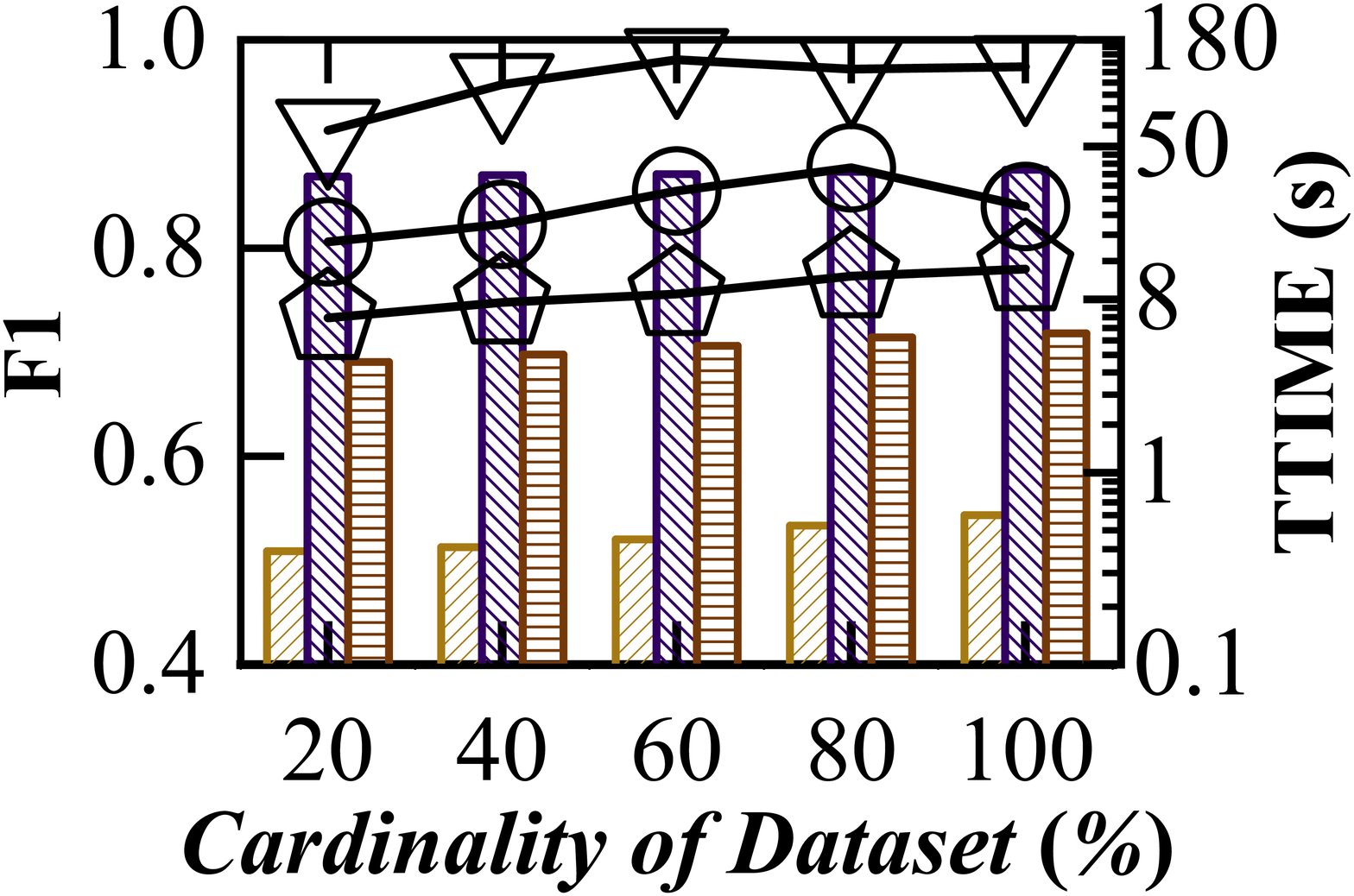}}
	\subfloat[Taxi/ACC/PTIME]{
		\includegraphics[width=0.24\textwidth]{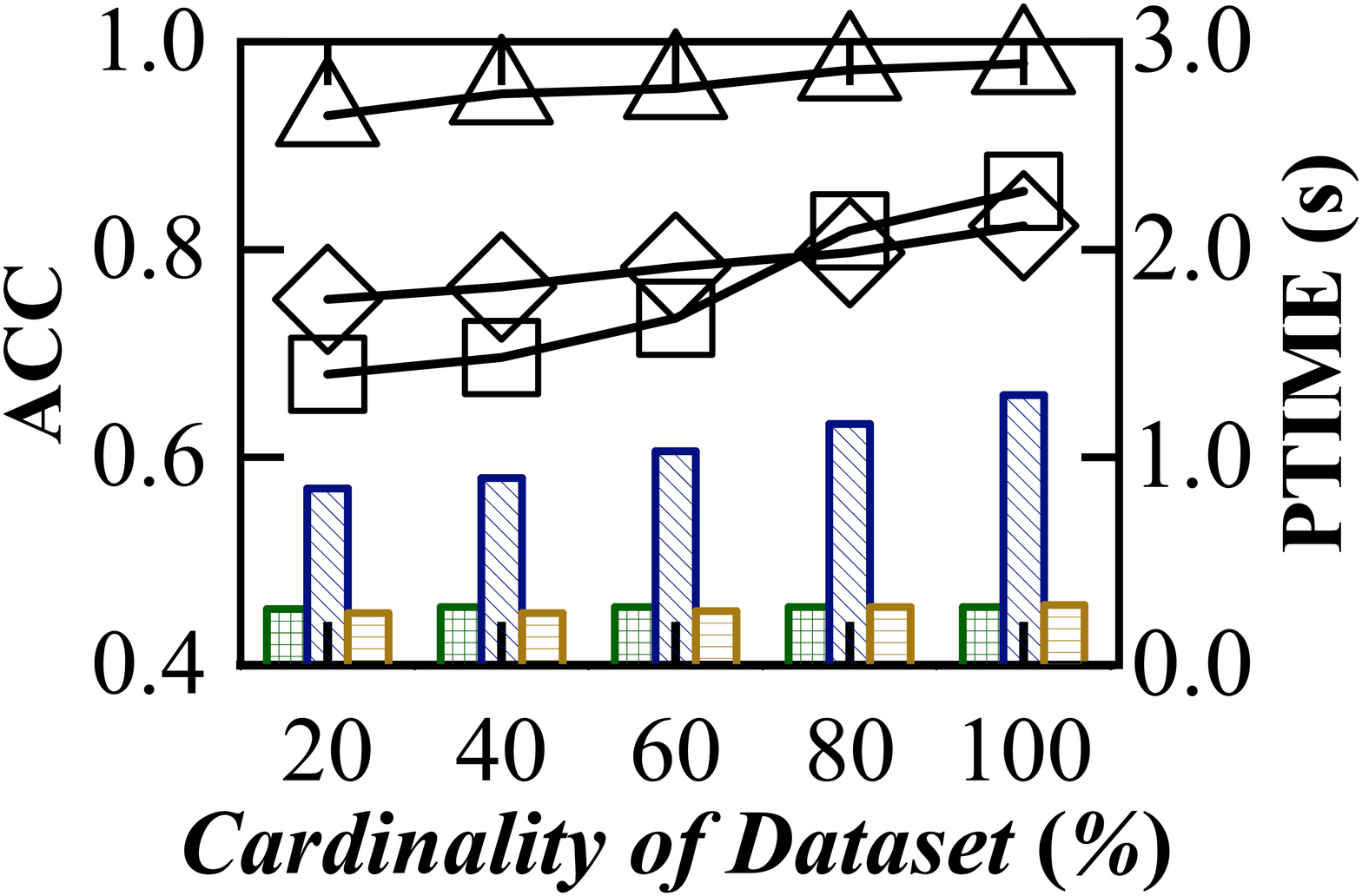}}
	\subfloat[Taxi/F1/TTIME]{
		\includegraphics[width=0.24\textwidth]{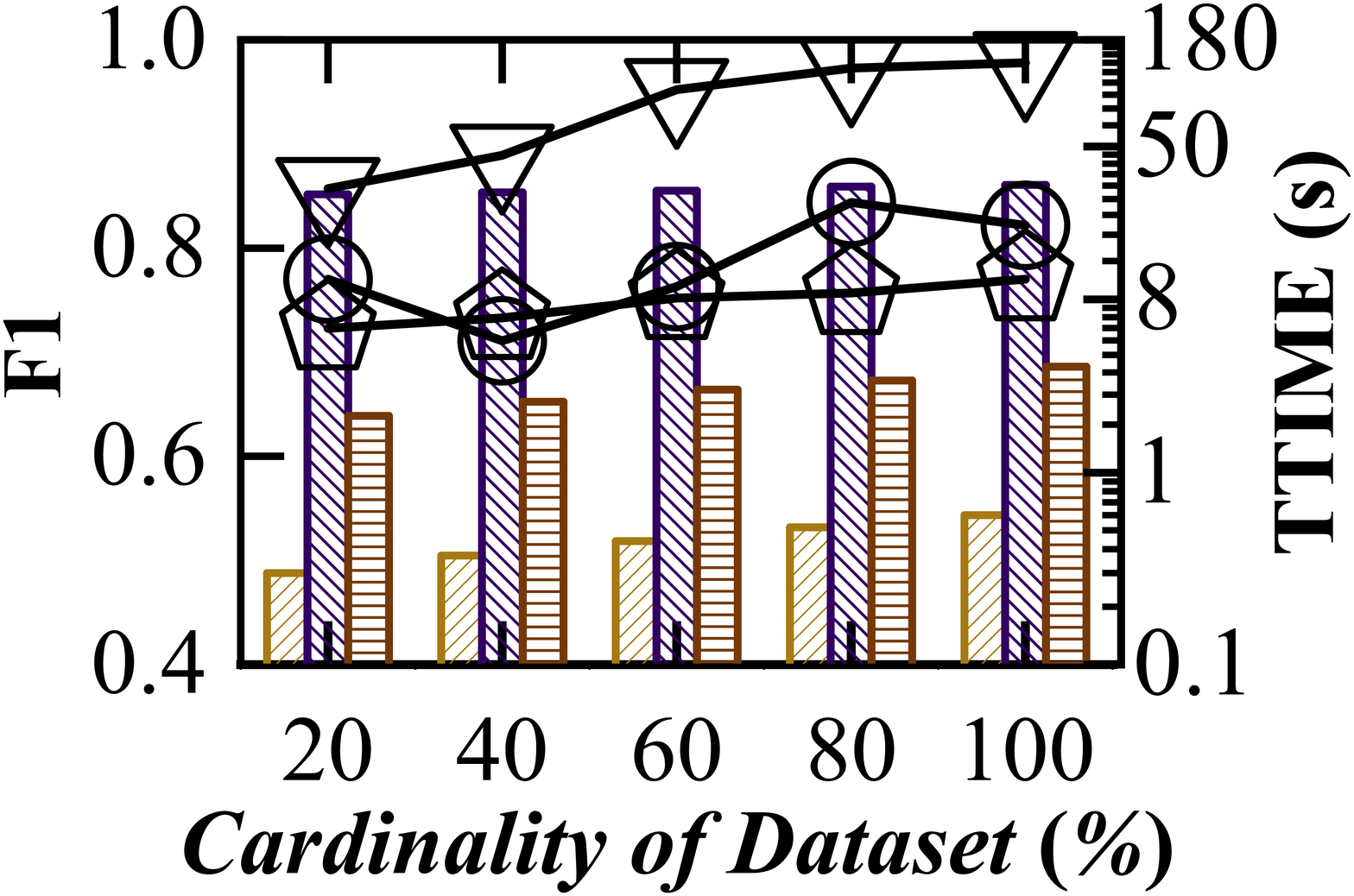}}\\
	\vspace{-2.5mm}
	\centering
	\hspace{-0.25cm}
	\subfloat[Train/ACC/PTIME]{
		\includegraphics[width=0.24\textwidth]{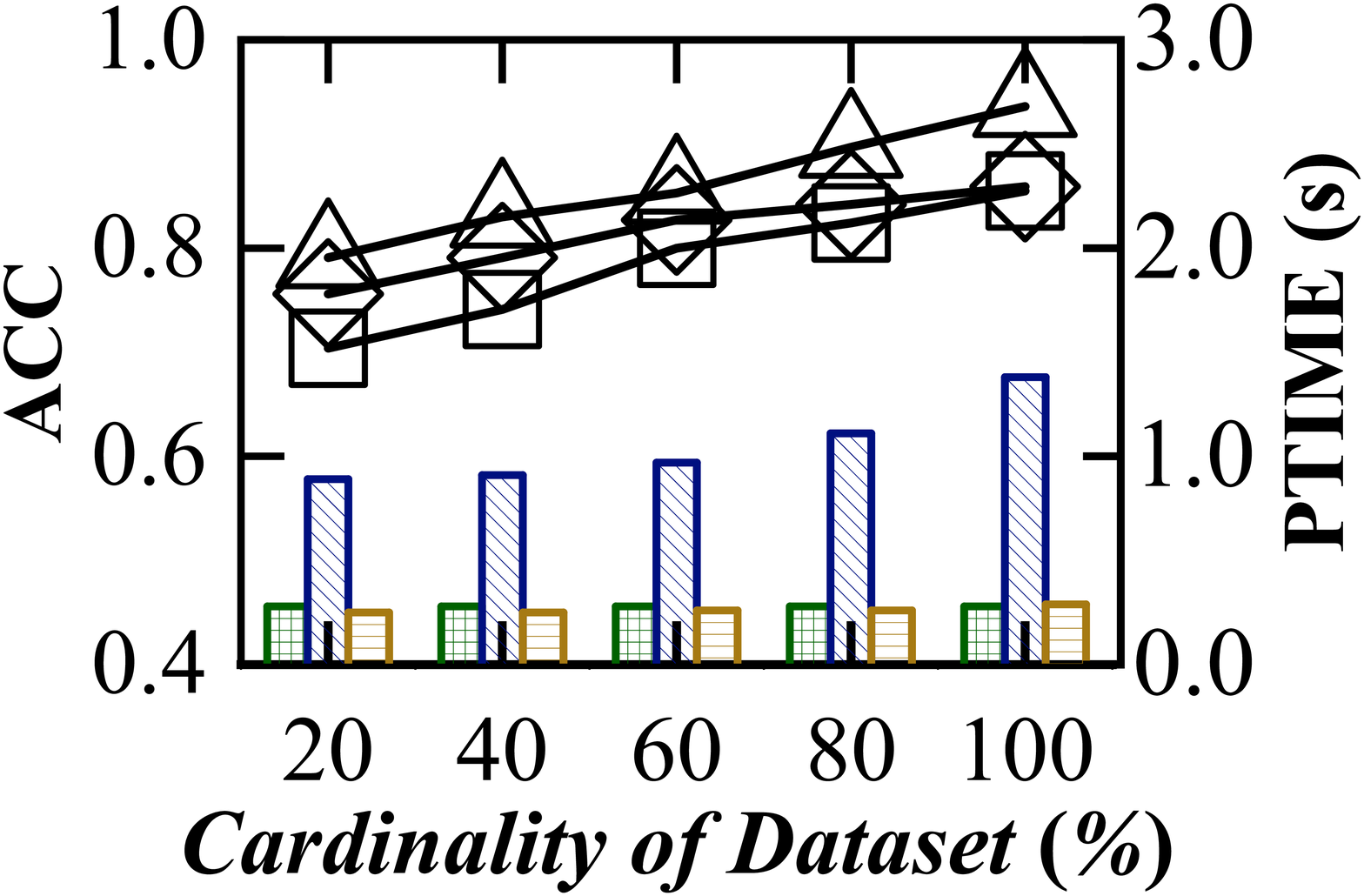}}
	\subfloat[Train/F1/TTIME]{
		\includegraphics[width=0.24\textwidth]{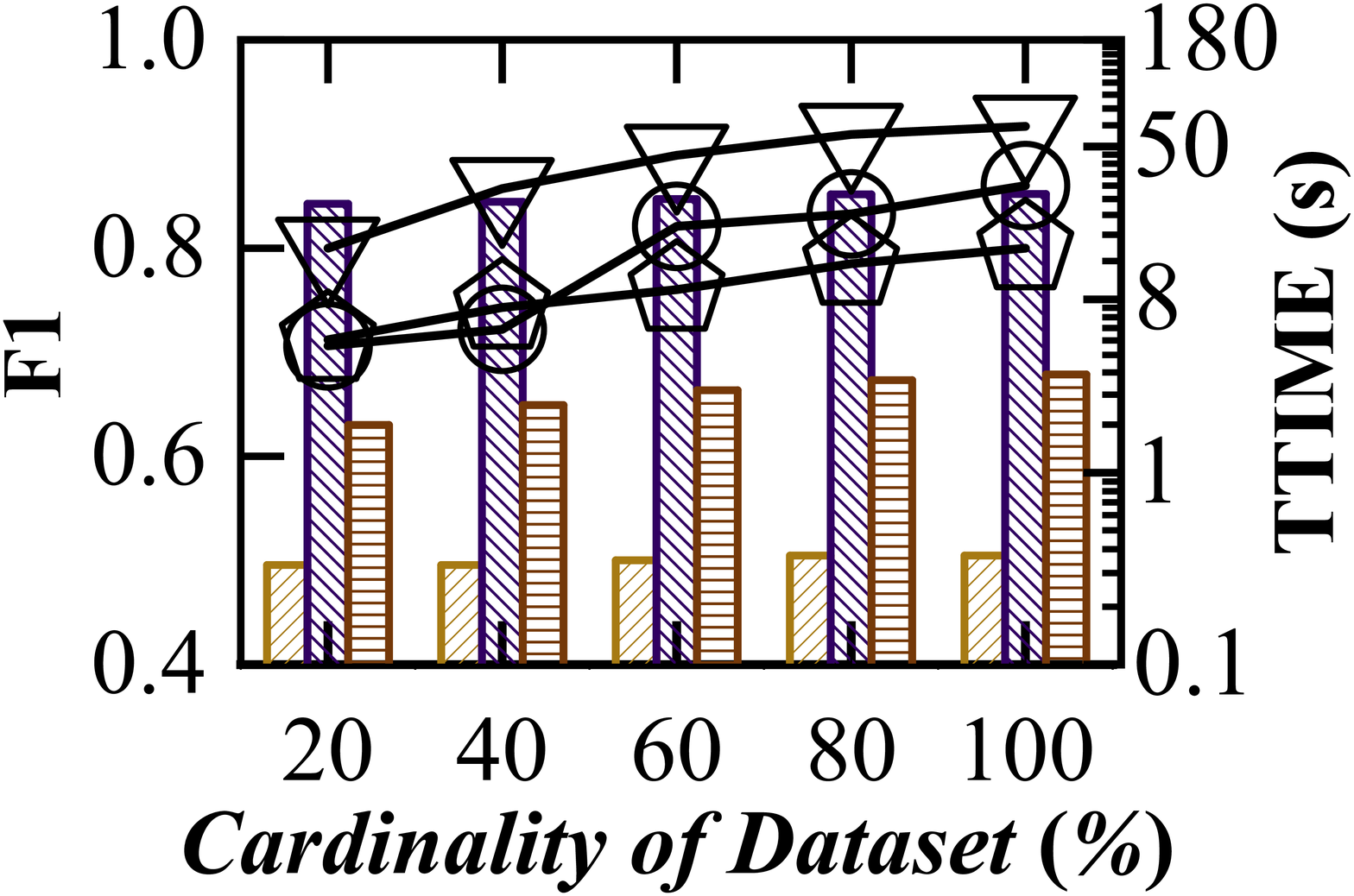}}
	\subfloat[GeoLife/ACC/PTIME]{
		\includegraphics[width=0.24\textwidth]{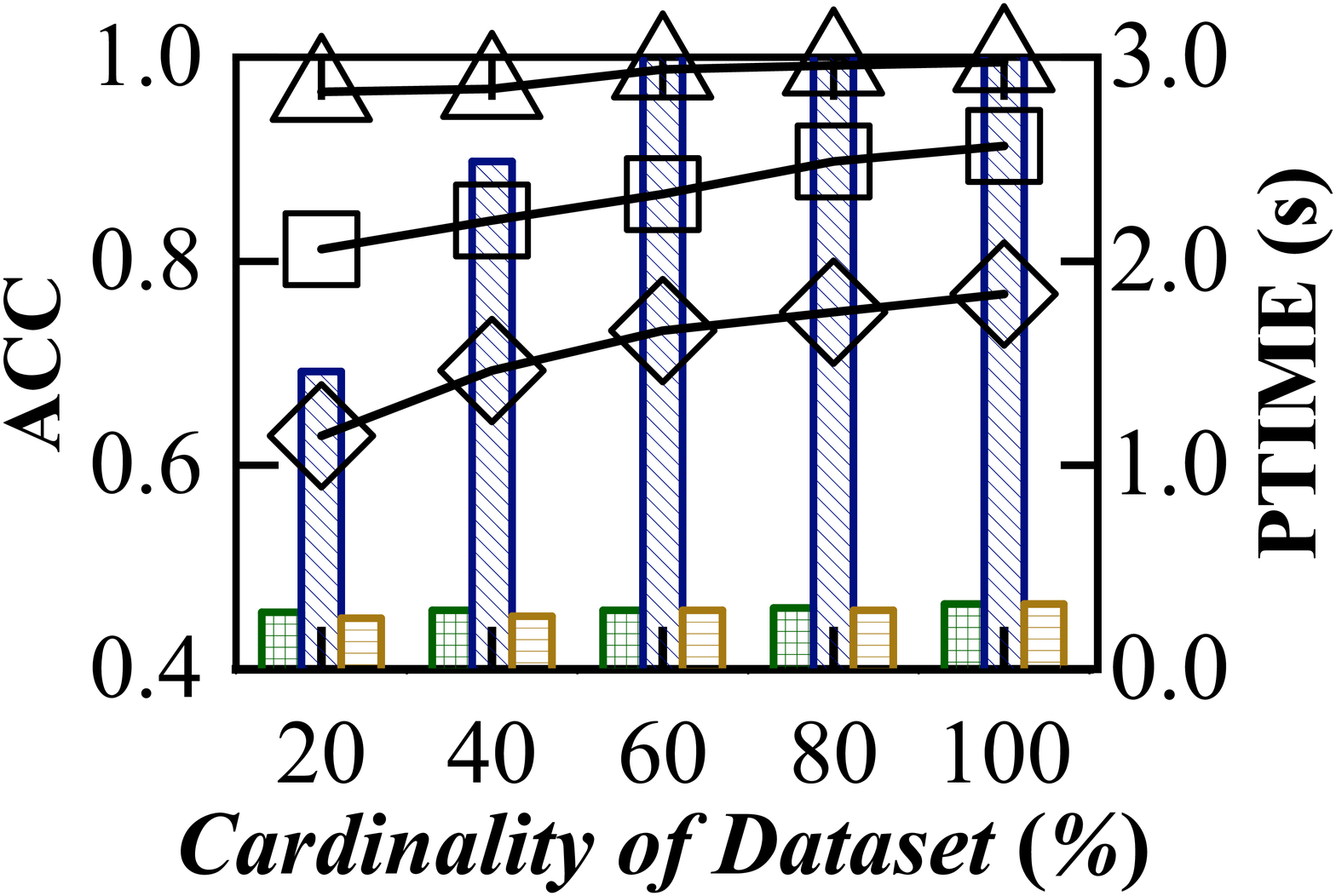}}
	\subfloat[Geolife/F1/TTIME]{
		\includegraphics[width=0.24\textwidth]{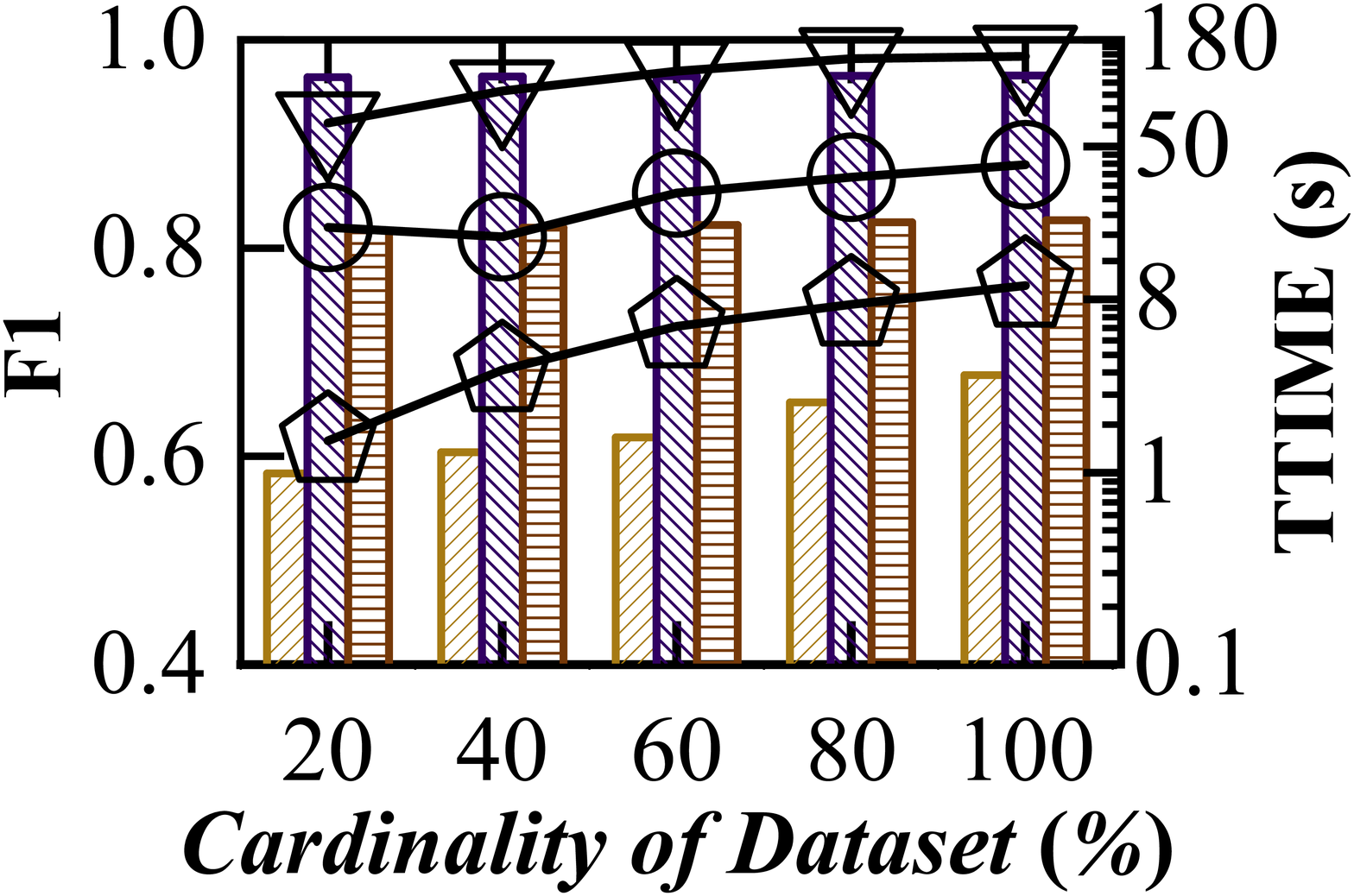}}
	\caption{Model Scalability Evaluation vs. Cardinality of Datasets}
	\label{fig:scalability}
    \vspace{-5mm}
\end{figure*}

\begin{figure*} [tb]
	\centering
	\vspace{-4mm}
	\hspace{0.7mm}
	\includegraphics[width=0.98\textwidth]{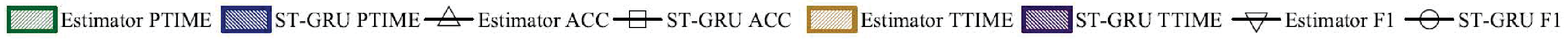}
    \vspace{-1mm}
	\hspace{-0.25cm}
	\subfloat[Walk/ACC/PTIME]{
		\includegraphics[width=0.24\textwidth]{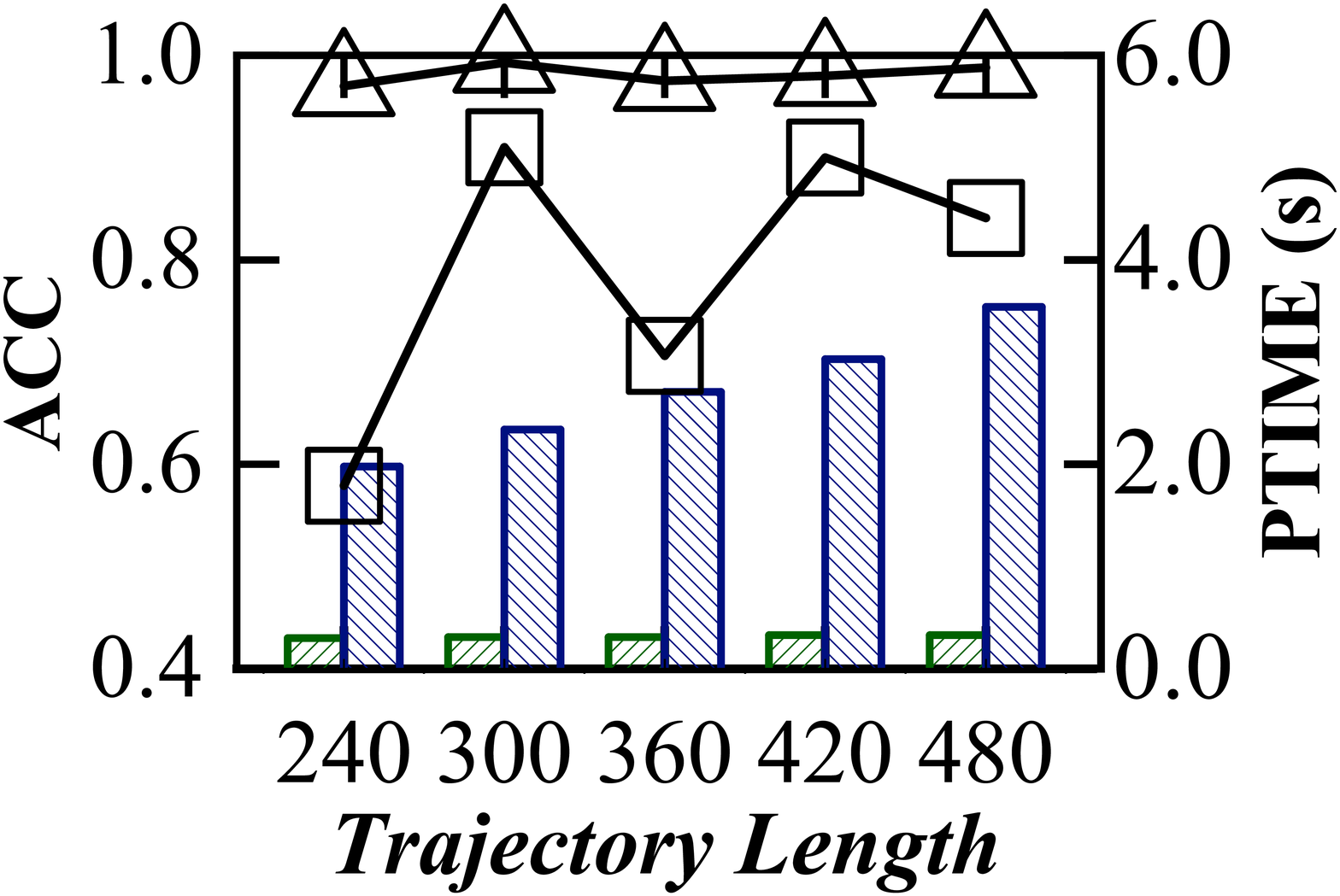}}
	\subfloat[Walk/F1/TTIME]{
		\includegraphics[width=0.24\textwidth]{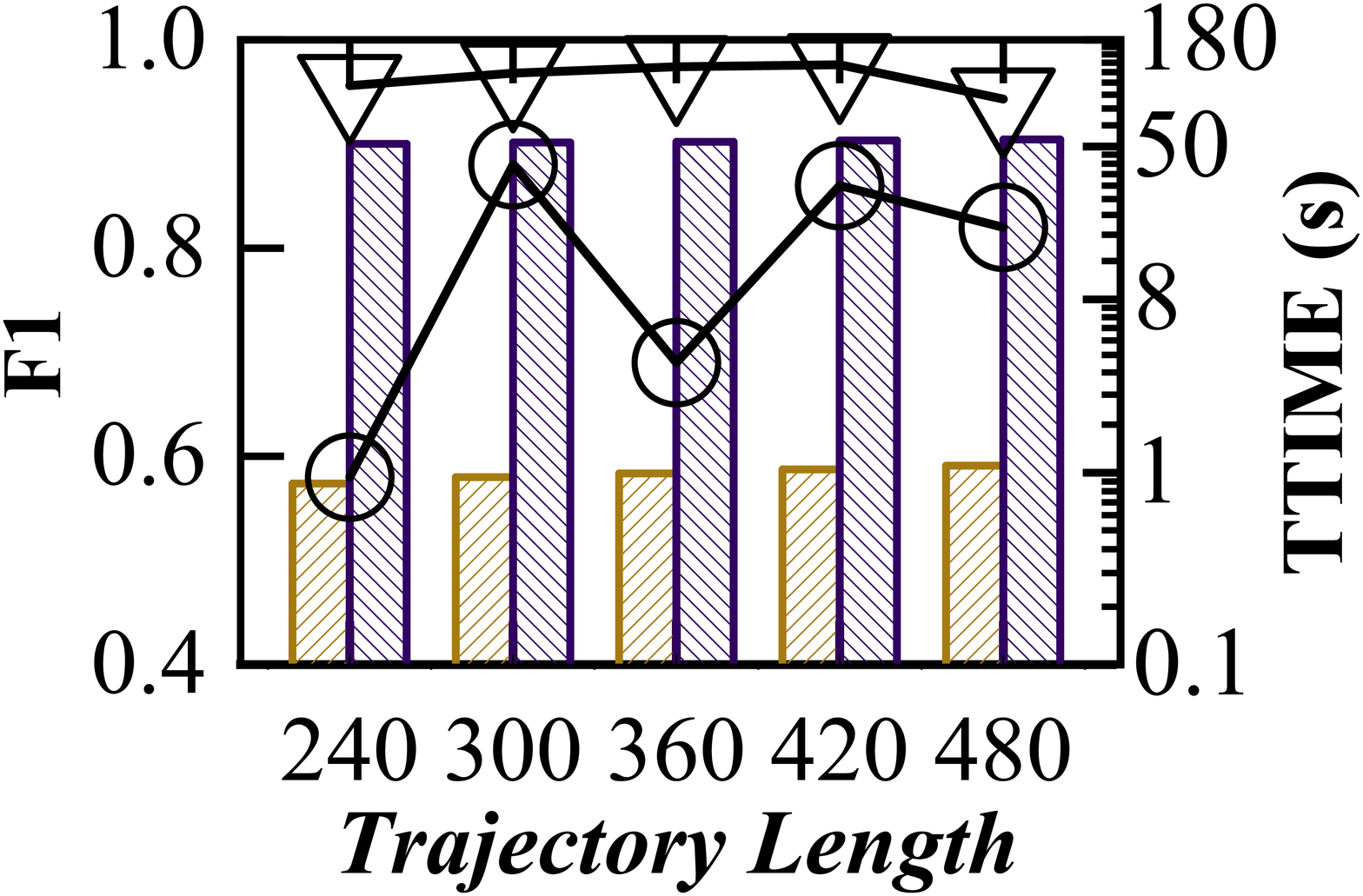}}
	\subfloat[Bike/ACC/PTIME]{
		\includegraphics[width=0.24\textwidth]{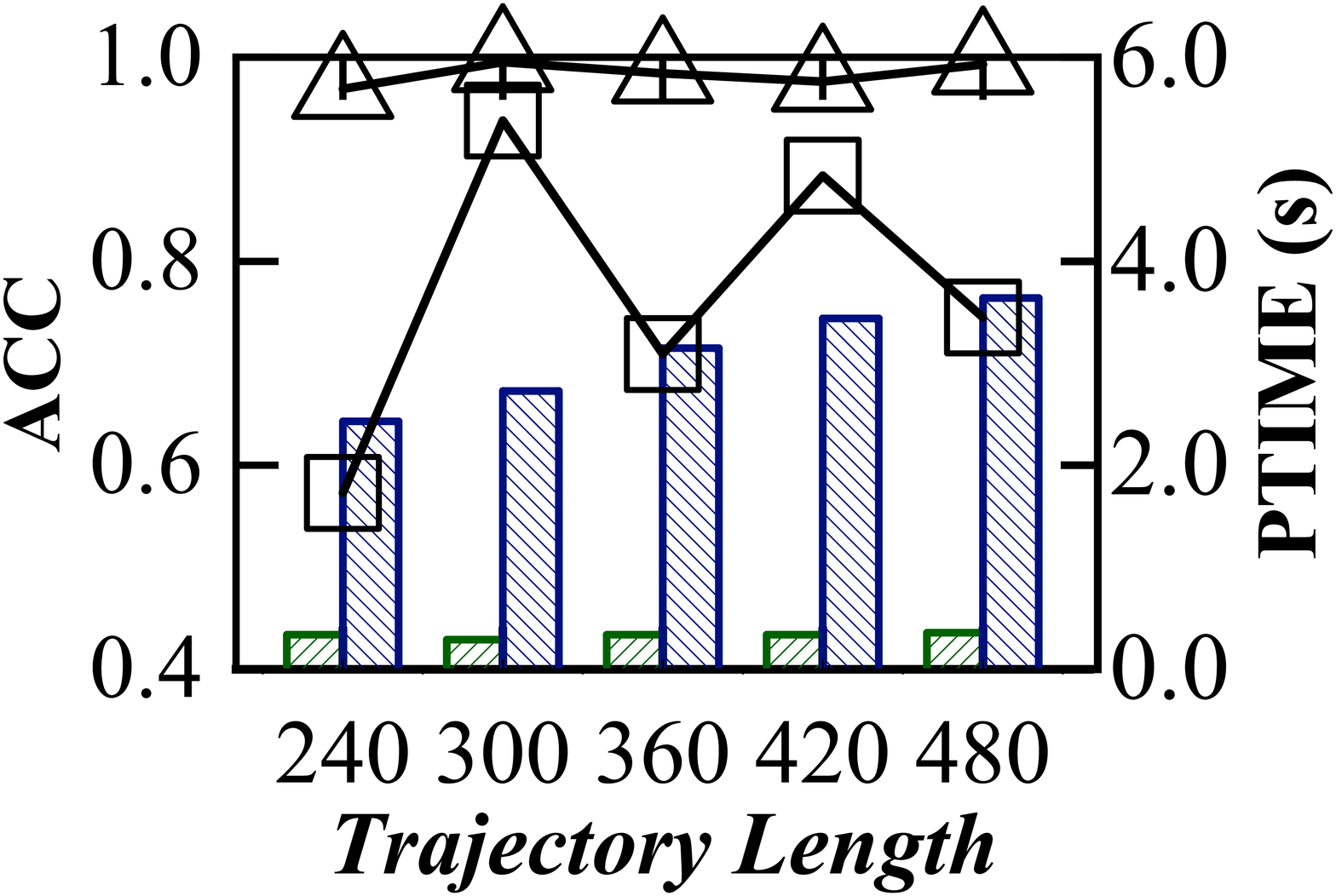}}
	\subfloat[Bike/F1/TTIME]{
		\includegraphics[width=0.24\textwidth]{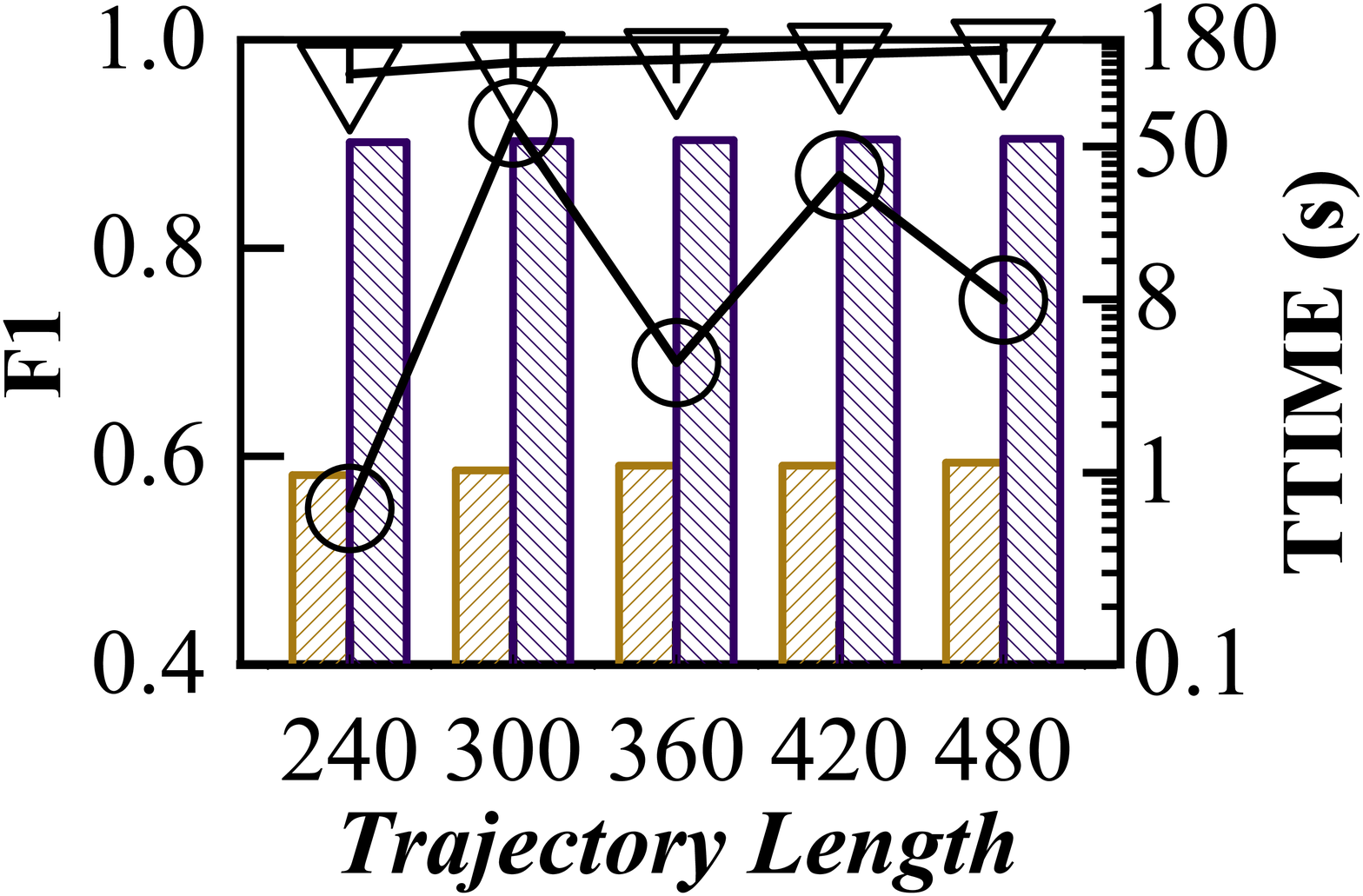}}\\
	\vspace{-2.5mm}
	\centering
	\hspace{-0.25cm}
	\subfloat[Bus/ACC/PTIME]{
		\includegraphics[width=0.24\textwidth]{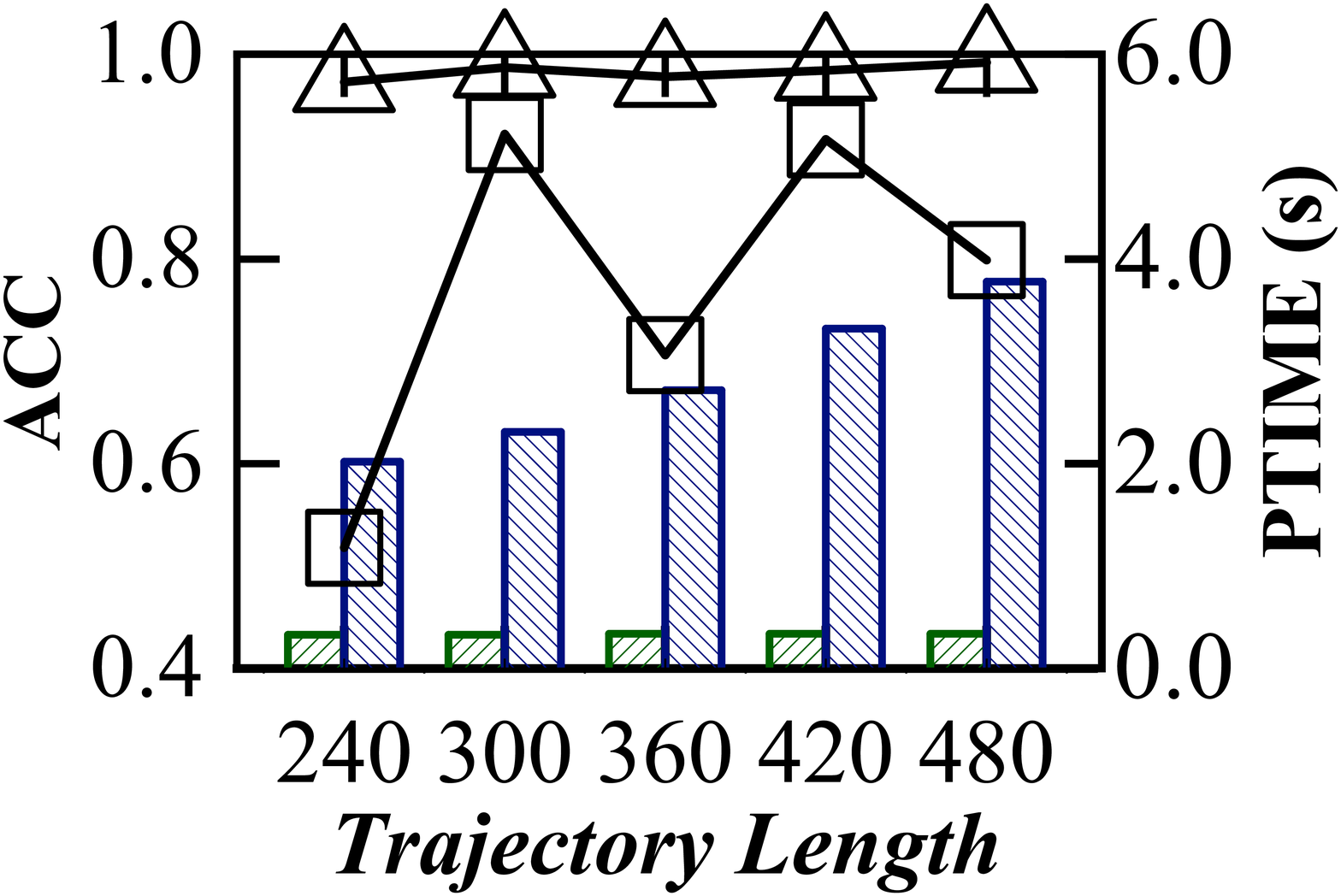}}
	\subfloat[Bus/F1/TTIME]{
		\includegraphics[width=0.24\textwidth]{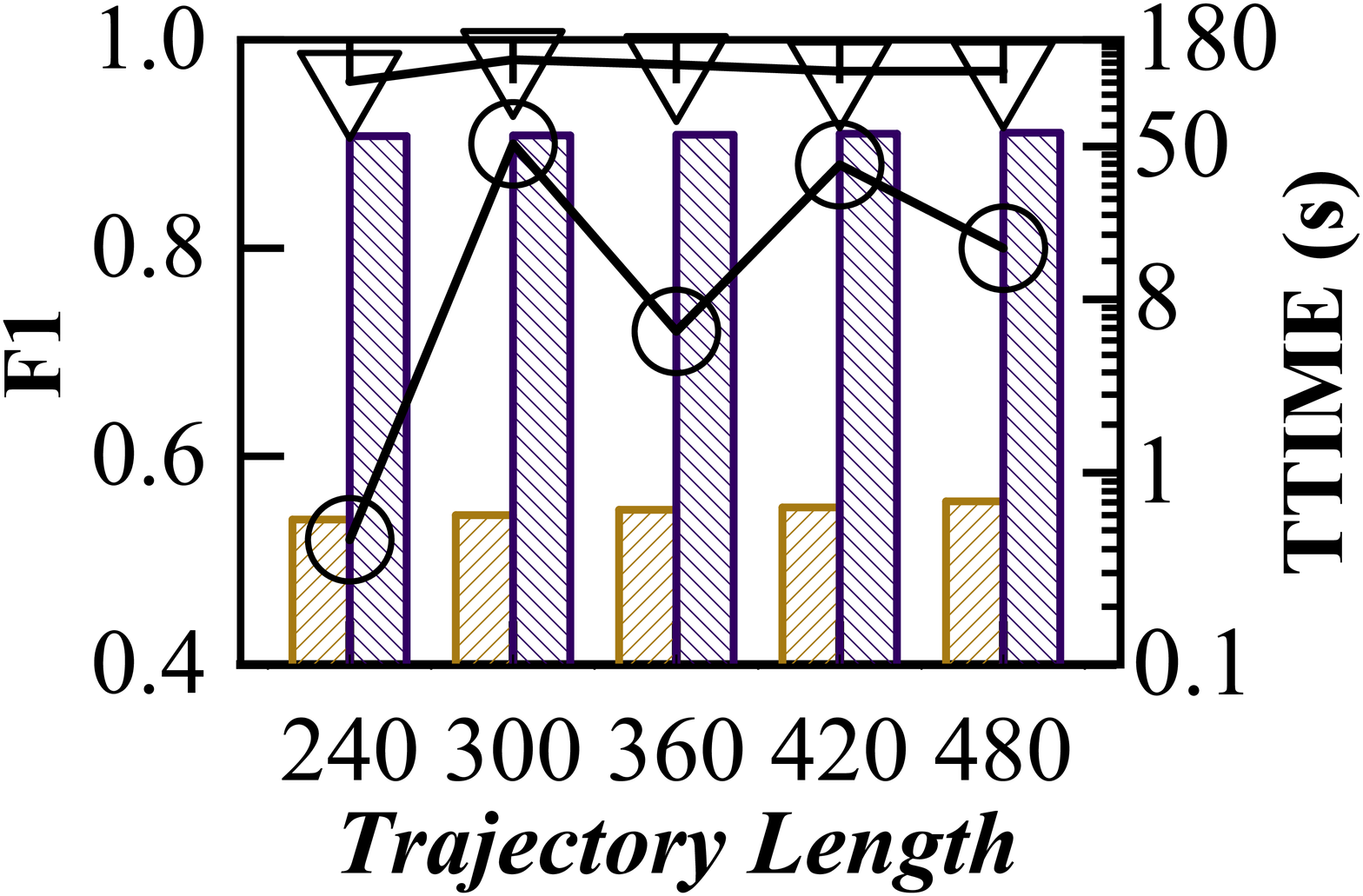}}
	\subfloat[Subway/ACC/PTIME]{
		\includegraphics[width=0.24\textwidth]{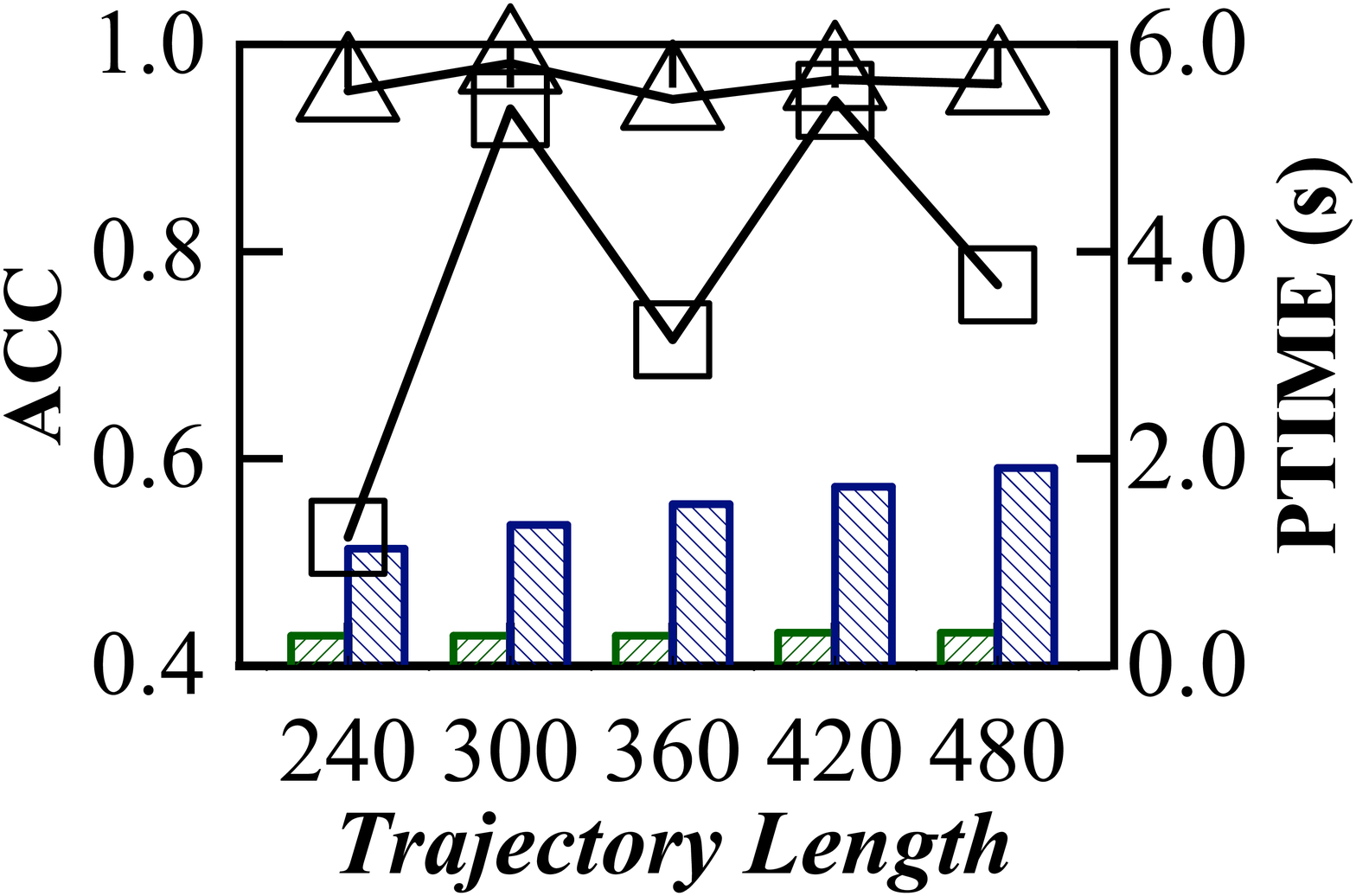}}
	\subfloat[Subway/F1/TTIME]{
		\includegraphics[width=0.24\textwidth]{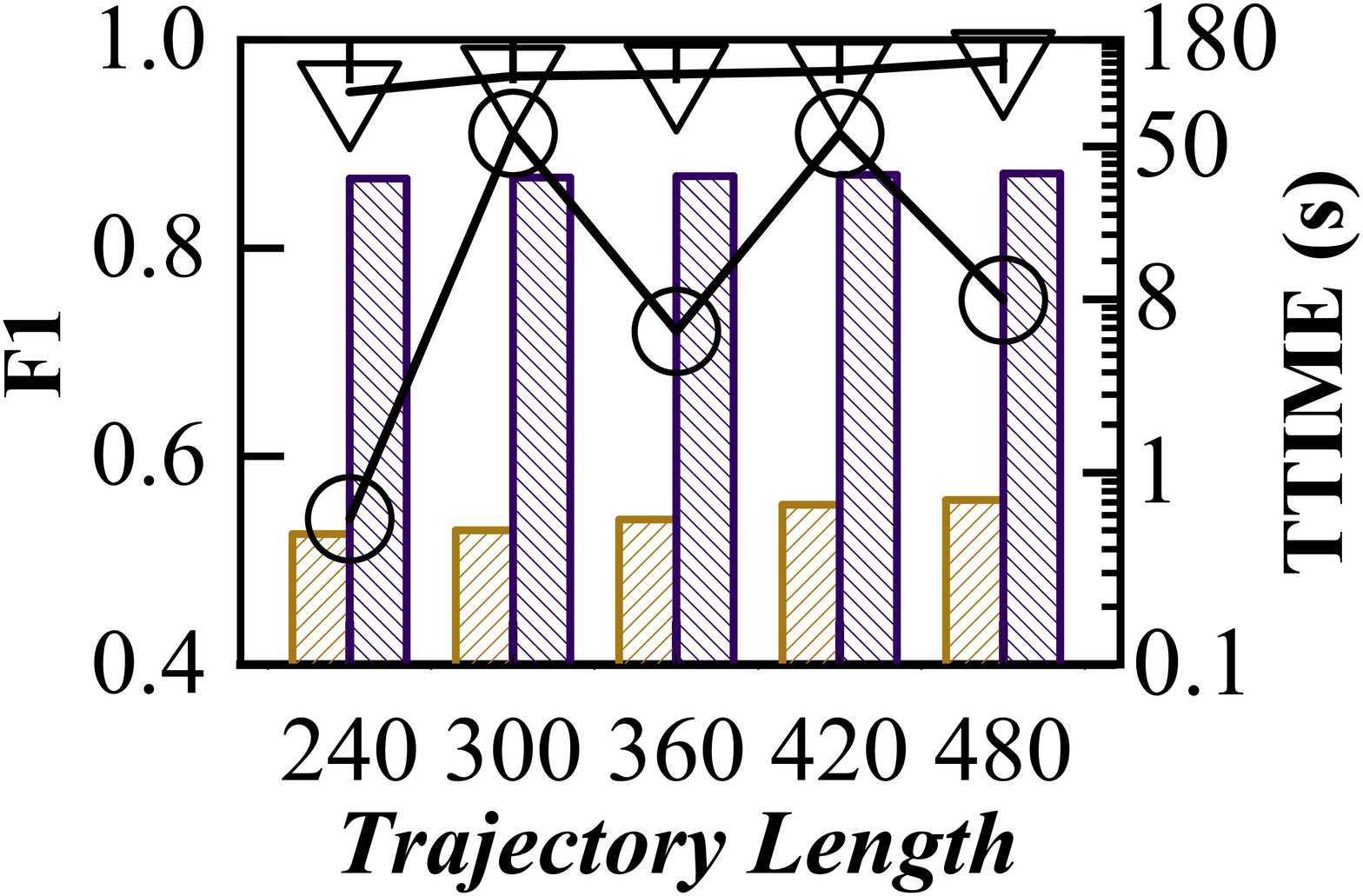}}\\
	\vspace{-2.5mm}
	\centering
	\hspace{-0.25cm}
	\subfloat[Private Car/ACC/PTIME]{
		\includegraphics[width=0.24\textwidth]{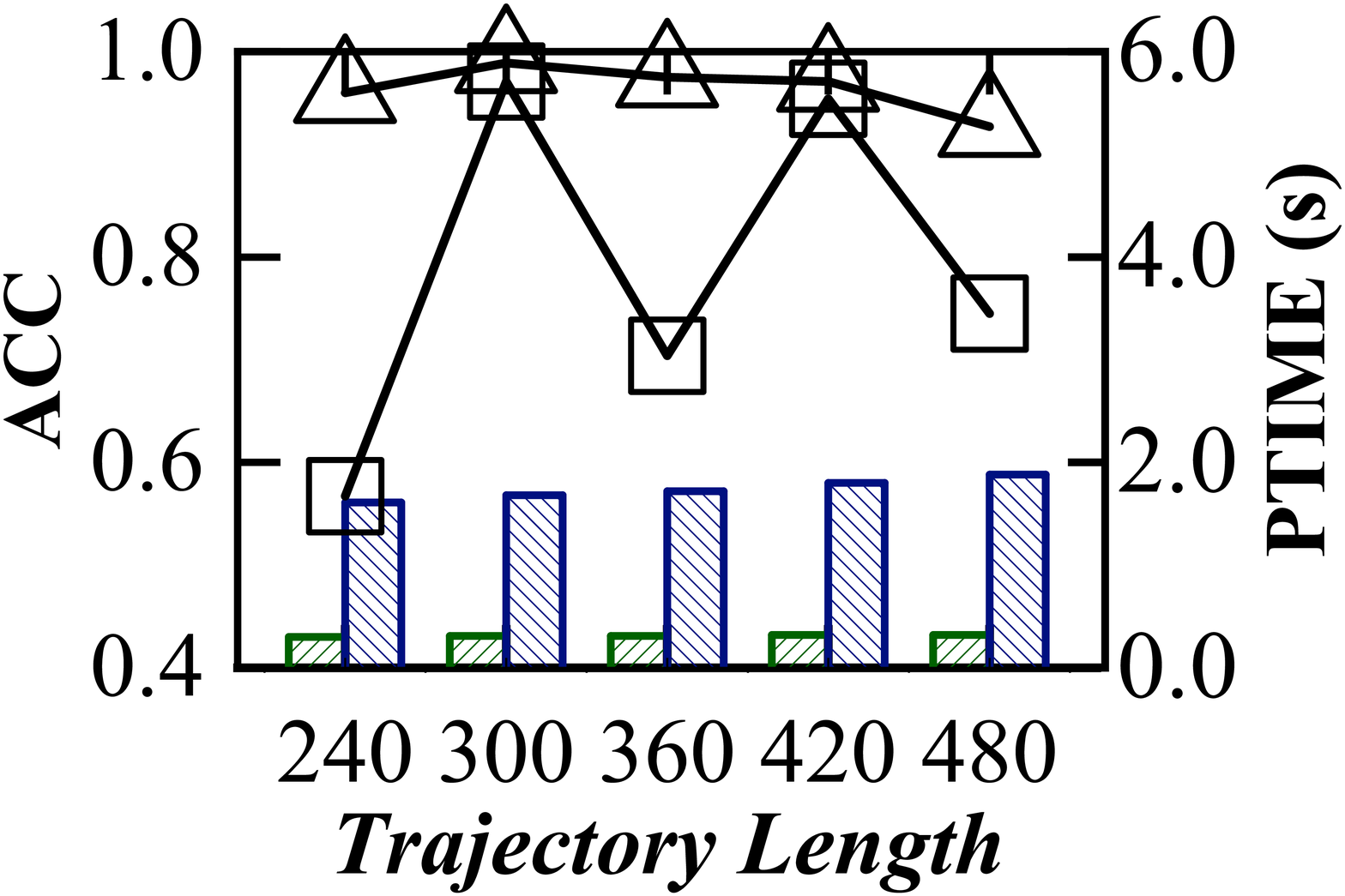}}
	\subfloat[Private Car/F1/TTIME]{
		\includegraphics[width=0.24\textwidth]{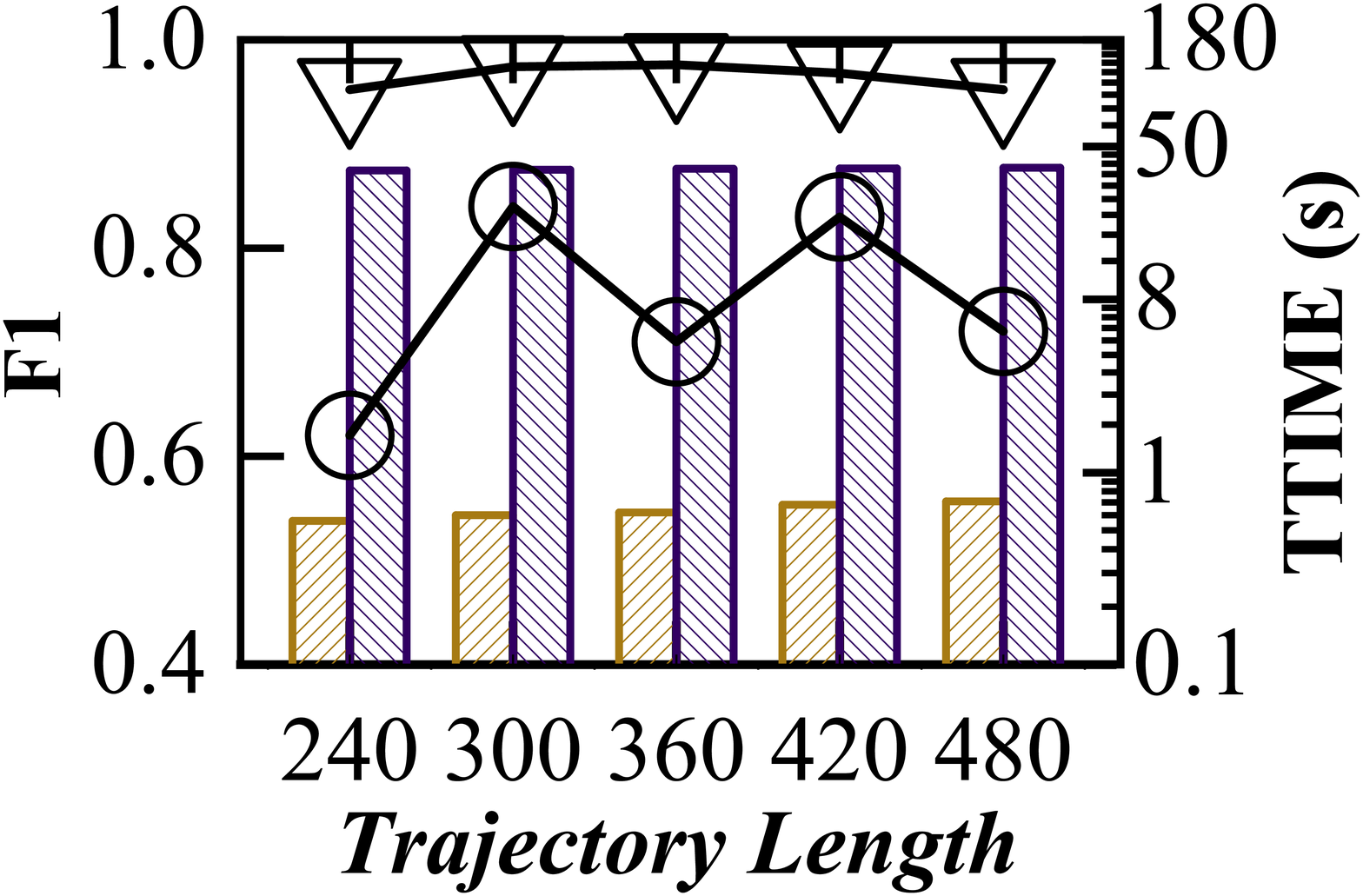}}
	\subfloat[Taxi/ACC/PTIME]{
		\includegraphics[width=0.24\textwidth]{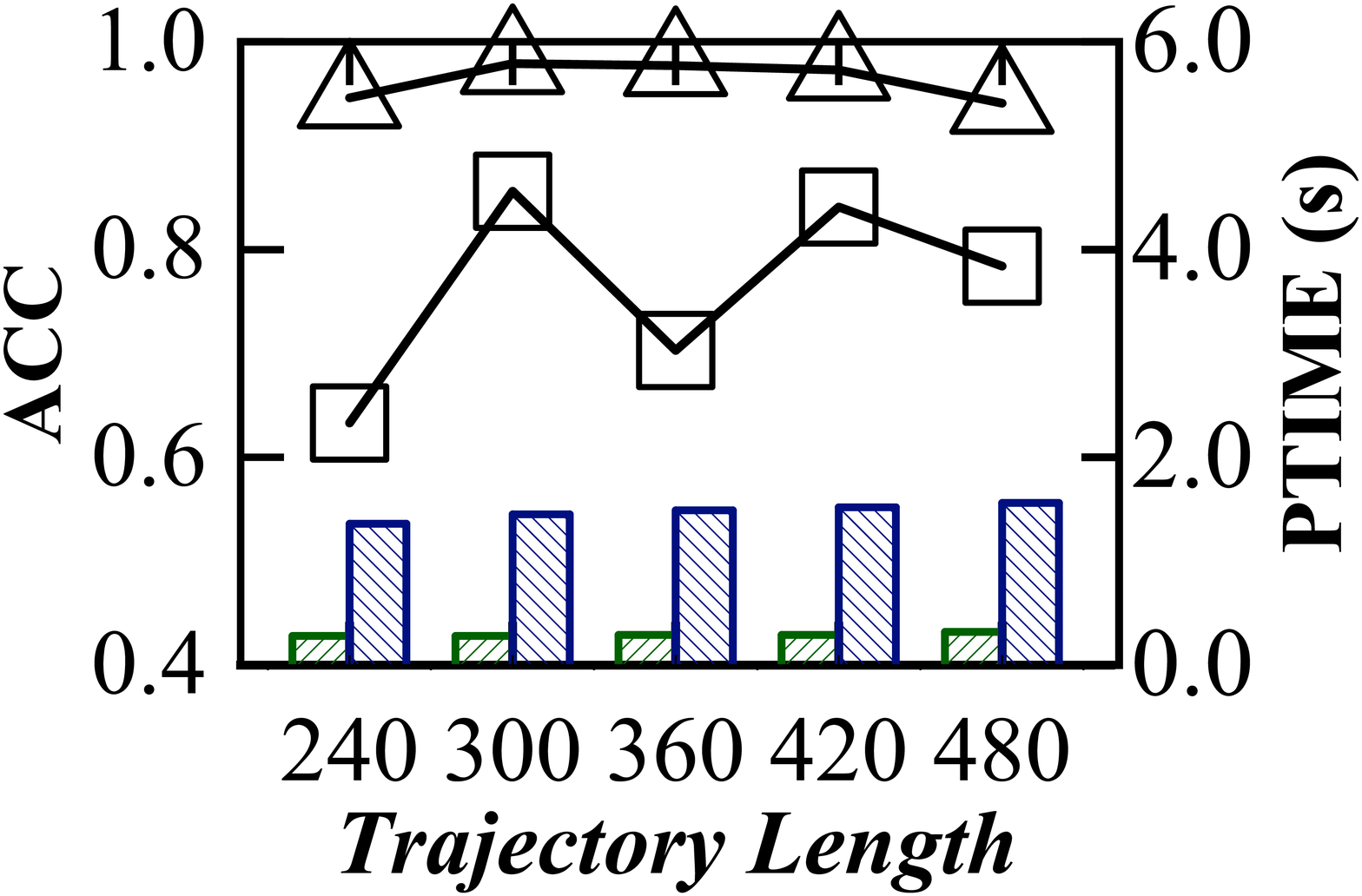}}
	\subfloat[Taxi/F1/TTIME]{
		\includegraphics[width=0.24\textwidth]{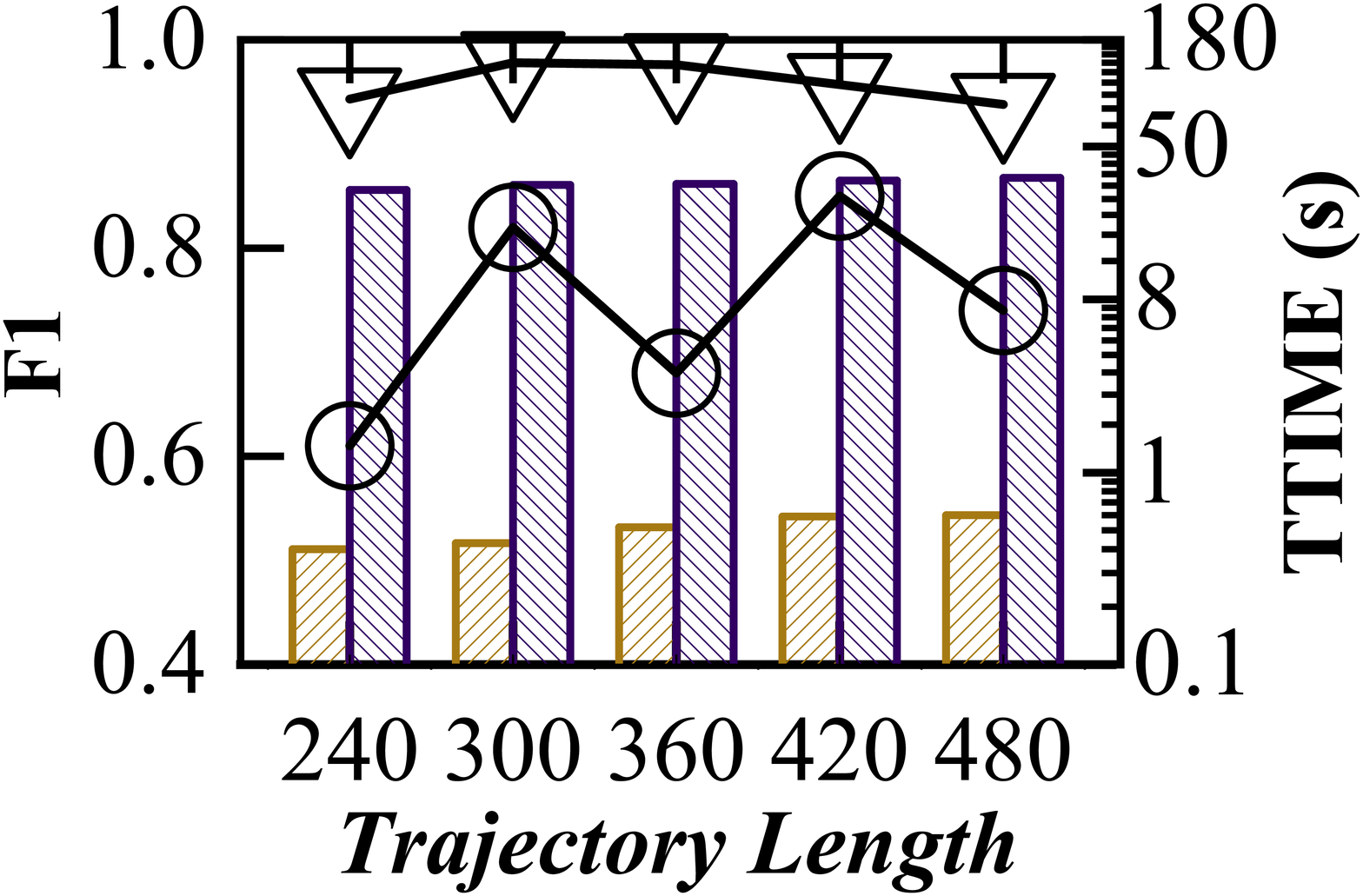}}\\
	\vspace{-2.5mm}
	\centering
	\hspace{-0.25cm}
	\subfloat[Train/ACC/PTIME]{
		\includegraphics[width=0.24\textwidth]{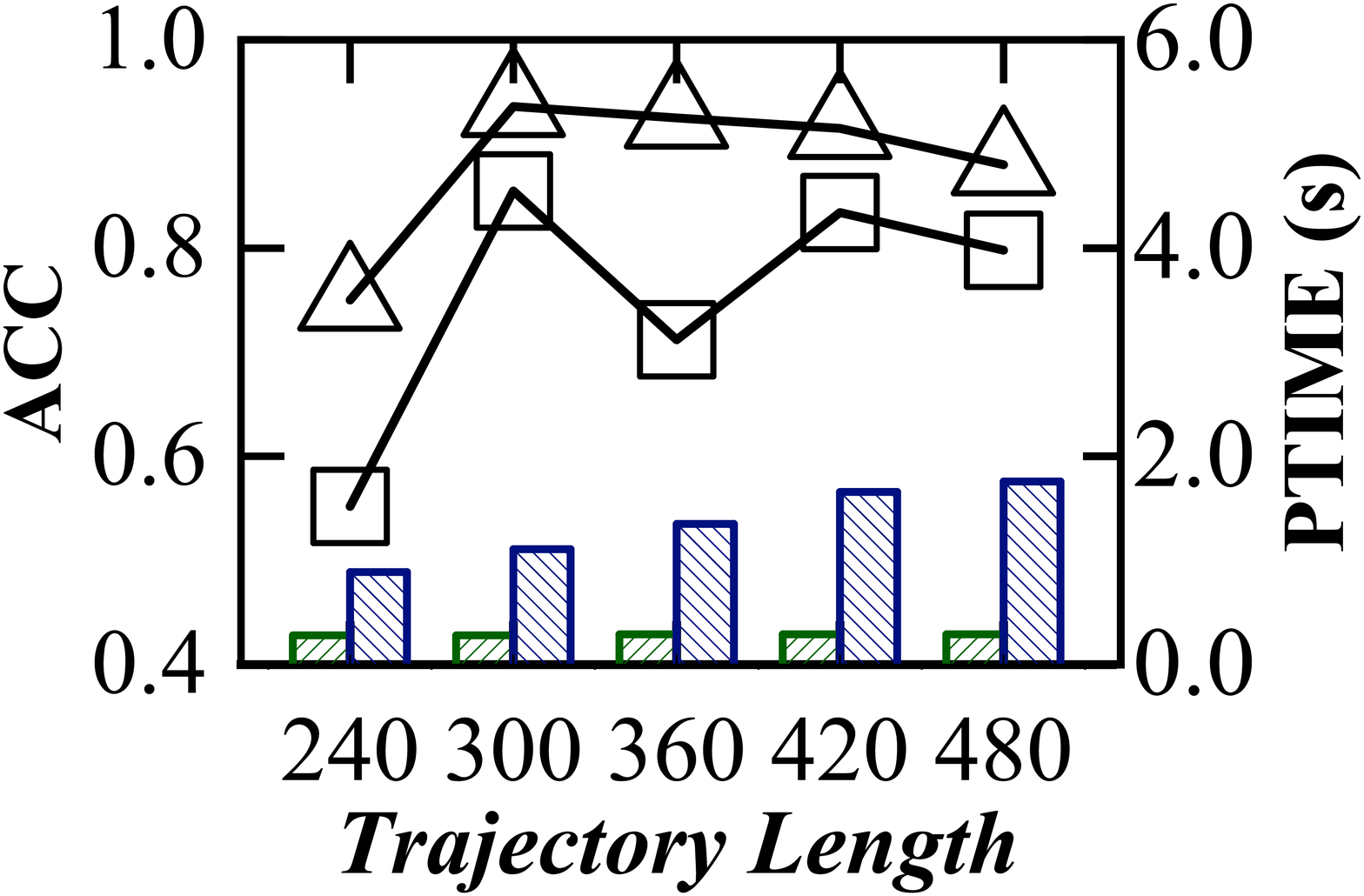}}
	\subfloat[Train/F1/TTIME]{
		\includegraphics[width=0.24\textwidth]{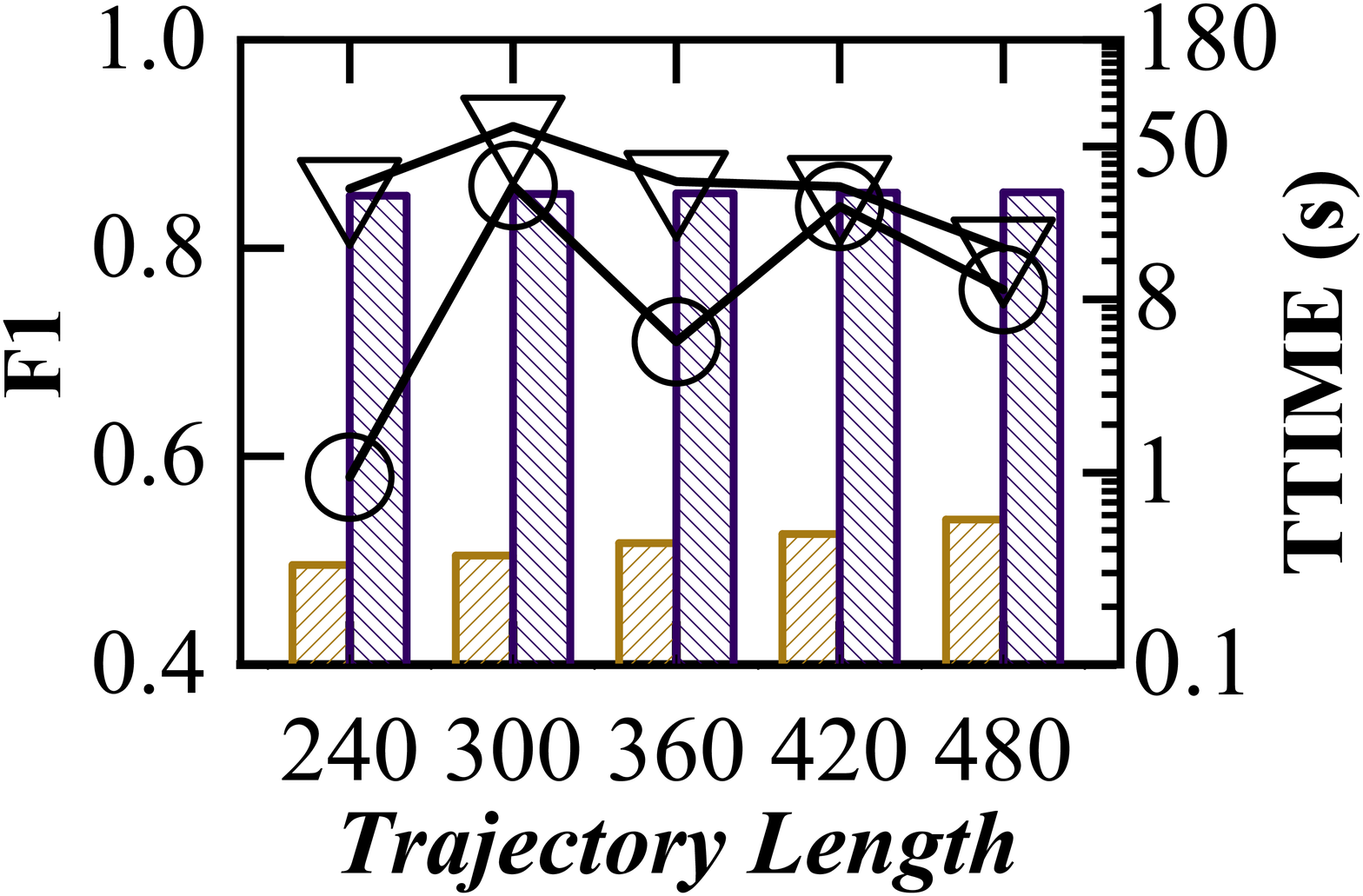}}
	\subfloat[GeoLife/ACC/PTIME]{
		\includegraphics[width=0.24\textwidth]{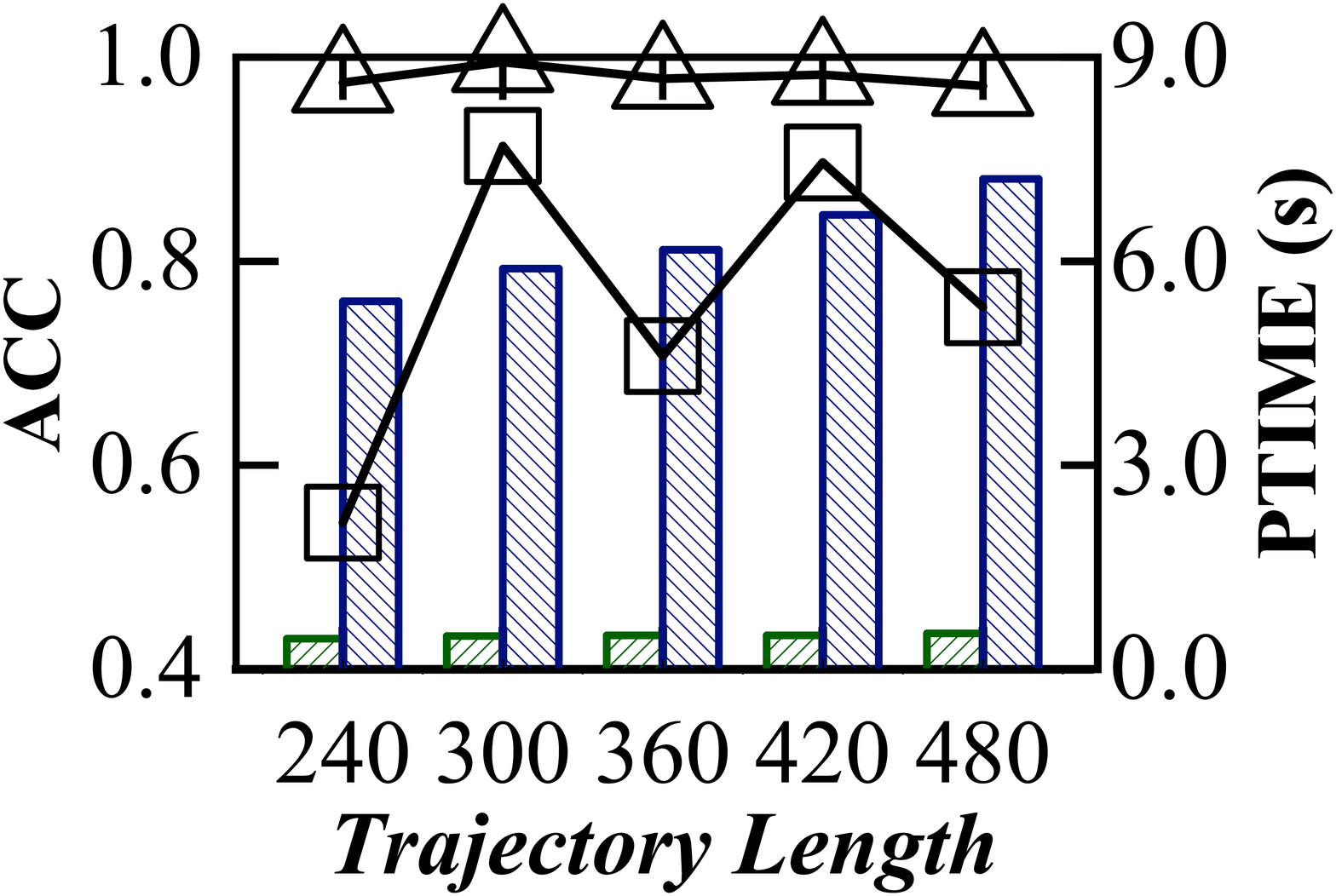}}
	\subfloat[Geolife/F1/TTIME]{
		\includegraphics[width=0.24\textwidth]{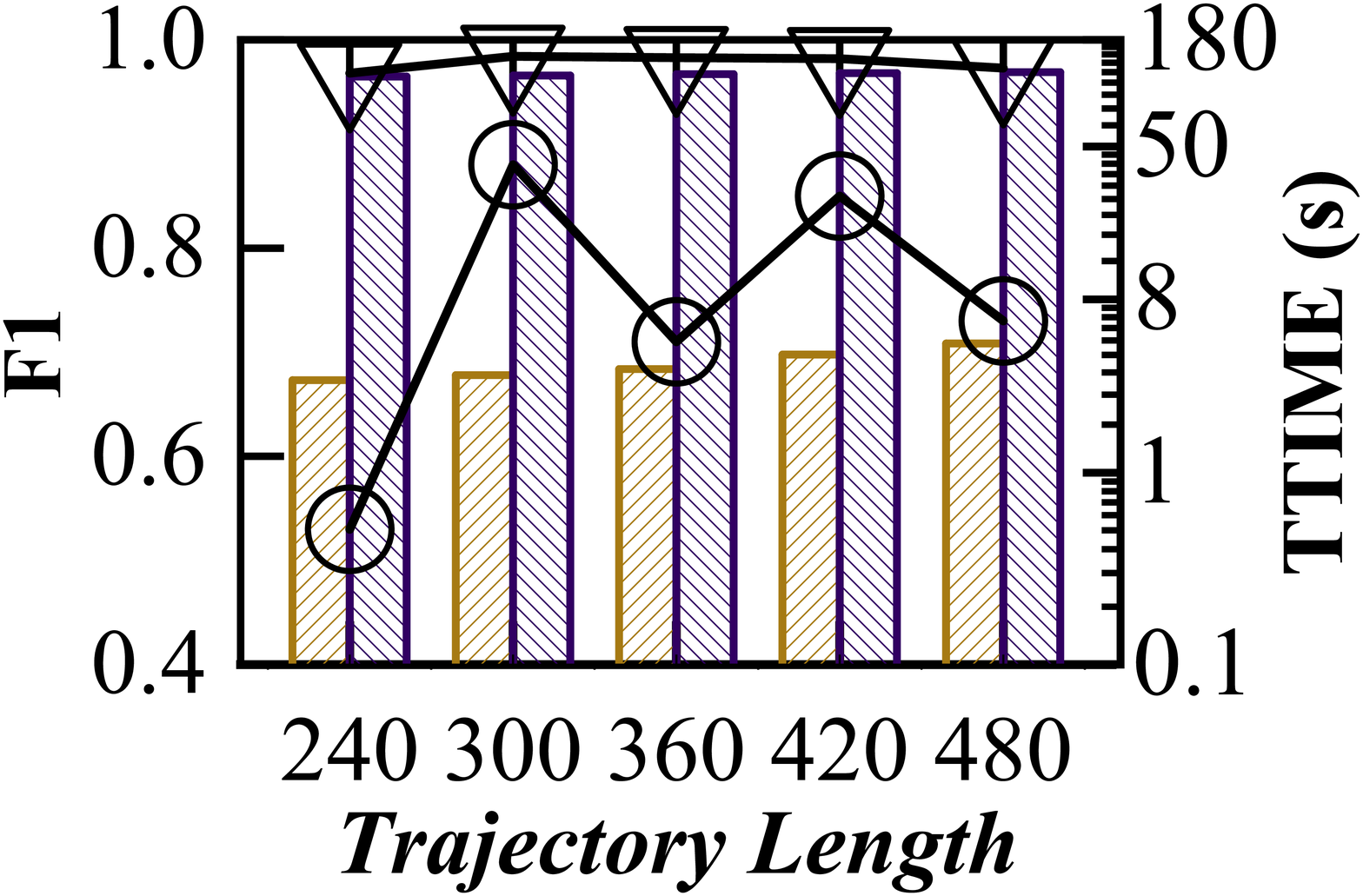}}
	\caption{Model Scalability Evaluation vs. Trajectory Length}
	\label{fig:scalabilitylength}
    \vspace{-5mm}
\end{figure*}

\subsection{Model Efficiency Evaluation}

We proceed to evaluate the model efficiency in terms of both the training phase (i.e., \textit{TTIME}) and predicting/testing phase (i.e., \textit{PTIME}), using the eight datasets. {Here, we only compare Estimator with start-of-the-art deep-learning based methods because the performance of trajectory mode classification of the machine-learning based methods is low. The results are presented in Table~\ref{tab:efficiency}.} The first observation is that Estimator outperforms all the competitors in both training and testing phases. In the training phase, Estimator runs up to 40 times faster than the state-of-the-art competitor ST-GRU. In the classification phase, Estimator achieves up to ${94.6\%}$ efficiency improvement on Geolife dataset. The second observation is that the efficiency of RNN-based ST-GRU is much lower than that of convolution based methods (i.e., Estimator and SECA). This is because RNN processes trajectory points one by one, while convolution methods process the trajectory in parallel. 

\subsection{Model Scalability Evaluation}
\label{subsec:scalability}

We study the scalability of Estimator by comparing with SECA and ST-GRU, where the data size of each dataset is varied from 20\% to 100\% of the
total dataset size.

Fig.~\ref{fig:scalability} depicts the experimental results by varying cardinality of datasets from $20\%$ to $100\%$, where the percentages denote the $20\%$, $40\%$, $60\%$, $80\%$, $100\%$ of the entire dataset. Note that, we keep the origin spatial region when sampling $20\%$, $40\%$, $60\%$, $80\%$, $100\%$ trajectories from the origin dataset. We use \textit{ACC}, \textit{F1}, \textit{TTIME} and \textit{PTIME} to evaluate the quality and efficiency. 
The first observation is that \textit{ACC} and ${F1}$ 
increase with the growth of data cardinality. The reason is that the model becomes better with the increasing amount of training data. Second, Estimator achieves the highest \textit{ACC} and ${F1}$. 
This is because we not only consider the trajectories' explicit features but also explore their implicit features, where the model learns comprehensive information and thus  classification is improved. The third observation is that \textit{TTIME} and \textit{PTIME} of ST-GRU increase rapidly with the growth of cardinality, because more training and testing data are used. However,  \textit{TIME} and \textit{PTIME} of SECA and Estimator increase slightly with the growth of cardinality. Moreover, SECA and Estimator have much smaller  \textit{TTIME} and \textit{PTIME} than ST-GRU model. This is because GRU (RNN) model sequentially processes trajectories, which degrades the parallel processing. On the other hand, SECA and Estimator utilize convolution layers to process trajectories in parallel. Overall, Estimator shows high scalability on both quality and efficiency, which proves its capability of performing large-scale classification.

Fig.~\ref{fig:scalabilitylength} reports the results when varying the trajectory length from ${240}$ to ${480}$. Here,  trajectory length denotes the average number of points of a trajectory. Note that SECA develops a 
segmentation method, which is unable to pre-define the trajectory length. Thus, we only compare Estimator with ST-GRU in this set of experiments. The first observation is that Estimator achieves the best performance. This is because, when the trajectory length is fixed, Estimator can extract more features of trajectories than ST-GRU. In the following experiments, we fix the trajectory length to 300 points for Estimator to achieve the best performance.
Second, ${ACC}$ and ${F1}$ of ST-GRU fluctuate drastically as the trajectory length increases. Specifically, ST-GRU achieves highest ${ACC}$ and ${F1}$ when the trajectory length equals to ${300}$ on Walk, Bike, Bus, Private Car, Taxi, Train and Geolife datasets, and achieves highest ${ACC}$ and ${F1}$ when  the trajectory length equals to ${420}$ on Subway dataset. 
This is because, on the one hand, ST-GRU can extract more deep features with the increasing length of data; on the other hand, {increasing data lengths require larger receptive field and more dilated convolutional layers, leading to more parameters and higher model complexity when capturing the features of a trajectory. This leads to over-fitting, which degrades the model performance.} 
By contrast, ${ACC}$ and ${F1}$ of 
Estimator are stable, which proves its robustness. In addition, the training time (\textit{TTIME}) and testing time (\textit{PTIME}) increase with the growth of trajectory length, as more sample points need to be processed. Overall, this set of experiments proves that Estimator is able to perform large-scale transportation mode classification.

\begin{table*}[]
\vspace{-2mm}
\caption{Ablation Study of Estimator vs. ACC, F1}
\vspace{-2mm}
 \setlength{\tabcolsep}{1.6mm}{
\begin{tabular}{|c|cc|cc|cc|cc|cc|cc|cc|cc|}
\hline
{\makebox[0.13\textwidth][c]{Variants}} & \multicolumn{2}{c|}{Geolife}       & \multicolumn{2}{c|}{Walk}          & \multicolumn{2}{c|}{Bike}          & \multicolumn{2}{c|}{Bus}           & \multicolumn{2}{c|}{Subway}        & \multicolumn{2}{c|}{Pri.Car}       & \multicolumn{2}{c|}{Taxi}          & \multicolumn{2}{c|}{Train}         \\ \cline{2-17} 
                          & \multicolumn{1}{c|}{ACC}   & F1    & \multicolumn{1}{c|}{ACC}   & F1    & \multicolumn{1}{c|}{ACC}   & F1    & \multicolumn{1}{c|}{ACC}   & F1    & \multicolumn{1}{c|}{ACC}   & F1    & \multicolumn{1}{c|}{ACC}   & F1    & \multicolumn{1}{c|}{ACC}   & F1    & \multicolumn{1}{c|}{ACC}   & F1    \\ \hline
CNN Model                 & \multicolumn{1}{c|}{0.731} & 0.722 & \multicolumn{1}{c|}{0.352} & 0.331 & \multicolumn{1}{c|}{0.867} & 0.846 & \multicolumn{1}{c|}{0.771} & 0.745 & \multicolumn{1}{c|}{0.345} & 0.352 & \multicolumn{1}{c|}{0.495} & 0.472 & \multicolumn{1}{c|}{0.513} & 0.484 & \multicolumn{1}{c|}{0.243} & 0.241 \\ \hline
CNN-TCN-B                 & \multicolumn{1}{c|}{0.964} & 0.913 & \multicolumn{1}{c|}{0.959} & 0.934 & \multicolumn{1}{c|}{0.961} & 0.941 & \multicolumn{1}{c|}{0.969} & 0.944 & \multicolumn{1}{c|}{0.873} & 0.887 & \multicolumn{1}{c|}{0.946} & 0.918 & \multicolumn{1}{c|}{0.946} & 0.923 & \multicolumn{1}{c|}{0.708} & 0.710  \\ \hline
CNN-TCN-H                 & \multicolumn{1}{c|}{0.984} & 0.939 & \multicolumn{1}{c|}{0.978} & 0.947 & \multicolumn{1}{c|}{0.962} & 0.941 & \multicolumn{1}{c|}{0.969} & 0.944 & \multicolumn{1}{c|}{0.875} & 0.887 & \multicolumn{1}{c|}{0.973} & 0.918 & \multicolumn{1}{c|}{0.961} & 0.930  & \multicolumn{1}{c|}{0.708} & 0.710  \\ \hline
CNN-TCN-P                 & \multicolumn{1}{c|}{0.992} & 0.974 & \multicolumn{1}{c|}{0.993} & 0.968 & \multicolumn{1}{c|}{0.995} & 0.978 & \multicolumn{1}{c|}{0.987} & 0.981 & \multicolumn{1}{c|}{0.982} & 0.950  & \multicolumn{1}{c|}{0.989} & 0.964 & \multicolumn{1}{c|}{0.979} & 0.938 & \multicolumn{1}{c|}{0.936} & 0.857 \\ \hline
\end{tabular}}
\label{tab:ablation}
\vspace{-3mm}
\end{table*}
\begin{table*}[]
\caption{Ablation Study of Estimator vs. TTIME, PTIME}
\vspace{-2mm}
 \setlength{\tabcolsep}{0.4mm}{
\begin{tabular}{|c|cc|cc|cc|cc|cc|cc|cc|cc|}
\hline
{\makebox[0.16\textwidth][c]{Variants}} & \multicolumn{2}{c|}{Geolife}       & \multicolumn{2}{c|}{Walk}          & \multicolumn{2}{c|}{Bike}          & \multicolumn{2}{c|}{Bus}           & \multicolumn{2}{c|}{Subway}        & \multicolumn{2}{c|}{Pri.Car}       & \multicolumn{2}{c|}{Taxi}          & \multicolumn{2}{c|}{Train}         \\ \cline{2-17} 
                          & \multicolumn{1}{c|}{TTIME} & PTIME & \multicolumn{1}{c|}{TTIME} & PTIME & \multicolumn{1}{c|}{TTIME} & PTIME & \multicolumn{1}{c|}{TTIME} & PTIME & \multicolumn{1}{c|}{TTIME} & PTIME & \multicolumn{1}{c|}{TTIME} & PTIME & \multicolumn{1}{c|}{TTIME} & PTIME & \multicolumn{1}{c|}{TTIME} & PTIME \\ \hline
CNN Model                 & \multicolumn{1}{c|}{14.34} & 4.53  & \multicolumn{1}{c|}{7.41}  & 2.23  & \multicolumn{1}{c|}{7.37}  & 2.22  & \multicolumn{1}{c|}{7.47}  & 2.21  & \multicolumn{1}{c|}{7.23}  & 2.21  & \multicolumn{1}{c|}{7.66}  & 2.24  & \multicolumn{1}{c|}{7.62}  & 2.2   & \multicolumn{1}{c|}{7.18}  & 2.16  \\ \hline
CNN-TCN-B                 & \multicolumn{1}{c|}{3.26}  & 0.31  & \multicolumn{1}{c|}{0.96}  & 0.27  & \multicolumn{1}{c|}{1.04}  & 0.31  & \multicolumn{1}{c|}{0.64}  & 0.32  & \multicolumn{1}{c|}{0.52}  & 0.30  & \multicolumn{1}{c|}{0.62}  & 0.29  & \multicolumn{1}{c|}{0.61}  & 0.29  & \multicolumn{1}{c|}{0.37}  & 0.29  \\ \hline
CNN-TCN-H                 & \multicolumn{1}{c|}{3.25}  & 0.30  & \multicolumn{1}{c|}{0.95}  & 0.27  & \multicolumn{1}{c|}{1.04}  & 0.31  & \multicolumn{1}{c|}{0.63}  & 0.32  & \multicolumn{1}{c|}{0.52}  & 0.30  & \multicolumn{1}{c|}{0.6}   & 0.29  & \multicolumn{1}{c|}{0.6}   & 0.28  & \multicolumn{1}{c|}{0.37}  & 0.29  \\ \hline
CNN-TCN-P                 & \multicolumn{1}{c|}{3.23}  & 0.30  & \multicolumn{1}{c|}{0.95}  & 0.27  & \multicolumn{1}{c|}{1.03}  & 0.29  & \multicolumn{1}{c|}{0.63}  & 0.29  & \multicolumn{1}{c|}{0.5}   & 0.29  & \multicolumn{1}{c|}{0.6}   & 0.28  & \multicolumn{1}{c|}{0.6}   & 0.28  & \multicolumn{1}{c|}{0.37}  & 0.28  \\ \hline
\end{tabular}}
\label{tab:ablation2}
\vspace{-4mm}
\end{table*}

\subsection{Ablation Study}

We study the effectiveness of four key components (i.e., CNN model, CNN-TCN based model, hidden features extraction, data partition and parallel computing) embedded in Estimator. 
Table~\ref{tab:ablation} and Table~\ref{tab:ablation2} report the results on eight datasets, where CNN-TCN-B denotes the CNN-TCN based model, CNN-TCN-H denotes CNN-TCN-B with hidden features extraction, and CNN-TCN-P denotes CNN-TCN-H with data partition and parallel computing. Obviously, CNN-TCN-P is Estimator essentially.

The first observation is that CNN-TCN-B outperforms CNN on all the datasets. This is because,  TCN component can greatly improve the CNN model performance in terms of both quality and efficiency for transportation mode classification, which verifies the superior performance of TCN model when processing trajectories. The second observation is that CNN-TCN-H outperforms CNN-TCN-B in terms of \textit{ACC} and ${F1}$ especially on \textit{Geolife}, \textit{taxi} and \textit{private car} datasets. This is because, employing the hidden periodic features between taxis and private cars can improve the model performance for transportation mode classification. Finally, CNN-TCN-P 
outperforms all the variants of Estimator in terms of \textit{ACC} and ${F1}$, because it considers the different traffic conditions in the city. 
Note that, \textit{ACC} and ${F1}$ are enhanced significantly on \textit{train} dataset in terms of \textit{ACC} and ${F1}$. The reason is that urban railways are mainly located in the suburbs, and  data partition (i.e., urban area, suburb, city center) can reduce the negative effects of other traffic modes on the classification. 
Overall, this set of experiments validate the effectiveness of four key components of Estimator. 
\vspace{-3mm}
\subsection{Model Hyperparameters Analytics}
We evaluate the performance of Estimator under different settings of hyperparameters, including the number of epochs, dilation base, the number of network layers, and the number of hidden units in each layer. Here we only report the experimental results on GeoLife dataset, because the performance on other datasets is similar to that on GeoLife dataset.
Note that, the number of epochs used for training reflects the convergence's speed of the model; dilation base used in  dilation convolutions reflects the receptive field in TCN; while the number of network layers and the number of hidden units in each layer are also important hyperparameters in deep learning, which reflects the network depth and the capability of capturing features. 

\begin{figure}[tb]
 \centering
 \hspace{-4mm}
 \includegraphics[width=0.48\textwidth]{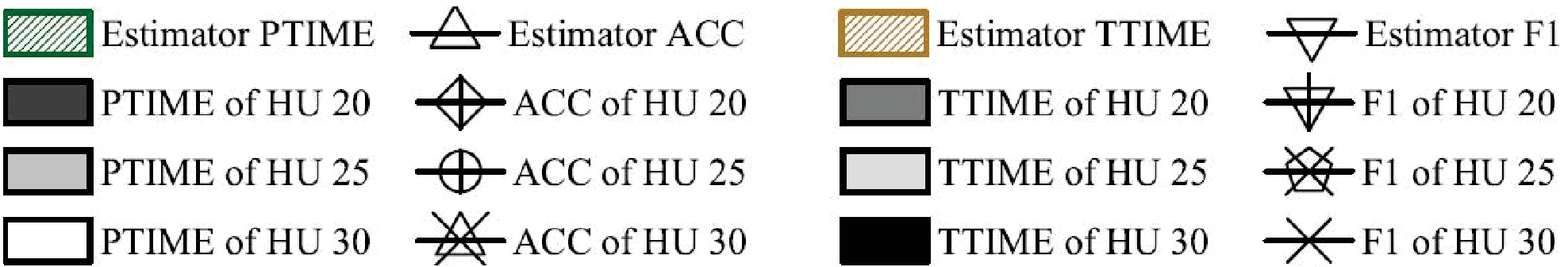}\\
 \hspace{-0.6cm}
 \vspace{-6mm}
\subfloat[\# Epochs/ACC/PTIME]{
		\includegraphics[width=0.24\textwidth]{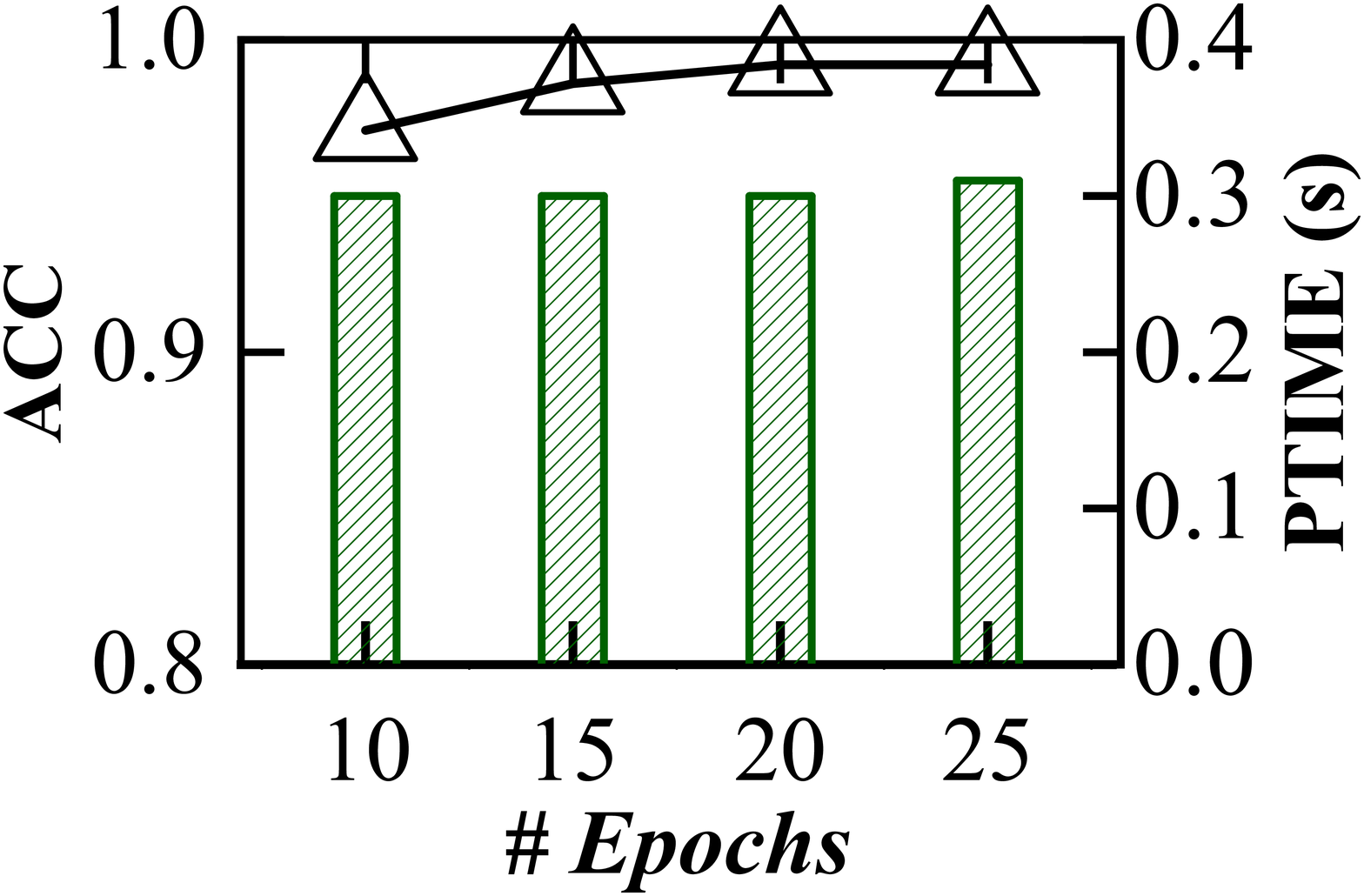}}
\subfloat[\# Epochs/F1/TTIME]{
		\includegraphics[width=0.24\textwidth]{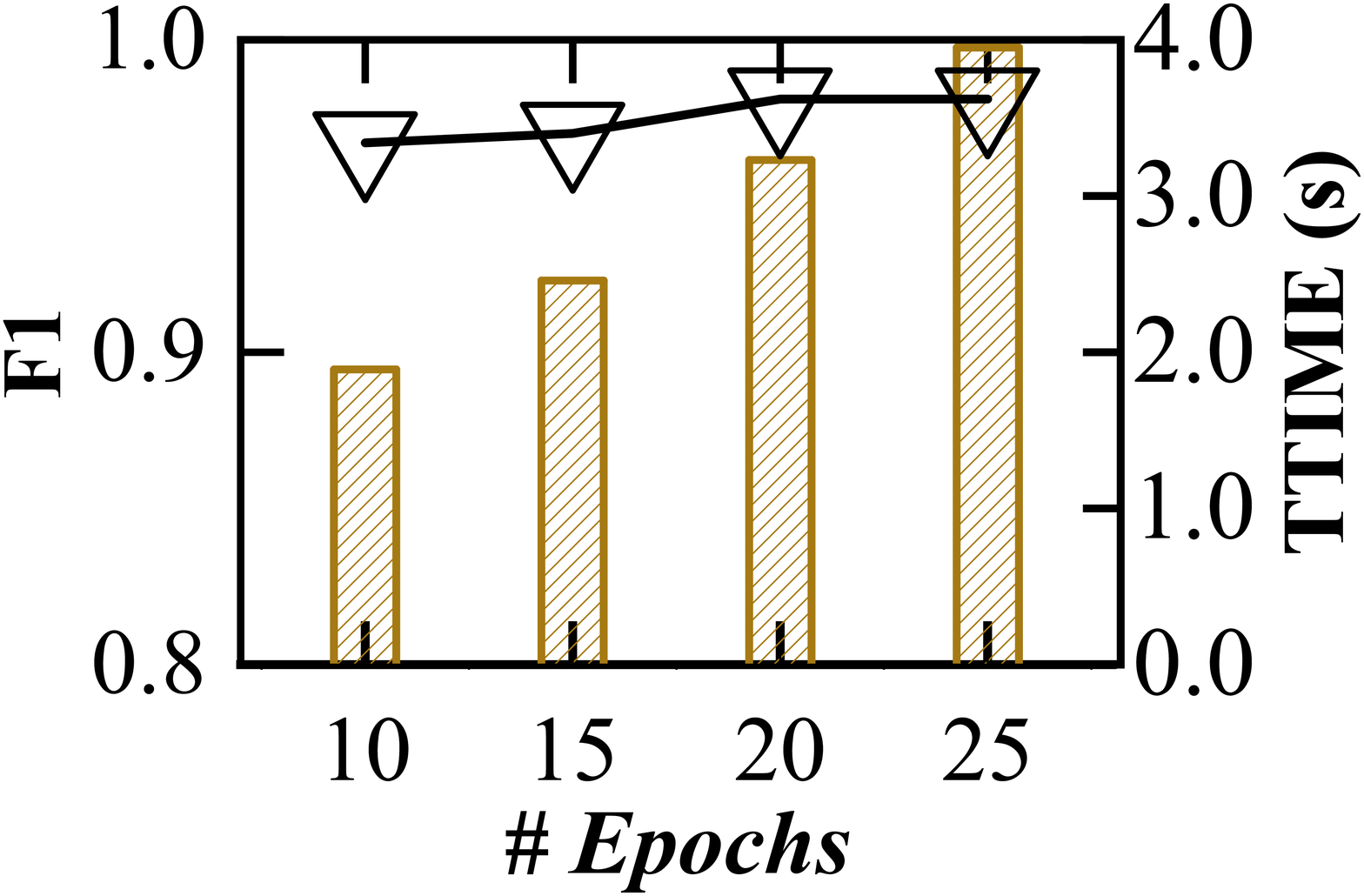}}\\
\vspace{2mm}
 \hspace{-0.6cm}
\subfloat[Dilation Base/ACC/PTIME]{
         \includegraphics[width=0.24\textwidth]{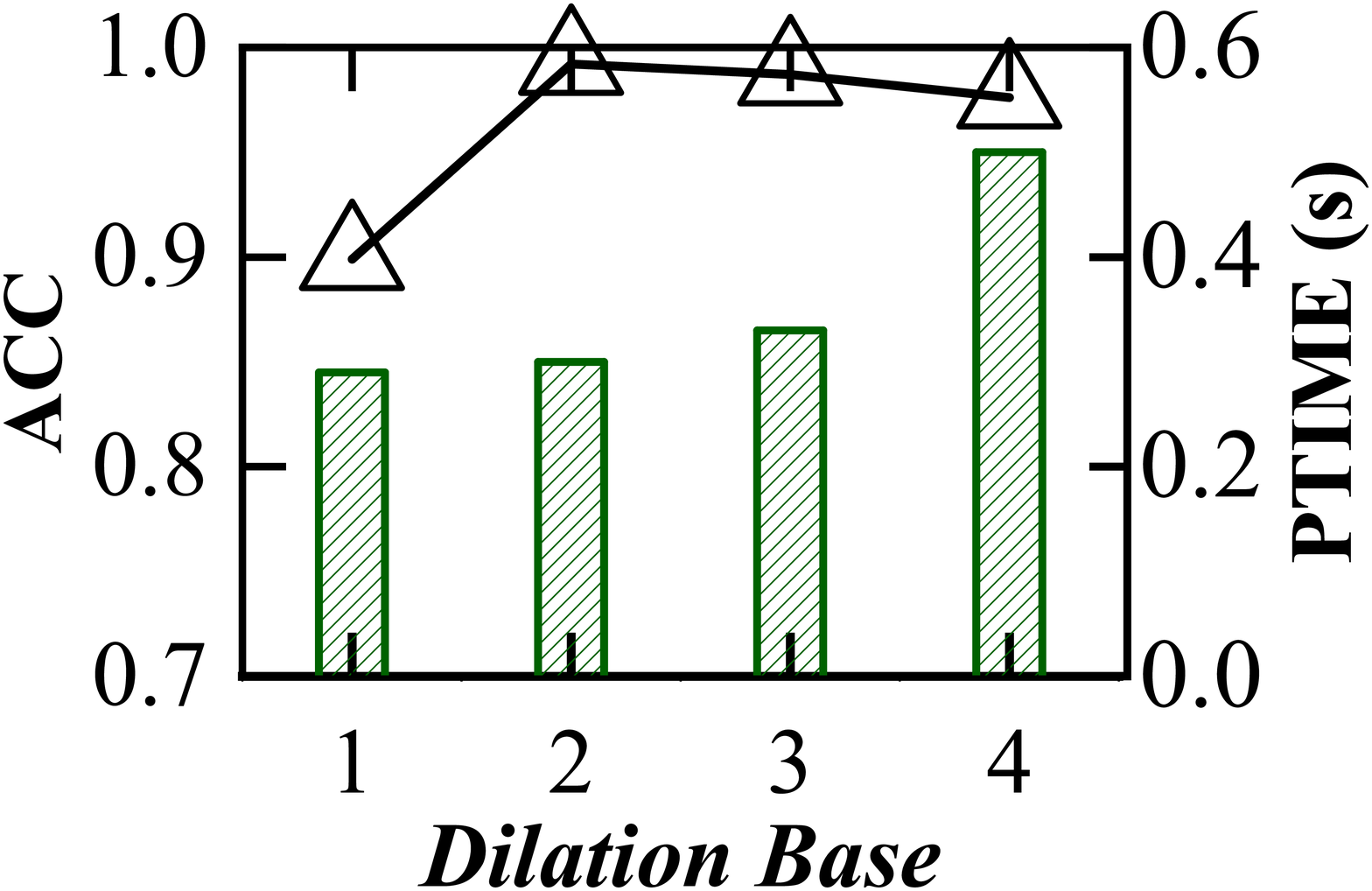}}
\subfloat[Dilation Base/F1/TTIME]{
		\includegraphics[width=0.24\textwidth]{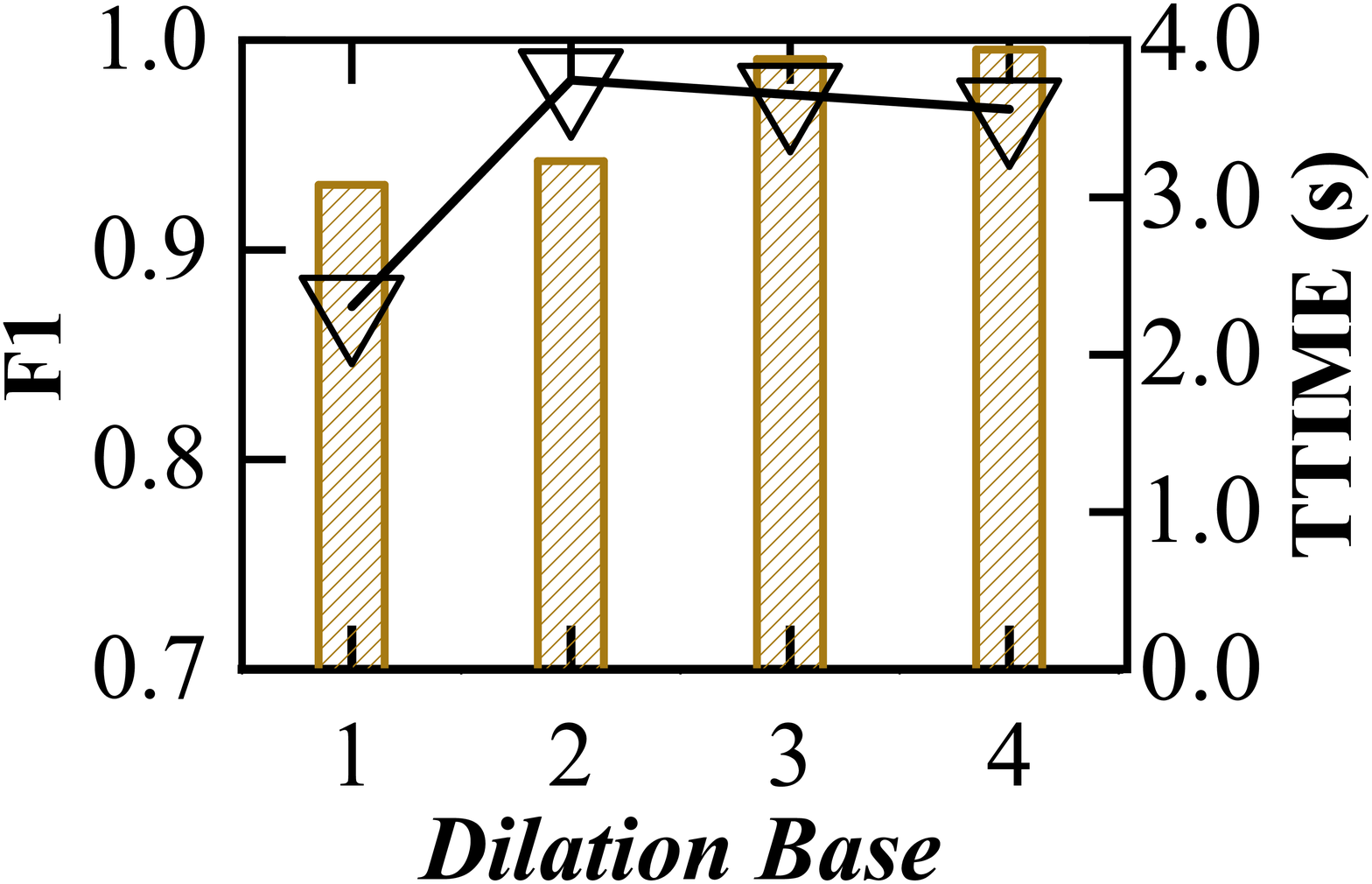}}\\
\vspace{-3mm}
 \hspace{-0.6cm}
\subfloat[\# Layers and Units/ACC/PTIME]{
		\includegraphics[width=0.24\textwidth]{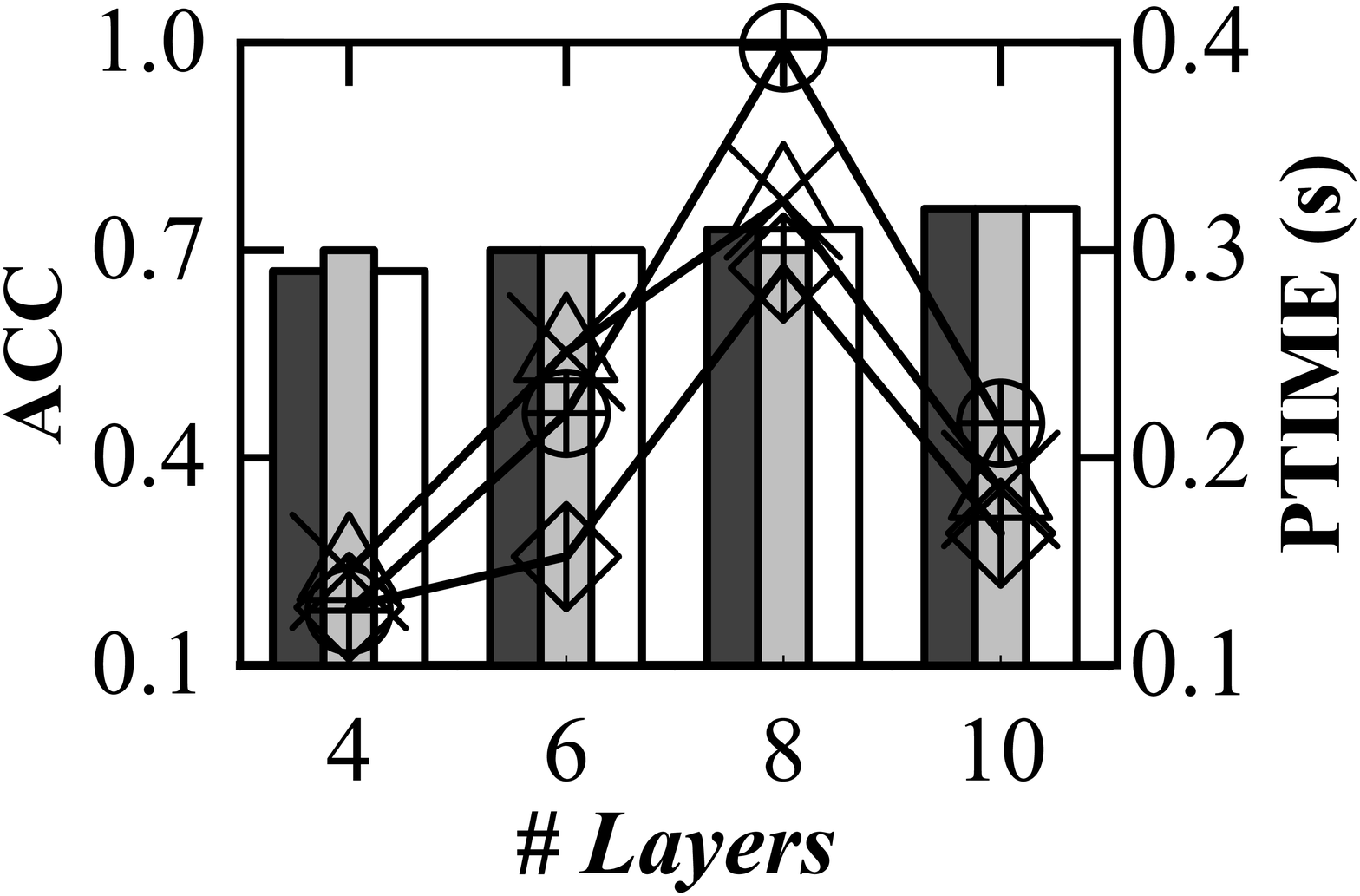}}
\subfloat[\# Layers and Units/F1/TTIME]{
         \includegraphics[width=0.24\textwidth]{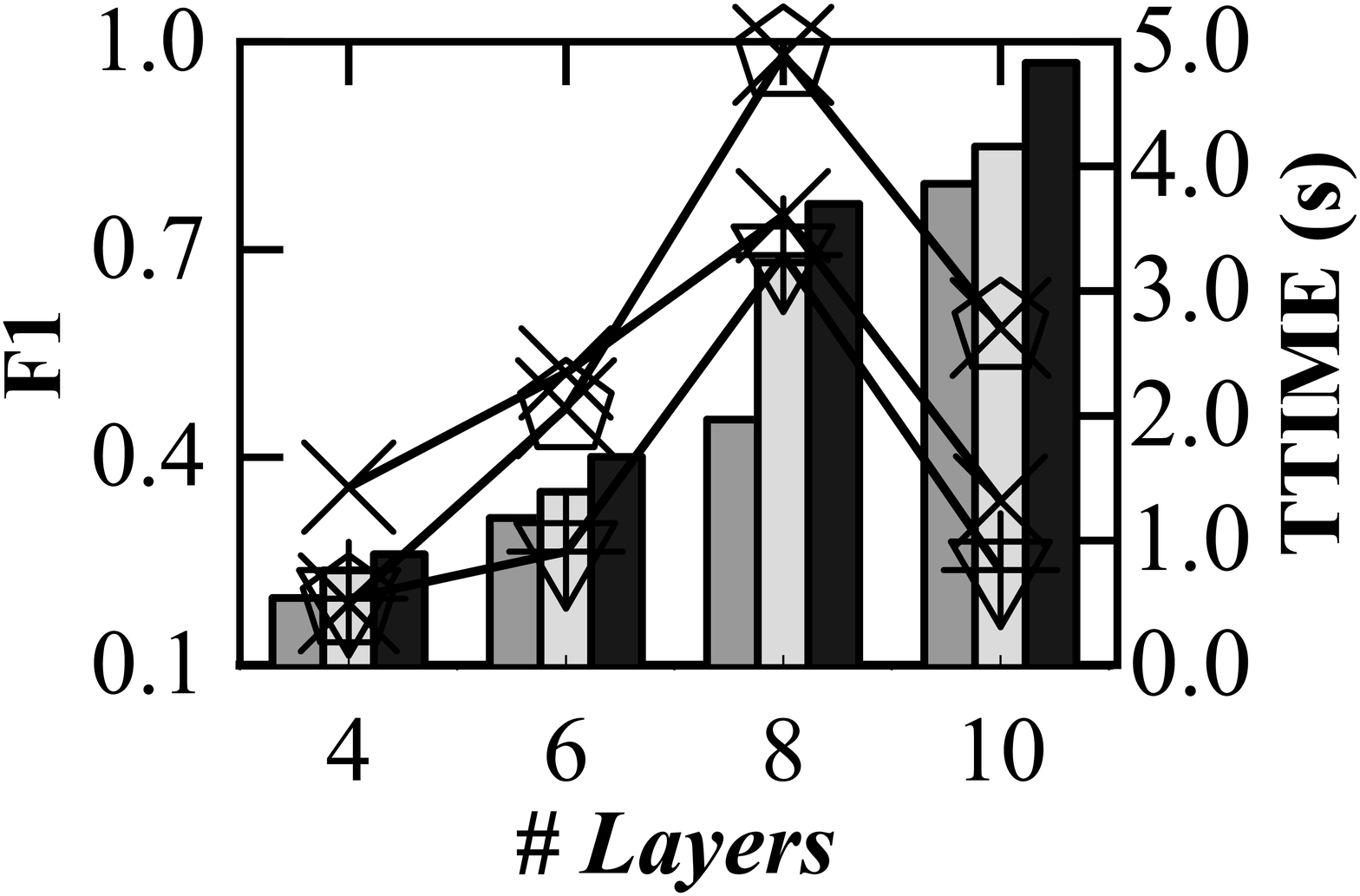}}\\
 \caption{Model Evaluation of Various TCN Hyperparameters} 
 \vspace{-6mm}
 \label{fig:parameters}
\end{figure}

Figs.~\ref{fig:parameters}(a) and~\ref{fig:parameters}(b) show the results when varying the number of epochs. As observed, $\textit{ACC}$ and ${F1}$ first increase with the growth of number of epochs and then keep stable (i.e., the model converges).  Note that, Estimator performs well in terms of $\textit{ACC}$ and ${F1}$ (i.e., more than 0.95) even when the number of epochs is 10. In addition, the training time \textit{TTIME} increases slightly with the growth of number of epochs, while the testing time \textit{PTIME} is relatively stable (i.e, always around {0.3s}). This verifies the robustness of Estimator.

Figs.~\ref{fig:parameters}(c) and~\ref{fig:parameters}(d) show the results under various  dilation coefficients. As the upward growth dilated coefficient can be computed as $d=b^i$, we set ${b=1,2,3,4}$ to adjust the dilation coefficient. The first observation is that, Estimator achieves the smallest $\textit{ACC}$ and ${F1}$ when $b = 1$. 
This is because that the receptive field does not been enlarged if the dilated coefficient is fixed to 1, which fails to capture the relevance between nodes with a long interval. The second observation is that Estimator performs the best when $b = 2$. 
This is because when $b = $ 3 or 4, the receptive field becomes too large. In this case, some local detailed features are lost, which causes negative effects on classification. In terms of efficiency, the training time $\textit{TTIME}$ first increases and then remains stable with the growth of $b$. This is because the receptive field of TCNs has covered the entire input trajectory when $b$ reaches 3, and there is no additional data to be trained. The testing time $\textit{PTIME}$ first slightly increases as $b$ grows from 1 to 3 but significantly increases when $b$ grows from 3 to 4. This is because, when $b$ grows from 3 to 4, the model processes lots of redundant data due to the large receptive field.

Figs.~\ref{fig:parameters}(e) and~\ref{fig:parameters}(f) show the results under various numbers of hidden units (denoted as HU = 20, 25, 30) when varying the number of convolutional layers from 4 to 10. 
The first observation is that, $\textit{ACC}$ and ${F1}$ first increase and then drop with the growth of the number of layers. The reason is that, on the one hand, more layers reduce the network errors and improve the accuracy; on the other hand, a large number of layers lead to over-fitting.
The second observation is that $\textit{ACC}$ and ${F1}$ are not linearly correlated with  the number of hidden units. This is because, given different numbers of network layers,  the number of hidden units that achieves the best performance varies. 
In addition, training time $\textit{TTIME}$ increases slowly with the growth of numbers of layers and units, while the testing time $\textit{PTIME}$ is stable (i.e., always around {0.3s}), which suggests the robustness of Estimator. As observed, Estimator achieves the highest $\textit{ACC}$, i.e., ${99.2\%}$, and the highest ${F1}$, i.e., ${0.981}$, when the depth of neural networks (i.e., the number of network layers) is 8 and the number of hidden units is 25. Hence, we set the number of convolutional layers to 8 and the number of hidden units in each layer to 25 for Estimator in order to achieve the best performance.

%% file: RelatedWork.tex
\section{Related Work}
\label{sec:related}
We proceed to review the related work on transportation mode classification, including machine learning based methods and deep learning based methods.

\subsection{Machine Learning based Classification} 
Earlier proposals of machine learning based trajectory mode classification typically contain three processing phases: trajectory representation, feature extraction, and classifier selection. First, a trajectory can be represented as different formats~\cite{15overview}, such as a series of timestamped sample points, a sequence of trajectory segments, and a set of image pixels.
Next, machine learning based methods manually extract spatio-temporal features from raw trajectories or trajectory images, which is called feature extraction. As an example, Xu et al. \cite{19temporalgraphs} extract the characteristics of every sample point in a trajectory considering both temporal information (e.g., stay time) and spatial information (e.g., speed and acceleration). Similarly, Dabiri et al.~\cite{19TKDE} transform each trajectory segment into a 4-channel characteristics tensor composed of the jerk, speed, acceleration, and relative distance. To extract features from mapped trajectory images, Endo et al.~\cite{16Endo} capture time interval characteristics and geographic location information of the trajectories with grid pixels and then use existing image classification models to classify transportation modes. Finally, these mobility features captured from either raw trajectories or trajectory images are fed into existing popular machine learning classifiers such as $k$-nearest neighbour ($k$NN)~\cite{08zhengyuMobility}, Hidden Markov Model (HMM)~\cite{14HMM}, Random Forest (RF)~\cite{10zhengyu}, and Support Vector Machine (SVM)~\cite{12transportationSVM}. However, they require to extract features manually,
which is time-consuming, noise sensitive, and fails to capture the non-linear and hidden dependencies of trajectories. Thus, their performance is limited. 
\vspace{-4mm}
\subsection{Deep Leaning based Classification}
Deep neural networks have been widely used in text classification~\cite{20Text,20Text2} and image classification~\cite{20image,20image2}, as it can be performed without manual feature extraction. 
This leads to the proposals of deep learning based transportation mode classification, including CNN-based methods, RNN-based methods, and CNN-RNN based methods. The CNN-based methods typically use convolution neural networks (CNNs) to capture moving object features. Specifically, Endo et al.~\cite{19TKDE} convert each trajectory to an image. An image is composed of a set of grids, each of which represent its geographic dependencies. Based on this,  the trajectory classification problem is transformed to the image classification problem.

There are several recent works that capture trajectory features by employing spatio-temporal graph attention networks (i.e, GNN-based models)~\cite{GNN1,GNN2}. Note that, spatio-temporal graph can only be applied to road network. However, the movements of some moving objects (e.g., people and bikes) do not follow network constraints. Thus, spatio-temporal graph is mainly applied for traffic prediction, and fails to support transportation mode classification. CNN-based methods only focus on the spatial aspect of trajectories while ignore the temporal correlations between trajectory points. To improve the performance of classification, recurrent neural networks (RNNs) are employed to capture time-series features of trajectories~\cite{19LSTM}. This is because, the RNN-type models are initially designed for time-series representation learning, which is easy to be adapted to trajectory classification~\cite{90RNN}. Compared with CNN-based methods, RNN-based methods attach more importance to the temporal aspect than the spatial aspect.

Recently, the state-of-the-art CNN-RNN based architecture has been developed, which feeds trajectories into both CNN models and RNN models simultaneously to capture both spatial and temporal mobility features of moving trajectories. Liu et al.~\cite{17ISKE} present an end-to-end framework based on Bi-LSTM model for transportation mode classification. Specifically, they design a spatio-temporal GRU model combined with CNN for classification analyses~\cite{19GRU}. Friedrich et al.~\cite{19LSTM} combine LSTM and CNN for transportation mode classification of trajectories from smartphone sensors. As we mainly focus on GPS-based trajectories (cf. Section~\ref{sec:problem}), we only compared the methods using GPS-based trajectories. Although CNN-RNN based models capture the spatio-temporal information of trajectories, RNNs can only scan a trajectory one by one, which is sub-optimal for parallel processing~\cite{18RNNcannot}. 
Consequently, existing RNNs involved methods (including both RNN-based and CNN-RNN based) are not able to support large-scale trajectory/transportation classification. Motivated by the parallel computing of CNNs, temporal convolutional networks (TCNs)~\cite{18TCN} are proposed, which enable reading and embedding time-series sequences in a parallel manner. TCN models have achieved great success in many time-series learning tasks such as action detection~\cite{17action}, probability predictiuon~\cite{20probability}, and machine translation~\cite{18TCN}. To the best of our knowledge, this is the first time to apply TCNs to transportation mode classification tasks.

%% file: Conclusion.tex
\section{Conclusions}
\label{sec:conclusion}

In this paper, we propose an effective and scalable framework, Estimator, for transportation mode classification over GPS trajectories. 
Estimator is able to capture hidden spatial-temporal features of trajectories. To improve classification effectiveness, Estimator exacts hidden features, and considers the varying traffic conditions. To improve classification efficiency and scalablity, Estimator combines CNN and TCN to parallel learn the hidden features. Extensive experiments conducted on eight real-life datasets confirm that Estimator is able to outperform the state-of-the-art methods
in terms of both effectiveness and efficiency. In the future, it is of interest to extend Estimator to other mining tasks such as road planning and transportation emissions detection.